# MorphDistill: Distilling Unified Morphological Knowledge from Pathology Foundation Models for Colorectal Cancer Survival Prediction

Hikmat Khan[1], Usama Sajjad[1], Metin N. Gurcan[2], Anil Parwani[1], Wendy L. Frankel[1], Wei Chen[1] and Muhammad Khalid Khan Niazi[1]

1 Department of Pathology, College of Medicine, The Ohio State University Wexner Medical Center, Columbus, OH 43210, USA

2 Center for Artificial Intelligence Research, Wake Forest University School of Medicine, Winston-Salem, NC 27101, USA

**Correspondence to:** Hikmat Khan, Department of Pathology, College of Medicine, The Ohio State University Wexner Medical Center, Pelotonia Research Center, Columbus, OH 43210, USA. | ORCID: 0009-0008-2550-1991 | Tel: +1-856-405-XXXX | E-mail: Hikmat.khan@osumc.edu | Hikmat.khan179@gmail.com.

**Videos:** A short overview of this work is available at:

- **YouTube Short:**
  - https://www.youtube.com/shorts/5JT7UU_JHwE
  - https://www.youtube.com/shorts/CLIxJ2WDj1A
- **Full Video:**
  - https://www.youtube.com/watch?v=oq78CdJhhto
  - https://www.youtube.com/watch?v=JLHg5vrwcTQ
  - https://youtu.be/zo3xrPfuX2g?si=boYAFC-ZpR-40cCu&t=469
  - https://www.youtube.com/watch?v=Dow8bv37CtI

**Abstract**

**Background**: Colorectal cancer (CRC) remains a leading cause of cancer-related mortality worldwide, and accurate survival prediction is essential for optimizing treatment stratification. While pathology foundation models trained via self-supervised learning capture rich morphological representations, they are typically optimized on heterogeneous datasets and overlook organ-specific features critical for CRC prognostication. The central challenge is that no single foundation model captures the full spectrum of prognostically relevant morphological information.

**Methods**: We introduce MorphDistill, a two-stage framework that unifies complementary morphological knowledge from multiple pathology foundation models into a compact, CRC-specific encoder. In Stage I, a student encoder is trained via dimension-agnostic multi-teacher relational distillation combined with supervised contrastive regularization on large-scale colorectal tissue datasets. This approach preserves inter-sample relational structures from ten state-of-the-art foundation models without requiring explicit feature projection, effectively synthesizing their collective knowledge. In Stage II, the trained encoder is frozen and used to extract patch-level features from whole-slide images, which are subsequently aggregated using attention-based multiple instance learning to predict five-year survival outcomes.

**Results**: Evaluated on the Alliance/CALGB 89803 cohort comprising 424 stage III CRC patients, MorphDistill achieves superior predictive performance with an AUC of 0.68 ± 0.08, representing an approximately 8% relative improvement over the strongest baseline foundation model (UNI v2, AUC 0.63). MorphDistill also attains a C-index of 0.661 and a hazard ratio of 2.52 (95% CI: 1.73–3.65), outperforming all ten foundation model baselines and five established MIL aggregation methods. We further validate the proposed approach on an external independent cohort of 562 patients from The Cancer Genome Atlas (TCGA) colorectal cancer cohort (READ and COAD), where it achieves a superior concordance index of 0.6151±0.07, outperforming existing MIL methods, demonstrating strong generalization across independent datasets. The unified representation consistently improves performance across different aggregation architectures and demonstrates robust generalizability across treatment regimens, sex, and tumor locations.

**Conclusion**: MorphDistill is a unified framework for task-specific representation learning in computational pathology. Rather than training new models from scratch or relying on a single encoder, our framework distills complementary knowledge from multiple foundation models into a specialized representation. This approach offers an efficient, scalable strategy for developing prognostic tools with potential applicability to diverse oncology tasks, although the current study is evaluated on a single multi-institutional clinical trial cohort of stage III colorectal cancer, and further validation on additional independent cohorts and broader disease stages will be necessary to assess generalizability. The source code and implementation are publicly available at https://github.com.mcas.ms/hikmatkhan/MorphDistill.

## 1. Introduction

Colorectal cancer (CRC) remains a critical global health challenge, ranking as the third most common malignancy and the second leading cause of cancer-related mortality worldwide [1]. Epidemiological projections indicate a 60% increase in incidence by 2030, with an estimated 2.2 million new cases and 1.1 million deaths annually [1]. In the United States alone, approximately 154,000 new diagnoses and 54,000 deaths are anticipated in 2025 [2]. Accurate prognostic assessment is complicated by the pronounced morphological heterogeneity of CRC: variations in glandular architecture, stromal composition, tumor budding, and immune infiltration all influence patient outcomes[3,4]. However, conventional histopathological assessment captures only a fraction of this prognostic information [3-8]. The digitization of tissue slides into whole-slide images (WSIs) offers unprecedented opportunities to quantitatively characterize morphological patterns at scale, but fully leveraging this potential requires computational approaches capable of learning CRC-specific prognostic representations from complex histological data [8,9].

Current computational pathology methods predominantly rely on multiple instance learning (MIL) with features extracted from foundation models—large-scale networks pretrained via self-supervised learning on millions of histological patches [10-20]. While these models have demonstrated strong generalization across tasks such as tissue classification, detection, treatment response prediction and molecular subtyping, their performance on prognostic and predictive tasks remains moderate, particularly for survival prediction in specific cancer types such as colorectal cancer [11,13,19,21-32]. This limitation partly stems from the fact that these models are typically trained on heterogeneous histopathology datasets and therefore remain largely domain-agnostic, which cause them to overlook specialized morphological features critical for CRC prognostication, such as glandular distortion, desmoplastic stromal reactions, and mucinous differentiation patterns [11,13,19,21-25]. Moreover, each foundation models is typically pretrained on different datasets and architectures, producing heterogeneous feature spaces that capture distinct aspects of tissue morphology. Relying on any single model therefore limits the ability to leverage the full spectrum of prognostically relevant morphological information available across the foundation model landscape[25,33].

This raises a central question: can the heterogeneous morphological knowledge encoded across multiple pathology foundation models be unified into a specialized representation that surpasses the capabilities of any individual encoder for CRC survival prediction? To address this challenge, we propose MorphDistill, a novel two-stage framework that distills complementary morphological knowledge from multiple foundation models into a compact CRC-specific encoder. In Stage I, we train a student encoder using dimension-agnostic multi-teacher relational distillation combined with supervised contrastive regularization on large-scale CRC tissue datasets [34-36]. Unlike conventional knowledge distillation approaches that require explicit alignment of feature dimensionalities, our method operates on batch-wise relational structures among samples, allowing knowledge transfer from ten state-of-the-art foundation models despite differences in architecture and embedding dimensionality. In Stage II, the resulting encoder is frozen and used to extract patch-level features from whole-slide images, which are subsequently aggregated using MIL to predict five-year survival outcomes.

The design of MorphDistill accounts for the heterogeneous representations of tissue morphology produced by different pathology foundation models. By preserving the relational structures each teacher induces among samples, our framework tires to integrate their complementary knowledge into a unified representation that is both comprehensive and tailored to CRC pathology. We evaluate MorphDistill on the Alliance/CALGB 89803 cohort, a rigorously curated dataset comprising 424 stage III CRC patients with long-term follow-up and demonstrate that it consistently outperforms all ten foundation model baselines and five established MIL aggregation methods for survival prediction. We further validate the generalizability of the learned representation on an external dataset comprising 562 patients from the TCGA colorectal adenocarcinoma (READ and COAD) cohorts.

The main contributions of this work are as follows:

1. We propose a dimension-agnostic multi-teacher distillation framework that transfers complementary morphological knowledge from heterogeneous pathology foundation models into a unified student encoder (MorphDistill) without requiring explicit feature projection. Unlike prior distillation frameworks that rely on feature alignment or teacher selection [33], MorphDistill introduces a dimension-agnostic relational alignment strategy that enables simultaneous knowledge transfer from heterogeneous encoders without architectural constraints, effectively synthesizing collective intelligence from diverse pretrained models.

2. We combine relational distillation with supervised contrastive regularization to learn CRC-specific representations that preserve global morphological relationships while emphasizing prognostically relevant tissue patterns.

3. Through extensive experiments on the multi-institutional Alliance/CALGB 89803 clinical cohort, we demonstrate that MorphDistill consistently outperforms ten state-of-the-art pathology foundation model baselines and five established MIL aggregation approaches for CRC survival prediction. Furthermore, we externally validate MorphDistill on the independent TCGA colorectal cancer cohorts (COAD and READ, n=562), where it achieves a superior C-index of 0.6151±0.07, outperforming multiple MIL baselines including ABMIL (0.603) and RRT-MIL (0.599), confirming its generalizability beyond the clinical trial setting. The learned representation exhibits strong generalization across treatment regimens, demographic subgroups, and tumor locations, highlighting the effectiveness of multi-teacher knowledge distillation for learning task-specific representations in computational pathology.

## 2. Methodology

We propose MorphDistill, a two-stage framework designed to synthesize complementary morphological knowledge from multiple pathology foundation models into a unified CRC-specific representation for slide-level survival prediction. An overview of the proposed framework is illustrated in Fig. 1. In Stage I, a patch-level student encoder is trained using multi-teacher relational distillation together with supervised contrastive regularization on large-scale CRC tissue datasets [34-36]. This stage enables the model to learn a CRC-specific morphological representation by distilling and unifying knowledge from multiple pretrained pathology foundation models, each capturing complementary aspects of tissue morphology. The resulting student encoder is referred to as the MorphDistill encoder, and we use the terms interchangeably. In Stage II, the trained encoder is frozen and applied to tessellated WSIs to extract patch-level embeddings, which are subsequently aggregated using ABMIL to produce slide-level representations for survival prediction [37]. Survival at a predefined clinical time horizon is formulated as a binary classification task and optimized using binary cross-entropy [38-41].

### 2.1. Stage I: Multi-Teacher Relational Distillation

**Problem Formulation:** The fundamental challenge addressed by MorphDistill in the context of colorectal cancer (CRC) survival prediction is that existing histopathology foundation models are typically pretrained using different datasets, architectures, and training objectives, resulting in heterogeneous feature representations with varying embedding dimensionalities and representational geometries. Although these models capture diverse morphological patterns, no single foundation model is optimized for CRC-specific prognostic features. Rather than selecting a single encoder, our goal is to distill and unify the complementary morphological knowledge embedded across multiple foundation models to learn a CRC-specific representation tailored for survival prediction. Let $\{T^{(k)}\}_{k=1}^{K}$ denote $K$ pretrained and frozen teacher pathology foundation encoders. For an input patch $x_i$, teacher $k$ produces an embedding $\mathbf{u}_\mathbf{i}^{(\boldsymbol{k})} = T^{(k)}(x_i) \in \mathbb{R}^{d_k}$, where the dimensionality $d_k$ may vary across teacher networks. Given a mini-batch $\mathcal{B} = \{(x_i, y_i)\}_{i=1}^{N}$, where $x_i \in \mathbb{R}^{224\times224\times3}$denotes an input image patch and $y_i \in \{1, \ldots, C\}$ represents the corresponding morphological tissue class label, the student encoder $S_\theta$ (MorphDistill) maps each patch to $z_i = S_\theta(x_i) \in \mathbb{R}^d$, with a fixed embedding dimension $d = 768$. The goal of Stage I is to pretrain $S_\theta$ the student encoder (MorphDistill) such that the relational structure induced by each teacher's heterogeneous embeddings, thereby distilling their collective morphological knowledge into a unified representation space.

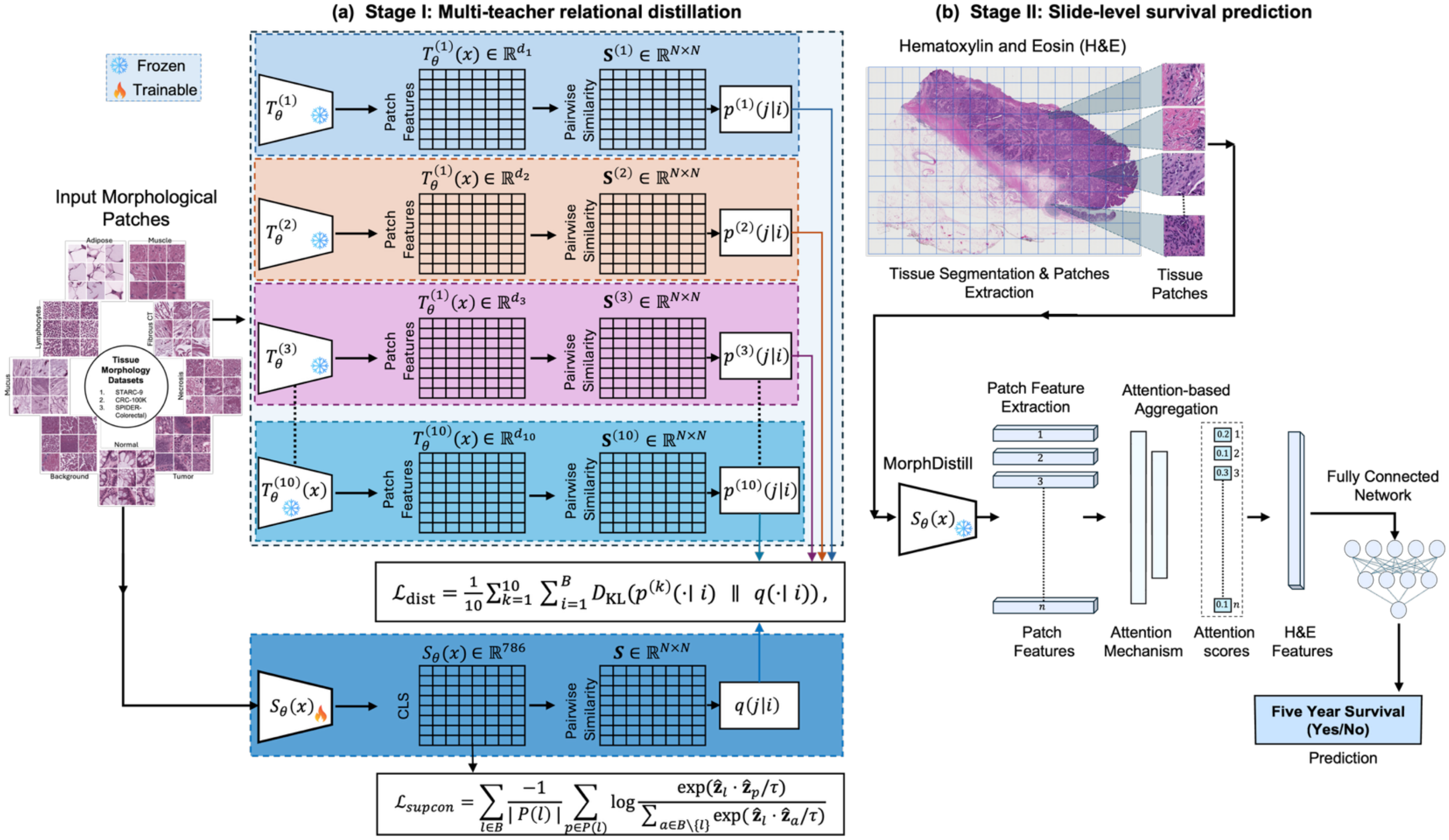

**Figure 1:** Schematic overview of the MorphDistill framework for colorectal cancer survival prediction from H&E whole-slide images (A) Stage I: Multi-teacher relational distillation. A student encoder (MorphDistill) learns CRC-specific morphological representations by distilling knowledge from ten pretrained pathology foundation models. Dimension-agnostic similarity alignment preserves inter-sample relational structures from each teacher without requiring feature projection, while supervised contrastive regularization grounds the representation in colorectal tissue morphology. (B) Stage II: Slide-level survival prediction. The frozen MorphDistill encoder extracts patch embeddings from WSIs, which are aggregated using attention-based multiple instance learning (ABMIL) to generate slide-level representations for five-year survival classification (best viewed in color).

**Dimension-Agnostic Relational Alignment:** To enable knowledge transfer across heterogeneous teacher encoders with varying embedding dimensionalities, MorphDistill performs alignment at the level of batch-wise relational structure level, circumventing the need for explicit feature-space matching. This design choice is critical: by operating on similarities rather than features, we preserve each teacher's original representational geometry without introducing projection bias. First, all embeddings are $\ell_2$-normalized to standardize their scale before computing pairwise similarities. Formally, the normalized teacher and student embeddings are given by:

$$\hat{\mathbf{u}}_i^{(k)} = \frac{\mathbf{u}_i^{(k)}}{\parallel \mathbf{u}_i^{(k)} \parallel_2}, \qquad \hat{\mathbf{z}}_i = \frac{\mathbf{z}_i}{\parallel \mathbf{z}_i \parallel_2}. \tag{1}$$

For each teacher $k$, a pairwise similarity matrix $\mathbf{S}^{(k)} \in \mathbb{R}^{N\times N}$ is constructed, where each entry $[\mathbf{S}^{(k)}]_{ij} = (\hat{\mathbf{u}}_i^{(k)})^\top \hat{\mathbf{u}}_j^{(k)}$ captures the cosine similarity between normalized embeddings of patches $x_i$ and $x_j$ within a batch, respectively. The corresponding student similarity matrix $\mathbf{S} \in \mathbb{R}^{N\times N}$ is defined as $[\mathbf{S}]_{ij} = \hat{\mathbf{z}}_i^\top \hat{\mathbf{z}}_j$. Since these similarity matrices depend only on batch size $B$ rather than individual embedding dimensionalities $d_k$, this formulation ensures the relational alignment remains entirely dimension-agnostic. By decoupling the distillation process from the specific feature dimensions of each teacher encoder, MorphDistill facilitates seamless transfer of knowledge across an ensemble of teachers with disparate architectural designs and training protocols while preserving their original representational geometries.

**Relational Knowledge Distillation:** To transfer the relational structure from each teacher to the student, we convert the pairwise similarities into probability distributions that capture the relative proximity of samples within a batch. This probabilistic formulation emphasizes the neighborhood structure induced by each teacher—which samples are considered similar—rather than absolute similarity values, making the distillation more robust to scale differences across teachers. For each anchor sample $i$, teacher $k$ induces a conditional distribution over all other samples $j \neq i$ by applying a softmax function to the corresponding similarity scores:

$$p^{(k)}(j \mid i) = \frac{\exp([S^{(k)}]_{ij}/\tau)}{\sum_{m \neq i} \exp([S^{(k)}]_{im}/\tau)}, \qquad j \neq i, \tag{2}$$

where $\tau > 0$ is a temperature parameter that controls the sharpness of the resulting distribution. Higher temperatures produce softer distributions, emphasizing broader relational patterns, while lower temperatures sharpen the focus on the nearest neighbors. Similarly, the student conditional distribution using its similarity matrix **S** is defined as follows:

$$q(j \mid i) = \frac{\exp([S]_{ij}/\tau)}{\sum_{m \neq i} \exp([S]_{im}/\tau)}, \tag{3}$$

The relational distillation objective then minimizes the discrepancy between the student's distribution and that of each teacher, encouraging the student to preserve the neighborhood structure induced by each teacher's embeddings—effectively distilling multiple complementary views of morphological similarity into a single representation. This is achieved by minimizing the Kullback–Leibler divergence between the two distributions, aggregated across all anchors and all teachers:

$$\mathcal{L}_{\text{dist}} = \frac{1}{K} \sum_{k=1}^{K} \sum_{i=1}^{B} D_{\text{KL}}(p^{(k)}(\cdot \mid i) \;\; \| \;\; q(\cdot \mid i)), \tag{4}$$

By minimizing this objective, the student embedding space is regularized to reflect the relational geometry encoded by each teacher, effectively distilling complementary notions of morphological similarity into a unified representation. Intuitively, this mechanism encourages the student to reproduce the consensus relationships captured by the teacher models. If multiple teachers agree that two patches are morphologically similar, the student learns to place them close together in its embedding space. Conversely, if a particular teacher captures a subtle morphological pattern—such as a CRC-specific tissue feature—the student still experiences the "pull" of that teacher's similarity distribution and preserves that relational structure in its own representation. As a result, the student encoder integrates complementary morphological signals from multiple foundation models into a single unified embedding space.

**Supervised Contrastive Regularization:** Relational distillation alone remains domain-agnostic. To encourage the student embedding space to capture CRC-specific morphological patterns, we incorporate supervised contrastive learning using labeled histopathology patches, enabling the model to focus on morphological heterogeneity relevant to downstream survival prediction, such as tumor epithelium versus stroma and high-grade versus low-grade dysplasia. Specifically, we utilize a CRC-specific collection of labeled patch-level datasets, including CRC-100K[34], STARC-9[35], and SPIDER[36], which together span 18 colorectal tissue morphological classes. For a mini batch of labeled patches, the supervised contrastive loss $\mathcal{L}_{supcon}$ is defined as:

$$\mathcal{L}_{supcon} = \sum_{l \in B} \frac{-1}{|P(l)|} \sum_{p \in P(l)} \log \frac{\exp(\hat{\mathbf{z}}_l \cdot \hat{\mathbf{z}}_p / \tau)}{\sum_{a \in B \setminus \{l\}} \exp(\hat{\mathbf{z}}_l \cdot \hat{\mathbf{z}}_a / \tau)}, \quad (5)$$

where $l$ denotes the anchor patch, $P(l)$ represents the set of indices of all patches within the mini-batch sharing the same morphological class label as the anchor, and $\tau$ is a temperature parameter controlling the sharpness of the similarity distribution. This objective pulls embeddings of patches from the same tissue class together while pushing apart embeddings from different classes, grounding the representation in CRC-relevant morphological distinctions. The overall Stage I training objective jointly optimizes relational distillation and supervised contrastive regularization:

$$\mathcal{L}_{total} = \lambda \mathcal{L}_{supcon} + (1 - \lambda) \mathcal{L}_{dist}, \quad (6)$$

where $\lambda$ controls the trade-off between preserving the general representational structure distilled from the teacher foundation models and incorporating CRC-specific morphological supervision. With $\lambda = 0.75$, we assign greater weight to the supervised contrastive objective, emphasizing CRC-relevant features while still benefiting from the broad morphological knowledge encoded in the teacher models. Optimizing the objective in Eq. (6) during Stage I yields a CRC-specific encoder (MorphDistill), which is subsequently used as a fixed feature extractor for whole-slide images (WSIs) in Stage II to enable five-year survival prediction. Figures S1-S26 illustrates examples of patch-level morphological classifications generated by the trained MorphDistill model.

## 2.2. Stage II: Slide-Level Survival Prediction

Once the student encoder $S_\theta$ is trained via multi-teacher relational distillation and supervised contrastive regularization, it is frozen and used as a fixed feature extractor for WSIs, providing CRC-specific morphological representations that synthesize knowledge from ten pathology foundation models for downstream survival prediction. In this stage, we utilize the multi-institutional Alliance/CALGB 89803 cohort, a collection of WSIs from stage III colorectal cancer patients with associated five-year survival outcomes, to train a slide-level survival prediction model. The encoder transforms each WSI into a bag of patch-level embeddings, which are then aggregated using ABMIL to produce a slide-level representation for binary survival classification [37].

**Patch embedding extraction:** Each WSI is tessellated into non-overlapping tissue patches at $20 \times$ magnification, with dimensions $224 \times 224 \times 3$. Patches containing less than $25\%$ of the tissue area are discarded using TRIDENT [42] to ensure that only informative regions are retained for downstream five-year survival prediction. For a given WSI containing $N$ tissue patches $\{x_n\}_{n=1}^{N}$, the frozen MorphDistill encoder $S_\theta$ extracts a corresponding set of patch-level embeddings:

$$h_n = S_\theta(x_n) \in \mathbb{R}^{768}, \qquad \mathcal{H} = \{h_n\}_{n=1}^{N} \quad (7)$$

The resulting set $\mathcal{H}$ constitutes a bag-of-instances representation, where each instance corresponds to a histologically meaningful patch and its embedding captures both generic and CRC-specific morphological features synthesized from multiple foundation models during Stage I.

**Attention-based multiple instance learning:** To aggregate the variable-sized bag of patch embeddings into a fixed-dimensional slide-level representation, we employ attention-based multiple instance learning [37]. This mechanism learns to assign importance weights to individual patches based on their prognostic relevance, enabling the model to focus on regions most indicative of patient outcome—such as invasive tumor fronts, desmoplastic stroma, or tumor-infiltrating lymphocytes. For each patch embedding $h_n$, an attention score is computed as:

$$a_n = \frac{\exp(\mathbf{w}^\top \tanh(\mathbf{V}\mathrm{h}_\mathrm{n}))}{\sum_{m=1}^{N} \exp(\mathbf{w}^\top \tanh(\mathbf{V}\mathrm{h}_\mathrm{m}))}, \quad (8)$$

where $\mathbf{V} \in \mathbb{R}^{L\times 768}$ and $\mathbf{w} \in \mathbb{R}^{L}$ are learnable parameters, and $L$ denotes the dimension of the attention hidden layer. These attention weights are normalized across all patches in the slide, ensuring that $\sum_{n=1}^{N} a_n = 1$. The final slide-level representation $\mathbf{g} \in \mathbb{R}^{768}$ is then computed as the attention-weighted sum of all patch embeddings:

$$\mathbf{g} = \sum_{n=1}^{N} a_n \, \mathbf{h}_n, \tag{9}$$

**Binary survival prediction objective:** Survival is formulated as a binary classification task at a predefined clinical time horizon of five years. For each slide, let $s \in \{0, 1\}$ denote the survival label, where $s = 1$ indicates that the patient died within five years of resection and$s = 0$ indicates survival beyond five years. The slide-level representation $\mathbf{g}$ is passed through a linear prediction head to produce a logit:

$$l = \mathbf{w}_c^{\top} \mathbf{g} + b, \qquad \hat{s} = \sigma(l), \tag{10}$$

where $\sigma(.)$ is the sigmoid activation function, $\mathbf{w}_c \in \mathbb{R}^{768}$ and $b \in \mathbb{R}$ are learnable parameters. The predicted probability $\hat{s}$ quantifies the likelihood of the patient belonging to the deceased group. The model is optimized by minimizing the binary cross-entropy loss across all training slides:

$$\mathcal{L}_{\mathrm{BCE}} = -\sum_{i=1}^{M} [s_i \log \, \hat{s}_i + (1 - s_i) \log \, (1 - \hat{s}_i)], \tag{11}$$

where $M$ denotes the number of slides in a mini batch (or the entire training set). Through this objective, the slide-level aggregation network learns to combine patch-level morphological features—now enriched with knowledge from ten foundation models—into a holistic representation predictive of five-year survival.

## 3. Experimental Setup

### 3.1. Datasets

We utilize two categories of datasets corresponding to the two stages of the MorphDistill framework: (i) large-scale patch-level datasets used for representation learning during Stage I, and (ii) a clinically curated survival cohort used for slide-level survival prediction in Stage II.

#### 3.1.1. Encoder Pre-Training Dataset

To train the Stage I morphology encoder, we construct a unified CRC morphological training dataset by combining three curated histopathology patch datasets: CRC-100K[34], STARC-9[35], and SPIDER[36], which together provide comprehensive coverage of colorectal tissue morphologies spanning 18 distinct tissue classes relevant to CRC pathology. All patches are extracted from H&E-stained whole-slide images and include expert-annotated tissue labels. Patient-disjoint splits are maintained throughout to prevent data leakage. Table S2 summarizes the dataset-wise distribution of morphological classes across CRC-100K, STARC-9, and SPIDER-Colorectal, including the number of training and validation patches contributed by each dataset. Table S3 presents the distribution of the unified set of 18 morphological classes used to train the MorphDistill encoder during Stage I.

**CRC-100K:** The CRC-100K dataset consists of two subsets: NCT-CRC-HE-100K and CRC-VAL-HE-7K [34]. The training subset contains 100,000 non-overlapping 224×224 patches extracted from 86 WSIs, while the validation subset contains 7,180 patches from 50 independent patients. The dataset includes nine tissue classes: adipose tissue (ADI), background (BACK), debris (DEB), lymphocytes (LYM), mucus (MUC), smooth muscle (MUS), normal colon mucosa (NORM), cancer-associated stroma (STR), and colorectal adenocarcinoma epithelium (TUM).

**STARC-9:** The STARC-9 (STAnford coloRectal Cancer) dataset is a large-scale histopathology benchmark designed for multi-class tissue classification in colorectal cancer [35]. A distinguishing feature of STARC-9 is its emphasis on morphological diversity and balanced class representation, with 70,000 tiles per class extracted from 200 CRC patients, which addresses common limitations of earlier CRC patch datasets that often suffer from sampling bias or limited tissue variability. Tiles are generated at 40× magnification (0.25 μm/pixel) with a spatial resolution of 256×256 pixels and subsequently verified by expert gastrointestinal pathologists. The final dataset includes nine tissue categories: adipose tissue (ADI), lymphocytes (LYM), smooth muscle (MUS), fibrous connective tissue (FCT), mucus (MUC), necrotic or cellular debris (NCS), blood (BLD), tumor epithelium (TUM), and normal tissue (NOR).

**SPIDER-Colorectal:** The SPIDER-Colorectal dataset contains 77,182 labeled patches (224×224) extracted from 1,719 WSIs at 20× magnification [36]. It includes 13 tissue and lesion classes spanning neoplastic and non-neoplastic patterns: adenocarcinoma (high-grade and low-grade), adenoma (high-grade and low-grade), hyperplastic polyp, sessile serrated lesion, inflammation, necrosis, mucus, muscle, fat, healthy stroma, and vessels. These annotations enable the model to learn detailed morphological features spanning the tumor microenvironment and epithelial lesions, which are essential for capturing prognostically relevant patterns in CRC histopathology.

### 3.1.2. Survival Alliance Cohort

The survival cohort is derived from the Alliance/CALGB 89803 randomized phase III trial of adjuvant therapy for stage III colorectal cancer. This rigorously curated clinical trial cohort provides high-quality outcome data with consistent treatment protocols and long-term follow-up, enabling robust evaluation of prognostic models. All WSIs underwent pathology review and quality control to ensure adequate tissue content and scan quality (e.g., tissue coverage, color fidelity, focus, and absence of scanning artifacts). Cases were excluded for the absence of tumor tissue, lymph-node-only specimens, non–disease-related deaths within five years, and loss to follow-up before five years. In addition, mucinous adenocarcinomas were excluded because their histologic architecture precludes reliable assessment of tumor budding. According to the World Health Organization (WHO) classification, mucinous carcinoma is defined by abundant extracellular mucin comprising ≥50% of the tumor volume, in which neoplastic epithelial cells are typically arranged as clusters or glandular structures floating within mucin pools. Accordingly, 50 cases of mucinous adenocarcinoma (≥50% mucin) were excluded from the analysis. After filtering, the final cohort comprised 424 patients with 431 WSIs, stratified into deceased (n=103) and surviving (n=321) groups based on five-year follow-up. For model development and evaluation, patient-disjoint five-fold splits were constructed using a stratified procedure that clusters patients by key covariates (age, BMI, and income) to preserve subgroup balance across folds. Slides were tessellated into 224×224 patches at 20× magnification, and patches with $< 25\%$ tissue were discarded. A detailed summary of the demographic and clinical characteristics of the study cohort is provided in Table S1.

### 3.1.3. **External Validation Cohort: TCGA READ and COAD**

To assess the generalizability of the MorphDistill framework beyond the Alliance/CALGB 89803 clinical trial cohort, we performed an independent validation on The Cancer Genome Atlas (TCGA) colon adenocarcinoma (COAD) and rectal adenocarcinoma (READ) cohorts. A total of 562 patients with available whole-slide images and survival data were included, comprising 423 COAD and 139 READ cases. Following the same preprocessing pipeline described in Section 3.1.2, whole-slide images were tessellated into $224 \times 224$ patches at 40× magnification, and patches with $< 25\%$ tissue content were discarded. Patients were stratified into deceased ($n = 117$) and surviving ($n = 445$) groups based on five-year follow-up. In contrast to the Alliance cohort, mucinous adenocarcinoma cases ($n = 69,\ 12.3\%$ of the cohort) were retained to assess model performance on a more diverse and heterogeneous dataset. This independent cohort serves to evaluate the robustness of the MorphDistill encoder against real-world data characterized by diverse scanning protocols, patient demographics, and tumor stages.

## 3.2. Baselines

To comprehensively evaluate the effectiveness of MorphDistill, we perform two complementary levels of comparison: (i) encoder-level benchmarking against state-of-the-art pathology foundation models, and (ii) aggregation-level benchmarking against established MIL frameworks. This evaluation strategy enables us to separately assess the contribution of the learned feature representation and the downstream aggregation model for the CRC survial prediction.

#### 3.2.1. Encoder-Level Baselines:

To evaluate the representation quality of the proposed MorphDistill encoder, we compare it against ten state-of-the-art pathology foundation models for five-year CRC survival prediction, including CONCH v1.5[11], CTransPath[12], Gigapath[13], Hoptimus0[14], Kaiko-ViTB8[15], Lunit-ViTS8[16], Phikon v2[17], UNI v1[18], UNI v2[19], and Virchow2[20]. To ensure a fair and controlled comparison, all encoders are used as frozen feature extractors, with identical preprocessing, patch tessellation, and feature extraction pipelines. The extracted patch embeddings are aggregated using the same ABMIL framework, and only the aggregation network and classification head are trained during slide-level survival prediction. This unified experimental setup isolates the effect of the learned feature representation, ensuring that differences in performance primarily reflect encoder representation quality rather than variations in downstream modeling choices.

#### 3.2.2. MIL Aggregation Baselines:

To evaluate whether the performance gains of MorphDistill arise from the feature representation itself rather than a specific aggregation architecture, we further compare against five established MIL aggregation methods, including ABMIL [37], CLAM [10], TransMIL [43], PATNTER [44], DSMIL [45], Nakanishi et al. [46], and RRT-MIL [47]. For this comparison, all aggregation models are evaluated using identical preprocessing, patch extraction, and patient-disjoint cross-validation splits. For baseline MIL approaches, we use UNI v2-derived patch embeddings as input, as UNI v2 represents the strongest-performing foundation model in our encoder-level benchmarking and is widely used in recent computational pathology studie and ensuring consistency with prior literature [48]. This experimental design allows us to isolate the effect of the aggregation strategy while keeping the feature representation fixed, enabling a rigorous comparison between MorphDistill embeddings and conventional foundation model features across multiple MIL architectures.

### 3.3. Evaluation Protocol

All experiments were conducted using patient-disjoint five-fold cross-validation. Data splits were constructed to preserve clinical and demographic subgroup balance across folds to minimize potential sampling bias. Within each cross-validation iteration, 85% of the training folds were used for model training, the remaining 15% were used for validation, and the held-out fold was used for testing. Model performance was evaluated using Area Under the Receiver Operating Characteristic Curve (AUC), balanced accuracy, sensitivity, and specificity. Performance metrics are reported as mean ± standard deviation across the five cross-validation folds. Sensitivity and specificity were computed based on the number of true positives (TP), true negatives (TN), false positives (FP), and false negatives (FN). Sensitivity (true positive rate) measures the proportion of positive samples correctly identified by the model and is defined as:

$$\text{Sensitivity (Sen)} = \frac{TP}{TP + FN}, \tag{12}$$

Specificity (true negative rate) measures the proportion of negative samples correctly identified and is defined as:

$$\text{Specificity (Spec)} = \frac{TN}{TN + FP}, \tag{13}$$

Balanced accuracy is defined as the average of sensitivity and specificity, ensuring that performance is evaluated independently of class imbalance:

$$\text{Balanced Accuracy (Bal-Acc)} = \frac{1}{2}\left(\frac{TP}{TP+FN} + \frac{TN}{TN+FP}\right). \tag{14}$$

### 3.4. Implementation Details

#### 3.4.1. Stage I: Encoder Training:

The student encoder $S_\theta$ is based on a Vision Transformer (ViT-S) architecture with a patch size of 16 and an embedding dimension of 768 [49]. For multi-teacher distillation, we select ten pathology foundation models as teachers models: CONCHv1.5[11], CTransPath[12], Gigapath[13], Hoptimus0[14], Kaiko-ViTB8[15], Lunit-ViTS8[16], Phikon v2[17], UNI v1[18], UNI v2[19], and Virchow2[20], representing diverse pretraining datasets, architectural designs, and self-supervised learning objectives. All teacher models were frozen during Stage I training. The student encoder is trained for 50 epochs using the AdamW optimizer with a learning rate of $1 \times 10^{-3}$ and weight decay of $1 \times 10^{-4}$ [50,51]. A cosine annealing learning rate scheduler is employed throughout training [52]. The batch size was set to 256. The temperature parameter $\tau$ in Equations (2), (3), and (5) was set to 0.1 for both relational distillation and supervised contrastive losses. The balancing hyperparameter $\lambda$ in Equation (6) was set to 0.75, assigning greater weight to the supervised contrastive objective to emphasize CRC-relevant morphological features while preserving general representational structure. All training patches were resized to 224×224 pixels and normalized using dataset-specific statistics. To improve generalization and mitigate overfitting, we applied standard data augmentations, including random horizontal and vertical flips, random rotations (up to 90 degrees), and color jittering (brightness, contrast, saturation, and hue) [53]. Time-to-event metrics (concordance index and hazard ratio) were computed from model probability scores using Cox proportional hazards regression. All experiments were conducted on NVIDIA A100 GPUs using PyTorch 2.0.

#### 3.4.2. Stage II: Survival Prediction:

For Stage II training, the pretrained MorphDistill encoder was frozen and used as a fixed feature extractor. The ABMIL aggregator comprised a gated attention network with a hidden dimension of 512. The slide-level survival prediction model is trained for 100 epochs using the Adam optimizer with a learning rate of $2 \times 10^{-4}$ and a batch size of 1 (each batch containing all patches from a single WSI) [54]. Early stopping was applied based on validation AUC with a patience of 10 epochs. L1 regularization with a coefficient of $5 \times 10^{-4}$ was applied to the attention network to promote sparsity and focus on the most prognostically relevant patches. All baseline methods were implemented using publicly available code repositories with their default hyperparameters, except where adjustments were necessary to ensure fair comparison (e.g., matching input feature dimensionality).

## 4. Results

### 4.1. Representational Diversity Across Pathology Foundation Models

To assess the discriminative capacity of feature representations learned by different pathology foundation models, we visualize patch embeddings from the validation set of the CRC morphological dataset using t-SNE (Fig. 2). The projections reveal that most models capture meaningful morphological structure, with patches belonging to similar tissue types forming localized clusters in the embedding space. However, the compactness of clusters and the degree of inter-class separation vary across encoders, indicating differences in how effectively each model captures histopathological patterns. Notably, UNI v2 and MorphDistill exhibit more compact clusters and clearer boundaries between tissue classes, suggesting stronger morphological discrimination compared with other foundation models. Importantly, while MorphDistill produces comparably well-localized clusters in the embedding space, these clusters are additionally grounded in CRC-specific morphological patterns relevant for downstream survival prognostication. This reflects the effect of the supervised contrastive regularization introduced during Stage I training, which anchors the distilled representation in CRC-specific tissue morphology while preserving relational knowledge transferred from multiple teacher models. As illustrated in Fig. 2, the resulting feature spaces differ substantially across encoders, reflecting variations in training datasets, architectural designs, and pretraining objectives. These observations highlight the complementary representational biases across pathology foundation models and motivate the design of MorphDistill. By distilling relational knowledge from multiple encoders into a unified representation, MorphDistill integrates complementary morphological information and produces a feature space that better captures the diversity of histopathological patterns relevant to colorectal cancer.

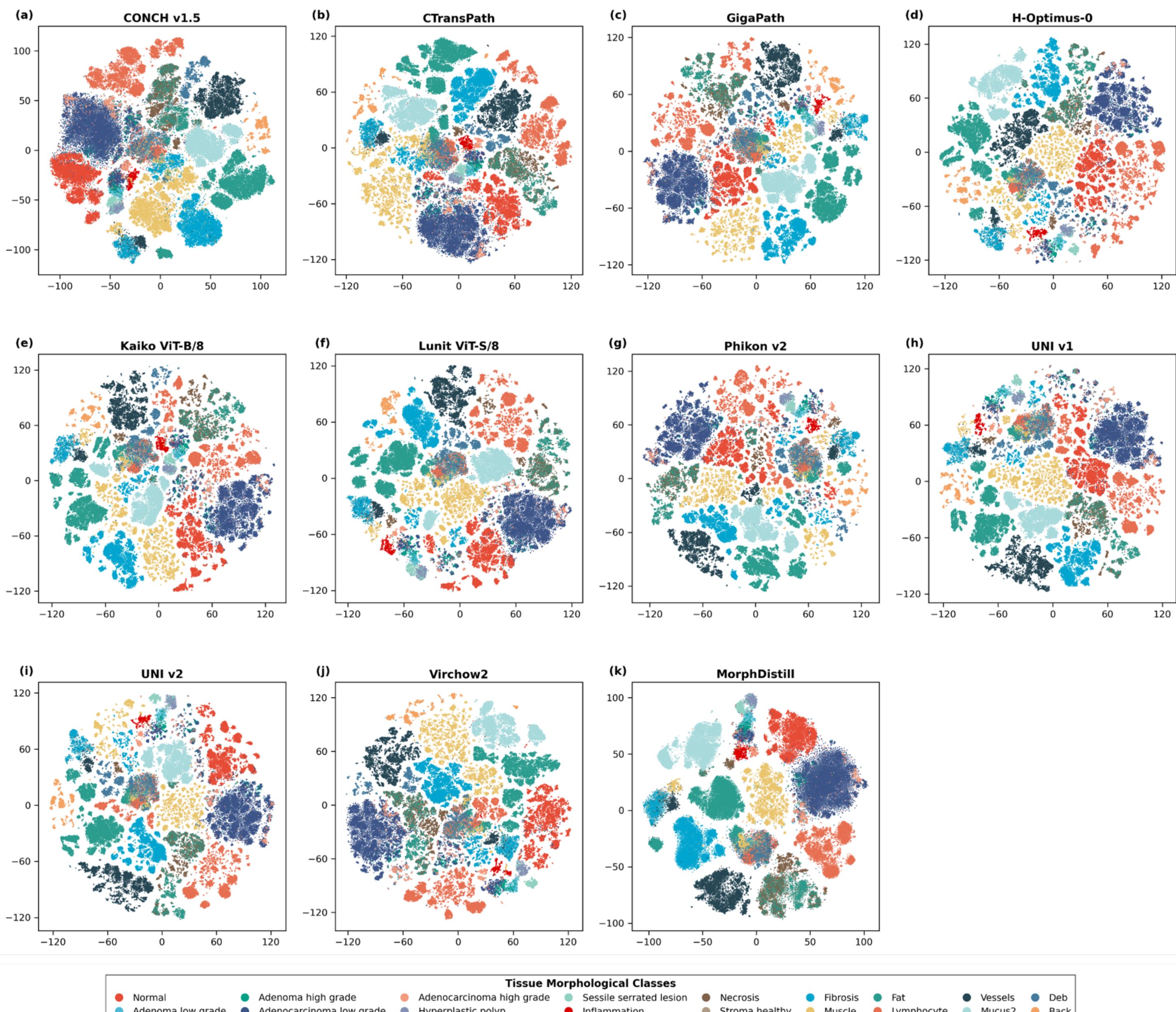

**Figure 2:** t-SNE visualization reveals heterogeneous representational geometries across ten pathology foundation models. Each point represents a tissue patch from the Alliance cohort, colored by morphological class. The degree of cluster compactness and inter-class separation varies substantially across encoders, reflecting differences in their training datasets and architectures. Models such as UNI v2 and MorphDistill exhibit more cohesive, well-separated clusters, indicating stronger discriminative capacity. This observed diversity confirms that different foundation models capture complementary aspects of tissue morphology, motivating our multi-teacher distillation approach (best viewed in color).

### 4.2. Five-Year Survival Prediction Performance Against State-of-the-Art Pathology Foundation Models

The central hypothesis of MorphDistill is that unifying heterogeneous knowledge from multiple expert models yields a specialized representation that surpasses any single foundation model for CRC prognostication. To test this, we benchmarked our distilled encoder against ten state-of-the-art pathology foundation models using an identical ABMIL aggregation framework—ensuring that differences in performance reflect representation quality rather than downstream modeling choices. As shown in Table 1, MorphDistill achieves the highest predictive performance among all evaluated encoders, with a mean AUC of $0.68 \pm 0.08$, outperforming the strongest baseline (UNI v2, AUC 0.63) by approximately 8%. MorphDistill also attains the highest balanced accuracy (64.11%) and sensitivity (60.24%), indicating improved identification of patients at high risk of mortality.

**Table 1.** Benchmarking MorphDistill against state-of-the-art pathology foundation models for five-year survival prediction. All encoders are frozen, and their extracted patch features are aggregated using a shared ABMIL framework to ensure a fair comparison of representation quality. MorphDistill achieves the highest AUC, balanced accuracy, and sensitivity among all evaluated models on the Alliance cohort in five-fold cross-validation. The best-performing metric in each column is highlighted in bold.

| Encoder | AUC | Bal-Acc (%) | Sensitivity (%) | Specificity (%) |
|---|---|---|---|---|
| CONCH v1.5[11] | 0.51 ± 0.07 | 50.81 ± 3.83 | 41.80 ± 15.76 | 59.81 ± 16.66 |
| CTransPath[12] | 0.58 ± 0.02 | 50.63 ± 1.97 | 21.90 ± 39.22 | **79.36 ± 37.38** |
| GigaPath[13] | 0.60 ± 0.04 | 56.37 ± 2.20 | 35.86 ± 6.10 | 76.89 ± 4.69 |
| Hoptimus0[14] | 0.61 ± 0.04 | 56.59 ± 4.13 | 36.71 ± 19.22 | 76.46 ± 18.38 |
| Kaiko-ViTB8[15] | 0.59 ± 0.11 | 58.38 ± 8.91 | 46.45 ± 30.02 | 70.31 ± 13.85 |
| Lunit-ViTS8[16] | 0.58 ± 0.04 | 53.22 ± 3.14 | 54.24 ± 15.52 | 52.21 ± 12.69 |
| Phikon v2[17] | 0.58 ± 0.04 | 58.05 ± 5.67 | 59.50 ± 15.38 | 56.61 ± 11.42 |
| UNI v1[18] | 0.62 ± 0.05 | 53.58 ± 2.44 | 38.14 ± 20.97 | 69.02 ± 21.79 |
| UNI v2[19] | 0.63 ± 0.03 | 60.59 ± 3.48 | 44.62 ± 8.81 | 76.53 ± 3.88 |
| Virchow2[20] | 0.58 ± 0.02 | 55.73 ± 3.84 | 52.71 ± 12.97 | 58.75 ± 19.85 |
| MorphDistill (ours) | **0.68 ± 0.08** | **64.11 ± 4.70** | **60.24 ± 5.38** | 66.57 ± 8.81 |

### 4.3. Risk Stratification Analysis Across State-of-the-Art Pathology Foundation Models

Beyond survival prediction, the capacity of each encoder for patient risk stratification was further assessed using Kaplan-Meier analysis. Figure 3 shows the survival curves for patients stratified into high-risk and low-risk groups based on model predictions. As shown, MorphDistill achieves the most consistent separation between the two groups across the five-year follow-up period, indicating improved prognostic discrimination compared with baseline encoders.

Time-to-event metrics further support these findings. MorphDistill achieves a concordance index (C-index) of 0.661 and a hazard ratio (HR) of 2.52 (95% CI: 1.73–3.65) for patients identified as high-risk. In comparison, the strongest baseline encoder (UNI v2) achieves a C-index of 0.633 and an HR of 2.08 (95% CI: 1.43–3.02). C-index and hazard ratio values for all evaluated models are presented in Figure 4.

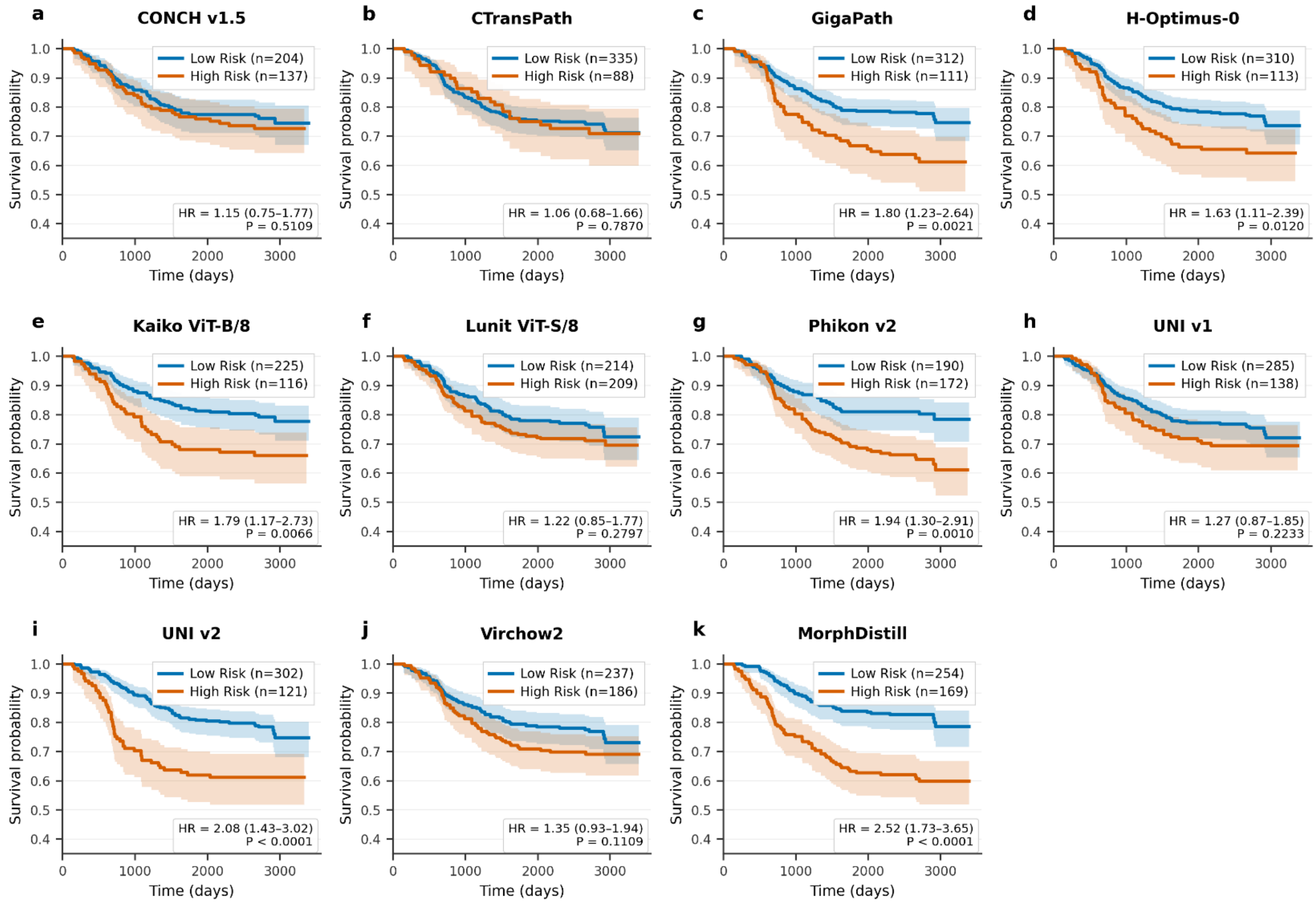

**Figure 3:** Kaplan-Meier survival curves for risk stratification. Patients are stratified into high-risk and low-risk groups based on prediction scores from each model. MorphDistill maintains separation between the curves throughout the five-year follow-up (best viewed in color).

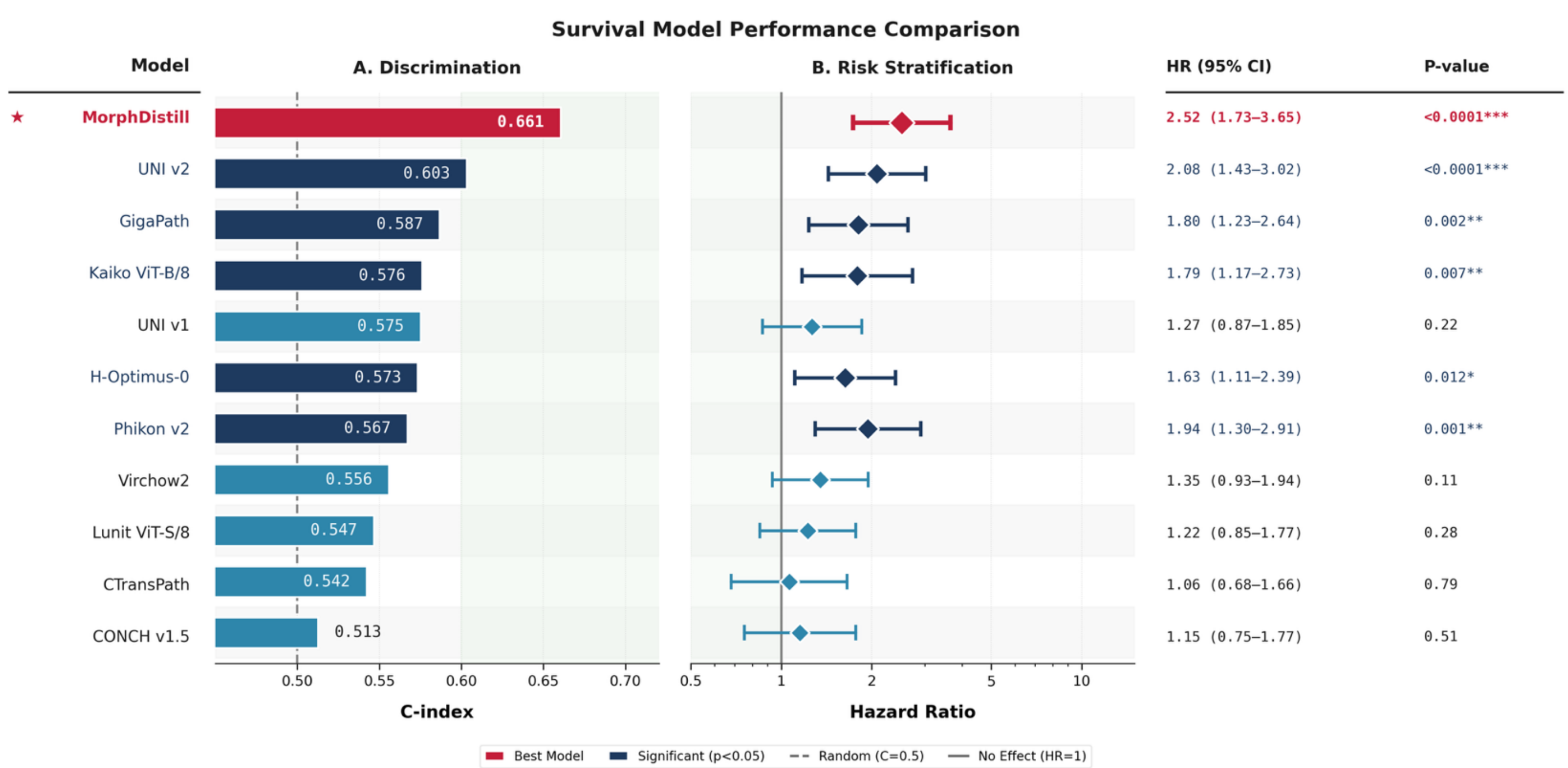

**Figure 4:** Time-to-event metrics for foundation encoders. (A) Concordance index (C-index) and (B) Hazard Ratio with 95% confidence intervals for each encoder. MorphDistill shows a higher C-index and hazard ratio compared to other models (best viewed in color).

### 4.4. Five-Year Survival Prediction Performance Across MIL Aggregation Methods

To determine whether the performance gains of MorphDistill arise from the learned representation rather than a specific aggregation architecture, we evaluated MorphDistill features against those from the top-performing baseline encoder (UNI v2) using five distinct MIL frameworks: ABMIL[37], CLAM[10], TransMIL[43], Nakanishi et al. [46], and RRT-MIL[47]. As summarized in Table 2, MorphDistill embeddings consistently outperform UNI v2 embeddings across all five MIL aggregation methods. For instance, when compared with CLAM, MorphDistill achieves an AUC of 0.68 compared to 0.66 for UNI v2. Performance improvements are particularly evident in sensitivity, where MorphDistill improves the detection of high-risk patients.

**Table 2**. Performance of MIL aggregation methods with UNI v2 and MorphDistill embeddings. Five-year survival prediction metrics (mean ± std) comparing MorphDistill against UNI v2 across five MIL aggregators. MorphDistill features consistently outperform UNI v2 features across all aggregation methods, demonstrating that performance gains stem from the representation itself rather than the aggregation architecture. The best-performing metric in each column is highlighted in bold.

| Model | AUC | Bal-Acc (%) | Sensitivity (%) | Specificity (%) | C-Index |
|---|---|---|---|---|---|
| ABMIL [37] | 0.63 ± 0.06 | 58.84 ± 4.50 | 51.18 ± 9.83 | 65.89 ± 4.50 | 0.607 |
| CLAM [10] | 0.66 ± 0.06 | 55.15 ± 3.37 | 25.61 ± 8.18 | 84.68 ± 8.89 | 0.603 |
| TransMIL [43] | 0.67 ± 0.05 | 52.50 ± 4.64 | 5.25 ± 9.11 | **99.74 ± 0.05** | 0.647 |
| Nakanishi et al. [46] | 0.61 ± 0.06 | 58.71 ± 5.77 | 53.49 ± 11.11 | 63.13 ± 2.75 | 0.604 |
| RRT-MIL [47] | 0.59 ± 0.05 | 56.76 ± 4.46 | 48.58 ± 23.92 | 64.94 ± 22.19 | 0.575 |
| PROGPATH [55] | - | - | - | - | 0.560 |
| MorphDistill | **0.68 ± 0.08** | **64.11 ± 4.70** | **60.24 ± 5.38** | 66.57 ± 8.81 | **0.661** |

### 4.5. Risk Stratification Analysis Across MIL Aggregation Methods

Kaplan–Meier survival analyses further support these findings. Figure 5 shows that models using MorphDistill embeddings consistently produce clearer separation between high and low-risk patient groups compared with models using UNI v2 features. In particular, the MorphDistill + ABMIL combination exhibits the most pronounced survival stratification, with survival curves diverging early and remaining well separated throughout the five-year follow-up period.

Time-to-event evaluation provides additional evidence of the superiority of MorphDistill representations. As shown in Figure 6 (A–B), MorphDistill embeddings yield higher concordance index (C-index) values and hazard ratios across all five MIL aggregation frameworks compared with UNI v2 features.

As shown in Figure 5 presents the Kaplan-Meier survival curves for patients stratified by models trained on MorphDistill and UNI v2 features. Notably, the combination of MorphDistill with ABMIL yields the most pronounced separation between risk groups, with survival curves that diverge early and remain well-separated throughout the five-year follow-up period—indicating that the unified representation captures prognostically relevant information that translates into clinically meaningful risk stratification regardless of the aggregation method. Figure 6 presents the C-index and hazard ratio for models using MorphDistill and UNI v2 features across the five MIL aggregators, further confirming the consistent superiority of MorphDistill embeddings.

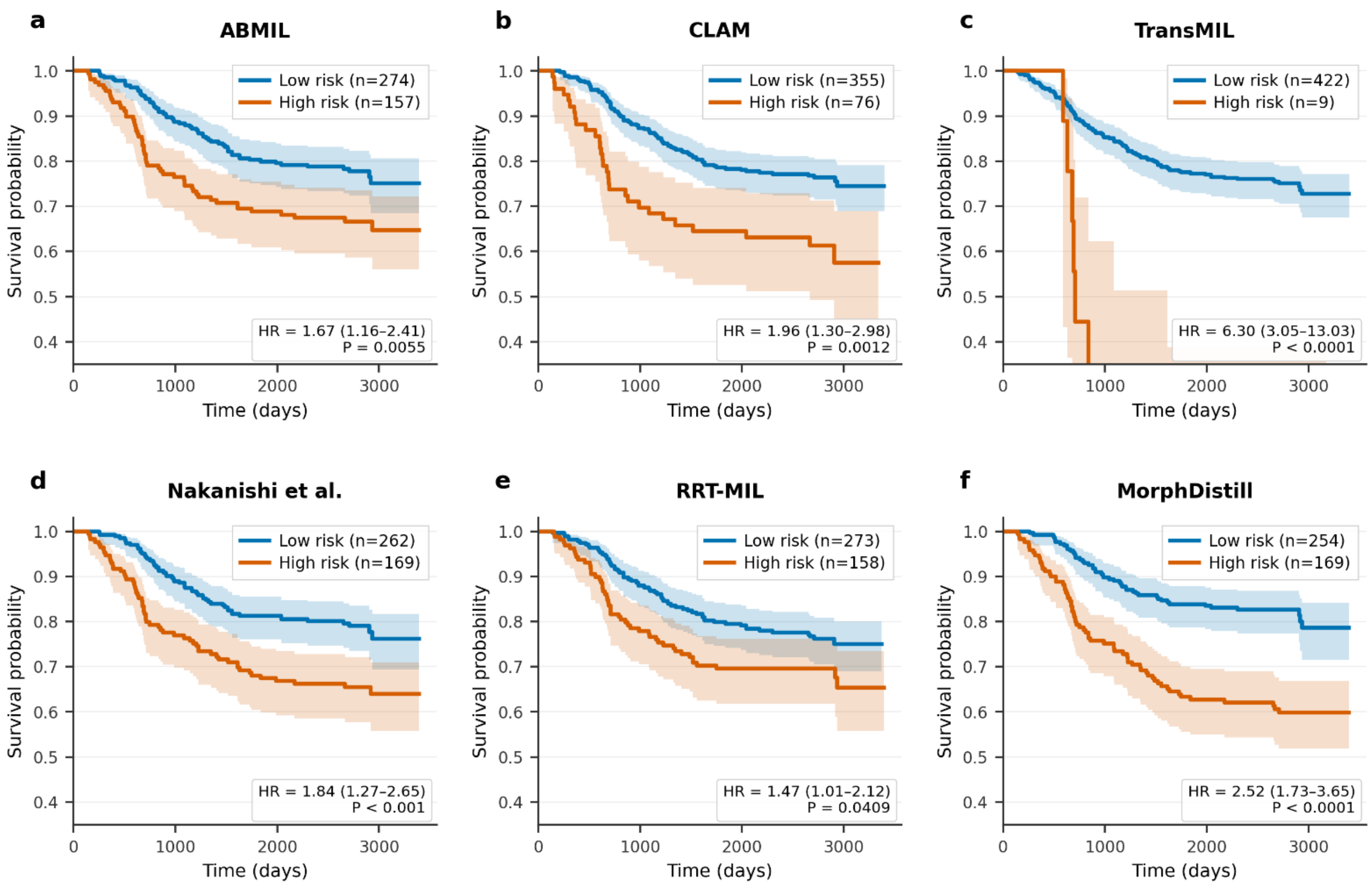

**Figure 5:** Comparison of risk stratification using MorphDistill and UNI v2 features across five MIL frameworks. MorphDistill consistently achieves better separation between high- and low-risk groups, with the most pronounced stratification observed when combined with ABMIL (best viewed in color).

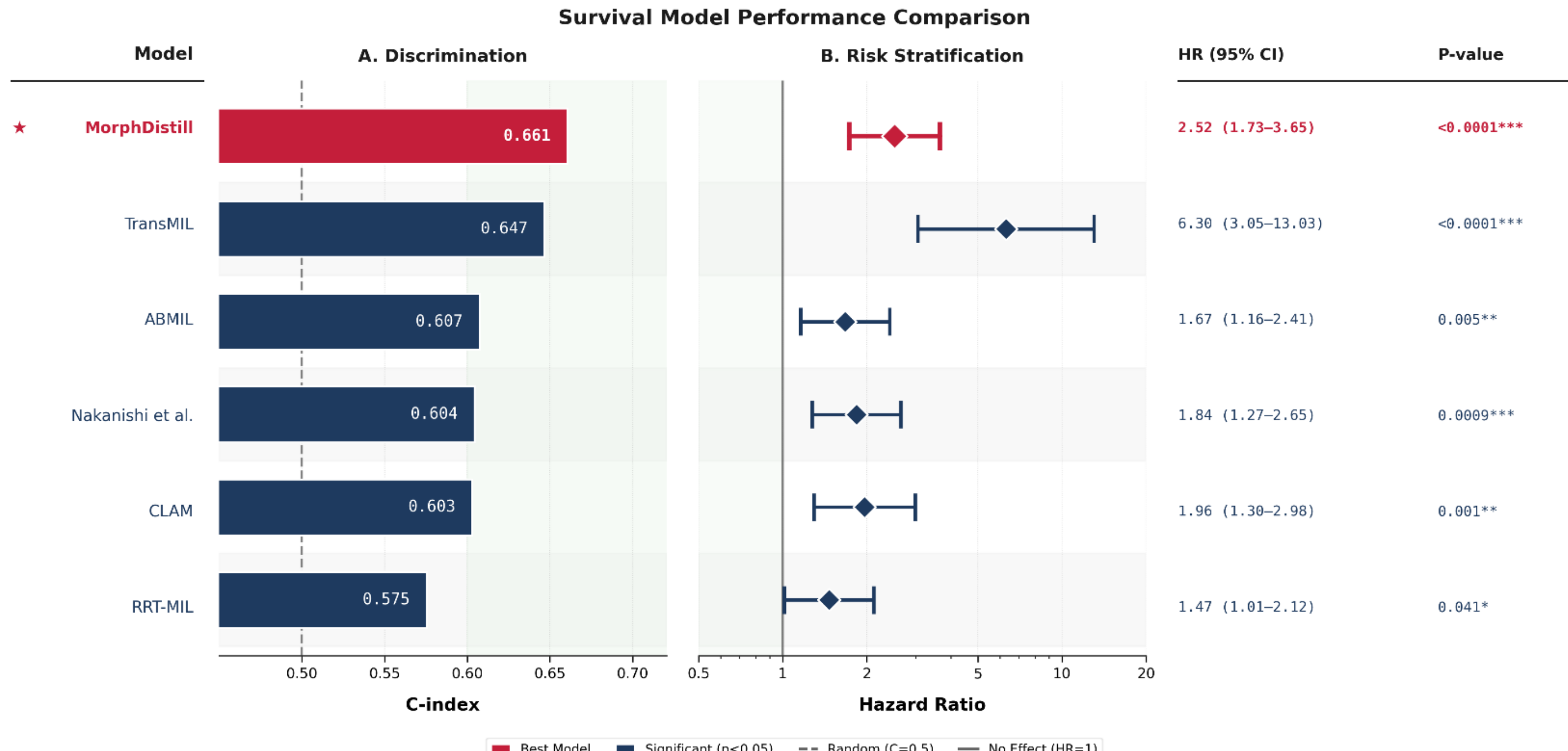

**Figure 6:** (A) C-index and (B) Hazard Ratio (with 95% CI) for five MIL methods using MorphDistill and UNI v2 features. MorphDistill embeddings achieve higher C-index values and hazard ratios across all aggregation methods, confirming that performance improvements are attributable to the unified representation rather than the choice of MIL architecture (best viewed in color).

### 4.6. Stratified Prognostic Performance Across Treatment Regimens, Sex, and Tumor Location

To assess the robustness and generalizability of MorphDistill, we evaluated its performance across clinically relevant subgroups of the Alliance cohort, including treatment regimen, sex, and tumor location. These analyses examine whether the learned representation remains predictive under varying clinical and biological conditions, a critical requirement for clinical deployment.

#### 4.6.1. Treatment-Stratified Analysis:

The cohort includes patients treated with either FL or IFL chemotherapy regimens. As shown in Table 3, MorphDistill achieves an AUC of 0.70 in the FL arm and 0.65 in the IFL arm, indicating stable performance across treatment regimens. In contrast, several baseline methods exhibit larger performance variability between treatment arms. For example, TransMIL achieves an AUC of 0.75 in the FL arm but only 0.62 in the IFL arm, suggesting greater sensitivity to treatment-specific variations. These results indicate that MorphDistill captures intrinsic morphological characteristics of CRC rather than treatment-related artifacts, supporting its potential for broader clinical applicability.

**Table 3**. Five-year survival prediction results of MorphDistill stratified by FL/IFL treatments in the Alliance cohort. MorphDistill maintains stable performance across treatment arms (AUC 0.70 vs. 0.65), demonstrating that the learned representation captures intrinsic morphological characteristics rather than treatment-specific artifacts. The best-performing metric in each column is highlighted in bold.

| Model | Treatment | AUC | Bal-Acc (%) | Sensitivity (%) | Specificity (%) |
|---|---|---|---|---|---|
| ABMIL [37] | FL | 0.65 ± 0.11 | **62.02 ± 13.76** | 51.93 ± 22.92 | 72.12 ± 08.70 |
| | IFL | 0.58 ± 0.07 | 54.13 ± 05.61 | 45.34 ± 08.90 | 62.92 ± 06.89 |
| CLAM [10] | FL | 0.61 ± 0.09 | 50.04 ± 05.96 | 14.12 ± 08.97 | 85.96 ± 08.47 |
| | IFL | 0.63 ± 0.15 | 60.89 ± 06.01 | 36.71 ± 13.97 | 85.07 ± 09.81 |
| TransMIL [43] | FL | **0.75 ± 0.10** | 54.75 ± 08.37 | 10.33 ± 16.13 | **99.16 ± 01.66** |
| | IFL | 0.62 ± 0.09 | 51.81 ± 03.63 | 03.63 ± 07.27 | **100.00 ± 0.00** |
| Nakanishi et al. [46] | FL | 0.65 ± 0.16 | 61.12 ± 12.85 | 48.38 ± 21.23 | 73.85 ± 07.18 |
| | IFL | 0.55 ± 0.07 | 56.02 ± 05.29 | 56.23 ± 10.32 | 55.80 ± 08.49 |
| RRT-MIL [47] | FL | 0.59 ± 0.15 | 55.42 ± 08.54 | 43.45 ± 25.65 | 67.38 ± 25.52 |
| | IFL | 0.55 ± 0.04 | 55.62 ± 03.61 | 46.90 ± 24.09 | 64.35 ± 25.62 |
| MorphDistill | FL | 0.70 ± 0.08 | 62.00 ± 6.11 | **58.26 ± 8.21** | 66.73 ± 7.53 |
| | IFL | **0.65 ± 0.15** | **63.44 ± 13.5** | **62.13 ± 17.40** | 64.74 ± 15.1 |

#### 4.6.2. Sex-Stratified Analysis:

Table 4 presents model performance stratified by patient sex. MorphDistill achieves an AUC of 0.65 for female patients and 0.69 for male patients, indicating comparable predictive performance across demographic groups. Balanced accuracy, sensitivity, and specificity also remain consistent between subgroups, suggesting minimal sex-related bias in the learned representation.

**Table 4.** Five-year survival prediction results stratified by sex in the Alliance cohort using five-fold cross-validation. Each row reports mean performance with standard deviation.

| Model | Sex | AUC | Bal-Acc (%) | Sensitivity (%) | Specificity (%) |
|---|---|---|---|---|---|
| ABMIL [37] | Female | 0.63 ± 0.11 | 57.86 ± 10.13 | 51.48 ± 17.68 | 64.25 ± 7.50 |
| | Male | 0.61 ± 0.05 | 57.68 ± 5.75 | 46.04 ± 12.87 | 69.32 ± 7.46 |
| CLAM [10] | Female | 0.61 ± 0.16 | 52.75 ± 9.29 | 20.33 ± 12.13 | 85.16 ± 13.61 |
| | Male | 0.63 ± 0.03 | 58.35 ± 3.95 | 31.38 ± 8.05 | 85.33 ± 5.95 |
| TransMIL [43] | Female | **0.69 ± 0.12** | 54.18 ± 5.99 | 8.37 ± 11.99 | **100.00 ± 0.01** |
| | Male | 0.67 ± 0.06 | 52.67 ± 6.19 | 6.00 ± 12.00 | **99.35 ± 1.20** |
| Nakanishi et al. [46] | Female | 0.60 ± 0.11 | **58.32 ± 9.70** | **53.43 ± 20.25** | 63.21 ± 7.87 |
| | Male | 0.61 ± 0.06 | 59.99 ± 5.78 | 56.01 ± 12.59 | 63.97 ± 3.65 |
| RRT-MIL [47] | Female | 0.56 ± 0.07 | 54.11 ± 5.35 | 45.98 ± 30.91 | 62.24 ± 24.09 |
| | Male | 0.59 ± 0.08 | 56.17 ± 6.29 | 44.19 ± 21.72 | 68.16 ± 25.79 |
| MorphDistill | Female | 0.65 ± 0.11 | 58.19 ± 7.41 | 52.26 ± 18.45 | 62.33 ± 10.47 |
| | Male | **0.69 ± 0.12** | **69.54 ± 8.81** | **66.63 ± 12.27** | 70.35 ± 9.71 |

#### 4.6.3. Tumor Location–Stratified Analysis:

We further evaluate model performance across primary tumor locations within the colon. As shown in Table 5, despite substantial differences in sample sizes across anatomical subsites, MorphDistill maintains stable predictive performance across locations. For example, AUC values range from 0.63 for tumors originating in the cecum to 0.83 for those in the splenic flexure, with overlapping confidence intervals suggesting consistent performance across sites. These results indicate that the distilled representation captures morphological features that generalize across diverse tumor microenvironments within colorectal cancer. Collectively, these findings further demonstrate the robustness of the proposed representation for survival prediction in heterogeneous clinical settings.

**Table 5**. Five-year overall survival (OS) results stratified by tumor location in the Alliance cohort using five-fold cross-validation. Each row reports mean performance with standard deviation. The best-performing metric in each column is highlighted in bold.

| | Tumor Location | AUC | Bal-Acc (%) | Sensitivity (%) | Specificity (%) | n(Pos/Neg) |
|---|---|---|---|---|---|---|
| **Right Colon** | Cecum | 0.63 ± 0.20 | 62.26 ± 17.48 | 61.29 ± 21.20 | 63.23 ± 25.54 | 99 (24 / 75) |
| | Ascending Colon | 0.71 ± 0.06 | 60.69 ± 11.53 | 60.00 ± 27.89 | 61.38 ± 14.36 | 62 (16 / 46) |
| | Hepatic Flexure | **0.75 ± 0.27** | **65.83 ± 28.63** | **63.33 ± 41.50** | **68.33 ± 33.54** | 28 (12 / 16) |
| | Transverse Colon | 0.53 ± 0.34 | 59.11 ± 27.13 | 46.67 ± 50.55 | 58.25 ± 17.88 | 45 (9 / 36) |
| | Combined | 0.66 ± 0.22 | 61.97 ± 21.19 | 57.82 ± 35.29 | 62.80 ± 22.83 | 239 |
| **Left Colon** | Splenic | **0.83 ± 0.23** | 61.33 ± 24.93 | 50.50 ± 50.50 | 62.67 ± 26.81 | 18 (5 / 13) |
| | Descending Colon | 0.79 ± 0.25 | **69.67 ± 24.51** | **60.00 ± 54.77** | 71.33 ± 27.85 | 19 (4 / 15) |
| | Sigmoid | 0.67 ± 0.13 | 59.46 ± 8.29 | 46.98 ± 21.20 | **71.95 ± 11.04** | 147 (30 / 117) |
| | Combined | 0.76 ± 0.20 | 63.49 ± 19.24 | 52.49 ± 42.16 | 68.65 ± 21.90 | 187 |

### 4.7. External Validation on The Cancer Genome Atlas (TCGA) Colorectal Cancer Cohort

To evaluate the generalizability of MorphDistill beyond the Alliance/CALGB 89803 clinical trial cohort, we performed an additional validation on an independent cohort of 562 patients from the TCGA colorectal cancer (READ and COAD) datasets. As shown in Table 6, MorphDistill achieves a C-index of 0.6281 ± 0.07, outperforming ABMIL (0.6034 ± 0.05), RRT-MIL (0.5991 ± 0.08), PANTHER (0.5832 ± 0.07), and DSMIL (0.5000 ± 0.01). These results demonstrate that the unified morphological representation learned by MorphDistill also generalizes to an external, multi-institutional dataset with different patient demographics, slide preparation protocols, and scanning equipment, supporting its potential for broader clinical application.

**Table 6**. External validation of MorphDistill on The Cancer Genome Atlas (TCGA) colorectal cancer cohorts (READ and COAD). The table presents the concordance index (C-index) comparing the proposed MorphDistill encoder against established multiple instance learning (MIL) aggregation methods. MorphDistill achieves the highest C-index of 0.6281, demonstrating robust generalizability to an independent, multi-institutional public dataset. Results are reported as mean ± standard deviation across cross-validation folds. The best-performing metric in each column is highlighted in bold.

| **Model** | **C-Index** |
|---|---|
| ABMIL [37] | 0.6034 ± 0.05 |
| PANTHER[44] | 0.5832 ± 0.07 |
| DSMIL [45] | 0.5000 ± 0.01 |
| RRT-MIL [47] | 0.5991 ± 0.08 |
| **MorphDistill** | **0.6151 ± 0.07** |

### 4.8. Ablation

#### 4.8.1. Computational Efficiency and Model Complexity

In addition to predictive performance, we evaluated the computational efficiency of MorphDistill compared with existing pathology foundation models. As shown in Figure 7 and Table 7, MorphDistill achieves significantly faster inference while maintaining competitive model capacity. Specifically, MorphDistill processes 1,000 patches in approximately 1.5 seconds, whereas the average runtime across foundation models is 3.06 seconds, corresponding to a 2.04× speedup. MorphDistill also maintains a relatively compact architecture with 86 million parameters and a 768-dimensional embedding space, which is substantially smaller than several large-scale foundation models while still preserving strong representational capacity. These results demonstrate that multi-teacher distillation not only improves predictive performance but also enables efficient deployment for large-scale whole-slide image analysis, where thousands of patches must be processed per slide.

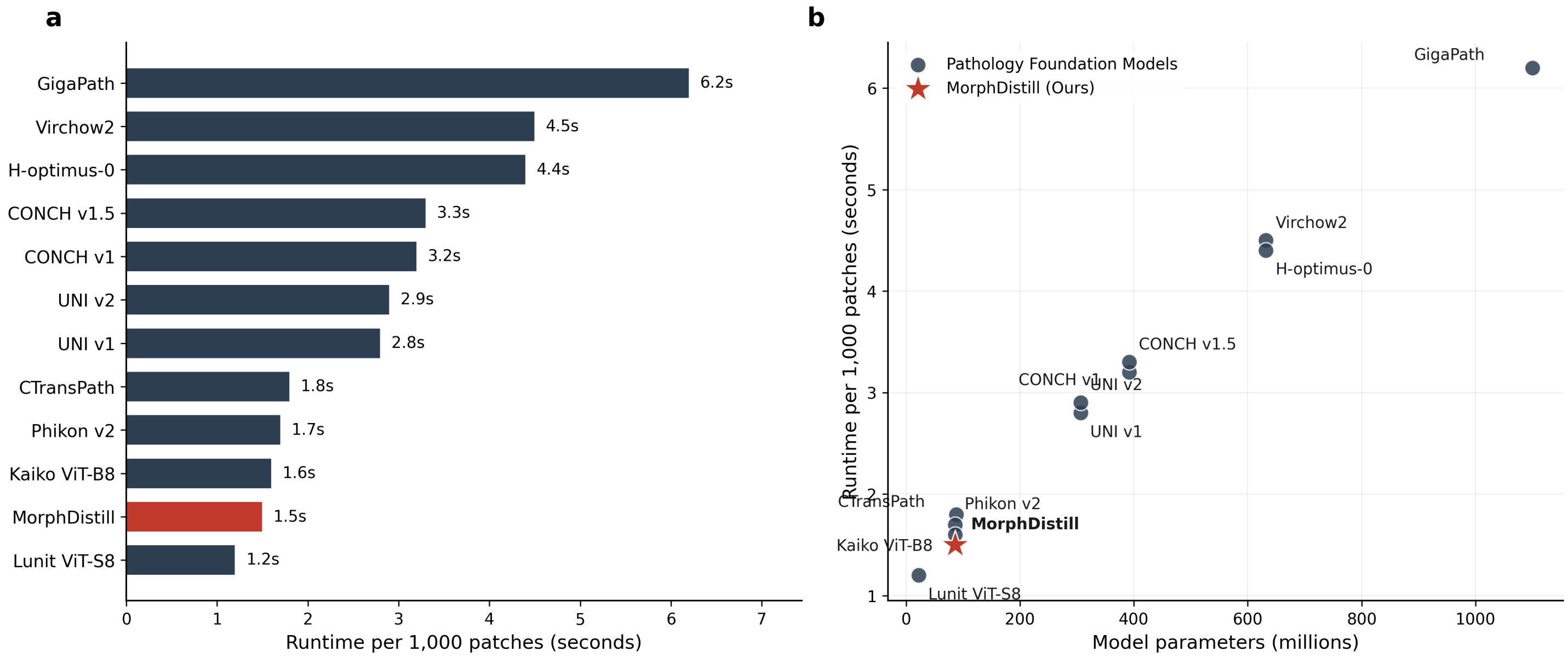

**Figure 7: Runtime efficiency of MorphDistill compared to pathology foundation models.** (a) Feature extraction runtime per 1,000 patches. (b) Model size versus runtime, demonstrating that MorphDistill achieves efficient inference (1.5s per 1,000 patches) with a compact 86M parameter architecture, 2.0x faster than the average foundation model (best viewed in color).

**Table 7. Runtime comparison of MorphDistill and pathology foundation models for patch-level feature extraction.** Computational efficiency was benchmarked across 12 models using the TRIDENT framework on an NVIDIA A100 GPU with batch size 32. Runtime was measured for 256×256-pixel patches extracted at 40× magnification. MorphDistill achieves 2.04× speedup compared to the average foundation model runtime (3.06 s/1K patches) while maintaining a compact architecture (86M parameters, 768-dimensional embeddings). Speedup metrics are calculated relative to the slowest model (GigaPath, 6.2 s/1K patches) and the average runtime across all foundation models.

| Model | Architecture | Parameters (M) | Embed-Dim | Pretraining | Training Data | Runtime per 1K patches (s) | Speedup vs Slowest | Speedup vs Avg Foundation Model |
|---|---|---|---|---|---|---|---|---|
| Lunit ViT-S8[16] | ViT-S/8 | 22 | 384 | DINO | 32K WSIs | 1.2 | 5.17 | 2.55 |
| MorphDistill (ours) | ViT-B/16 | 86 | 768 | Multi-teacher distillation | CRC dataset | 1.5 | 4.13 | 2.04 |
| Kaiko ViT-B8[15] | ViT-B/8 | 86 | 768 | DINOv2 | Public datasets | 1.6 | 3.88 | 1.91 |
| Phikon v2[17] | ViT-B/16 | 86 | 768 | iBOT | 43M patches | 1.7 | 3.65 | 1.8 |
| CTransPath[12] | Swin-T | 88 | 768 | SRCL | 15M patches | 1.8 | 3.44 | 1.7 |
| UNI v1[18] | ViT-L/16 | 307 | 1,024 | DINOv2 | 100M+ patches | 2.8 | 2.21 | 1.09 |
| UNI v2[19] | ViT-L/16 | 307 | 1,024 | DINOv2 | 100M+ patches | 2.9 | 2.14 | 1.05 |
| CONCH v1[11] | ViT-B/16 + GPT-2 | 392 | 512 | CLIP-style | 1.17M patches | 3.2 | 1.94 | 0.95 |
| CONCH v1.5[11] | ViT-B/16 + GPT-2 | 392 | 512 | CLIP-style | 1.17M patches | 3.3 | 1.88 | 0.93 |
| H-optimus-0[14] | ViT-H/14 | 632 | 1,280 | DINOv2 | 500K+ WSIs | 4.4 | 1.41 | 0.69 |
| Virchow2[20] | ViT-H/14 | 632 | 1,280 | DINOv2 | 3M+ WSIs | 4.5 | 1.38 | 0.68 |
| GigaPath[13] | ViT-g/14 | 1,100 | 1,536 | DINOv2 | 170K WSIs | 6.2 | 1 | 0.49 |

### 4.8.2. Evaluating Learned Representations for Downstream Tissue Classification

To better understand the contribution of the proposed representation learning framework, we conducted an ablation study to evaluate the impact of different Stage I training strategies on the learned representation. Table 8 compares the embedding quality of MorphDistill with existing pathology foundation models using Linear Probe (LP) and k-Nearest Neighbor (KNN) evaluation protocols for colorectal tissue morphology classification. These experiments provide a proxy evaluation of the discriminative capacity of the learned representations independent of the downstream survival task. Across both LP and KNN protocols, MorphDistill demonstrates competitive or improved performance relative to existing encoders, indicating that the distilled representation preserves meaningful morphological structure and produces embeddings suitable for colorectal tissue morphology classification.

**Table 8.** Comparison of Linear Probe (LP) and k-Nearest Neighbors (KNN) classification performance across pretrained pathology foundation models and the proposed MorphDistill representation. The evaluation assesses the quality of learned embeddings for basic colorectal tissue morphology classification. For KNN, the reported results correspond to the best-performing value of $K$. Performance is reported using Accuracy, Macro-F1, and Weighted-F1 metrics, with the best result for each method highlighted in bold. MorphDistill achieves improved classification performance, indicating that the distilled representation captures discriminative morphological features relevant for tissue classification. The best-performing metric in each column is highlighted in bold.

| Model | LP Acc | LP Macro-F1 | LP Weighted F1 | KNN (K) | KNN Acc | KNN Macro-F1 | KNN Weighted F1 |
|---|---|---|---|---|---|---|---|
| CONCH v1.5[11] | 84.91 | 81.29 | 85.76 | 30 | 83.64 | 80.67 | 84.10 |
| CTransPath[12] | 83.99 | 80.24 | 84.79 | 30 | 84.20 | 80.87 | 84.73 |
| GigaPath[13] | 85.28 | 82.04 | 85.98 | 30 | 84.28 | 82.01 | 84.81 |
| H-Optimus-0[14] | 85.25 | 81.43 | 85.96 | 30 | 85.53 | 82.78 | 86.14 |
| Kaiko ViT-B/8[15] | 85.25 | 82.01 | 85.93 | 30 | 85.16 | 81.89 | 85.70 |
| Lunit ViT-S/8[16] | 83.26 | 80.25 | 84.13 | 30 | 83.82 | 81.07 | 84.40 |
| Phikon v2[17] | 84.48 | 79.84 | 85.31 | 30 | 83.56 | 80.89 | 84.06 |
| UNI v1[18] | 85.98 | 82.18 | 86.68 | 30 | 85.32 | 82.49 | 85.91 |
| UNI v2[19] | 86.21 | 82.28 | 86.87 | 30 | 84.65 | 82.40 | 85.15 |
| Virchow2[20] | 85.92 | 82.76 | 86.54 | 30 | 85.87 | 83.10 | 86.50 |
| MorphDistill (ours) | **86.93** | **83.30** | **87.60** | **30** | **86.26** | **83.40** | **86.90** |

### 4.8.3. Impact of Training Strategies on Representation Learning for Downstream Survival Prediction

We further analyzed the impact of different representation learning strategies on downstream survival prediction using the Alliance cohort. Specifically, we evaluated six configurations that systematically vary the objective function derived from Equation (6): (i) sup: supervised learning using only cross-entropy loss (removing both $\mathcal{L}_{supcon}$ and $\mathcal{L}_{dist}$); (ii) sup-distill: supervised learning with $\mathcal{L}_{dist}$ replacing cross-entropy; (iii) unsup: contrastive learning using only $\mathcal{L}_{supcon}$ (simplified to instance-level contrastive loss without class labels); (iv) unsup-distill: contrastive learning combined with $\mathcal{L}_{dist}$; (v) supcon: supervised contrastive learning using only $\mathcal{L}_{supcon}$ (with $\lambda = 1$); and (vi) supcon-distill (MorphDistill): full objective with $\lambda = 0.75$, combining $\mathcal{L}_{supcon}$ and $\mathcal{L}_{dist}$. As shown in Table 9, models trained with different combinations of supervision, contrastive learning, and relational distillation show distinct levels of prognostic performance. Incorporating multi-teacher relational distillation consistently improves performance across training paradigms. In particular, the supervised contrastive learning with relational distillation configuration (MorphDistill) achieves the strongest results, with an AUC of 0.68 ± 0.08 and balanced accuracy of 64.11%. Kaplan–Meier analyses (Fig. 8) demonstrate clearer separation between predicted low- and high-risk patient groups, while time-to-event metrics (Fig. 9) show higher concordance and hazard ratios compared with alternative training strategies. Together, these findings suggest that combining supervised contrastive learning with multi-teacher relational distillation enables the model to learn representations that more effectively capture prognostically relevant morphological patterns in colorectal cancer histopathology.

**Table 9.** Ablation study of training strategies used during MorphDistill representation learning and their impact on downstream five-year survival prediction. The MorphDistill encoder is trained under different training configurations during Stage I, and the resulting pretrained encoders are used as frozen feature extractors to encode whole-slide images from the Alliance cohort. The extracted features are then used for downstream five-year survival prediction to evaluate how different

representation learning strategies influence prognostic performance. The best-performing metric in each column is highlighted in bold.

| MorphDistill (MD) Variant | Distillation | Training Strategy | AUC | Bal-Acc (%) | Sensitivity (%) | Specificity (%) |
|---|---|---|---|---|---|---|
| $MorphDistill_{sup}$ | ✗ | Supervised | 0.60 ± 0.07 | 57.62 ± 7.13 | 55.52 ± 14.77 | 59.72 ± 12.79 |
| $MorphDistill_{sup-distill}$ | ✓ | Supervised | 0.64 ± 0.05 | 59.87 ± 4.68 | 60.38 ± 14.61 | 59.35 ± 10.90 |
| $MorphDistill_{unsup}$ | ✗ | Contrastive Learning | 0.62 ± 0.02 | 58.67 ± 5.61 | 36.76 ± 12.99 | **80.59 ± 8.75** |
| $MorphDistill_{unsup-distill}$ | ✓ | Contrastive Learning | 0.63 ± 0.07 | 60.95 ± 7.16 | 44.14 ± 20.33 | 77.75 ± 9.36 |
| $MorphDistill_{supcon}$ | ✗ | Supervised Contrastive Learning | 0.62 ± 0.09 | 58.57 ± 10.10 | **63.29 ± 25.51** | 53.86 ± 17.80 |
| $MorphDistill_{supcon-distil}$ | ✓ | Supervised Contrastive Learning | **0.68 ± 0.08** | **64.11 ± 4.70** | 60.24 ± 5.38 | 66.57 ± 8.81 |

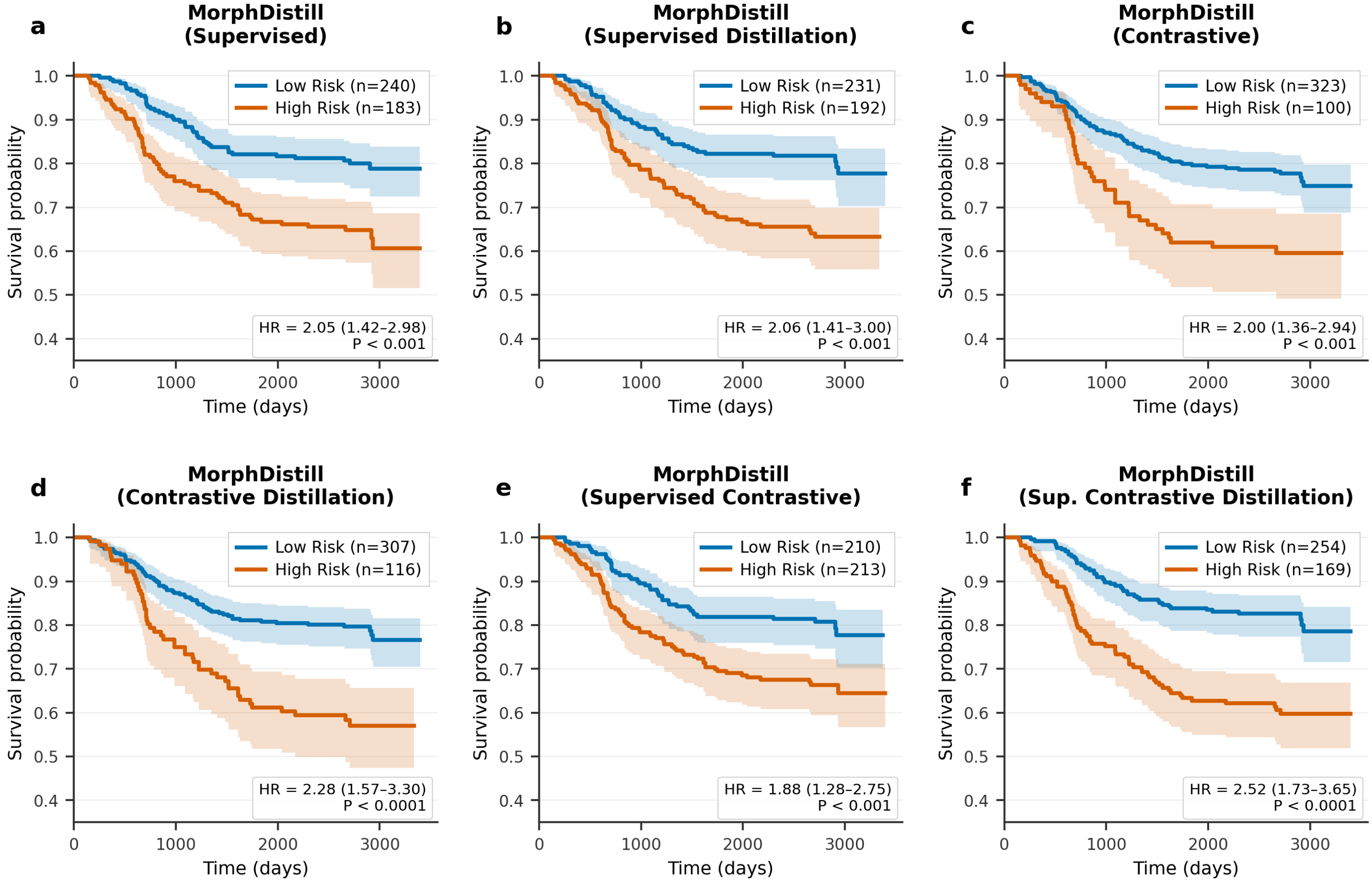

**Figure 8:** Kaplan–Meier survival curves for the MorphDistill ablation study evaluating different Stage I representation learning strategies. Each panel shows survival probability over time for patients stratified into low-risk (blue) and high-risk (orange) groups based on model predictions. The evaluated configurations include: (a) supervised learning with cross-entropy loss; (b) supervised learning with knowledge distillation; (c) contrastive learning without distillation; (d) contrastive learning with knowledge distillation; (e) supervised contrastive learning without distillation; and (f) supervised contrastive learning with knowledge distillation (MorphDistill). Hazard ratios (HR) with 95% confidence intervals and log-rank p-values are reported for each configuration. Shaded regions represent 95% confidence intervals. The supervised contrastive distillation configuration (f) yields the strongest survival stratification among the evaluated settings (HR = 2.52, 95% CI: 1.73–3.65) (best viewed in color).

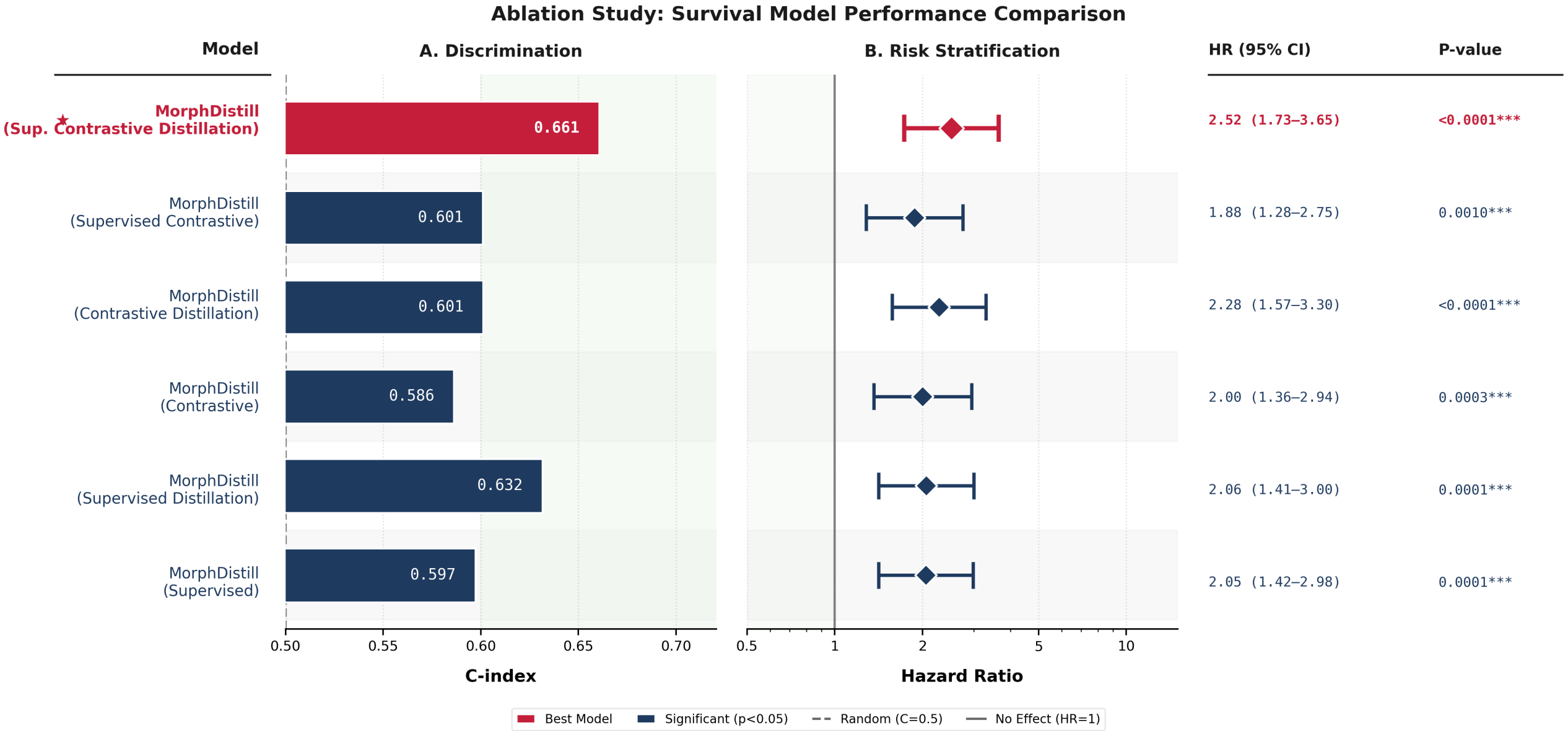

**Figure 9:** Performance comparison of MorphDistill ablation configurations. Panel (A) shows discrimination performance measured by the concordance index (C-index), and panel (B) shows risk stratification measured by hazard ratio (HR) with 95% confidence intervals across the evaluated training strategies. The supervised contrastive distillation configuration (MorphDistill, highlighted in red) achieves the highest performance with a C-index of 0.661 and HR of 2.52 (95% CI: 1.73–3.65, $P < 0.0001$). The dashed vertical line in panel (A) indicates random prediction (C-index = 0.5), while the solid vertical line in panel (B) denotes the null effect threshold (HR = 1). Across the ablation configurations, models incorporating knowledge distillation demonstrate improved survival discrimination and risk stratification (best viewed in color).

## 5. Discussion

We present MorphDistill, a multi-teacher knowledge distillation framework that unifies heterogeneous morphological knowledge from ten pathology foundation models into a compact, CRC-specific encoder for survival prediction. The central finding of this study is that synthesizing complementary representations from multiple pretrained models yields a more prognostically informative representation than relying on any single foundation encoder. On the Alliance/CALGB 89803 cohort, MorphDistill improves predictive performance relative to the strongest baseline encoder (UNI v2), achieving an approximately 8% improvement in AUC (0.68 vs. 0.63) along with stronger risk stratification, reflected by a C-index of 0.661 and hazard ratio of 2.52 (95% CI: 1.73–3.65) compared with HR 2.08 for UNI v2 (Table 1, Figure 3, and 4). The stronger separation between predicted high-risk and low-risk groups suggests improved clinical stratification, indicating that the unified representation learned by MorphDistill captures morphological patterns associated with survival outcomes more effectively than individual foundation encoders. The generalizability of these findings is further supported by external validation on the independent TCGA colorectal cancer cohort (READ and COAD), where MorphDistill achieved a concordance index of 0.628, outperforming established MIL aggregation methods (Table 6). These results demonstrate that the collective knowledge embedded in existing foundation models can be effectively distilled and integrated to create task-specific representations that outperform individual encoders.

The key methodological contribution of MorphDistill is its dimension-agnostic relational distillation strategy, which addresses a central challenge in computational pathology: the heterogeneous feature representations produced by foundation models trained on different datasets, architectures, and learning objectives. Rather than aligning individual feature vectors, our framework aligns batch-wise relational structures among samples, enabling knowledge transfer across teachers with varying embedding dimensionalities while preserving each model's original representational geometry. When combined with supervised contrastive regularization, this approach encourages the learned embedding space to capture both global morphological relationships and CRC-specific tissue semantics, producing a representation tailored for survival prediction. Ablation analyses presented in the supplementary material further demonstrate that combining supervised contrastive learning with multi-teacher relational distillation consistently improves survival prediction and risk stratification compared with alternative training strategies (Table 9, and Figures 8 and 9).

The results further demonstrate that the benefits of MorphDistill extend beyond a single aggregation architecture. When evaluated across multiple MIL frameworks—including ABMIL, CLAM, TransMIL, Nakanishi et al., and RRT-MIL—MorphDistill embeddings consistently improved survival prediction compared with UNI v2 features (Table 2). Performance gains were particularly evident in sensitivity, suggesting that the unified representation more effectively captures morphological patterns associated with poor prognosis. Kaplan–Meier analyses across aggregation methods further confirmed stronger patient risk stratification when MorphDistill embeddings were used (Fig. 5), while higher C-index values indicate improved concordance between predicted risk and observed survival outcomes and the consistently larger hazard ratios reflect stronger separation between predicted risk groups across aggregation frameworks (Fig. 6). Together, these results demonstrate that the performance gains of MorphDistill arise primarily from the learned representation itself rather than a specific MIL architecture.

From a clinical perspective, accurate survival prediction in stage III colorectal cancer is critical for guiding adjuvant therapy decisions. Patients with poor prognosis may benefit from more aggressive treatment strategies, whereas those with favorable outcomes may avoid unnecessary treatment-related toxicity. The consistent performance of MorphDistill across treatment regimens, demographic subgroups, and tumor locations (Tables 3, 4 and 5) suggests that the learned representation captures intrinsic morphological characteristics of CRC rather than dataset-specific artifacts. The observed hazard ratio of 2.52 indicates that patients identified as high-risk by MorphDistill exhibit more than twice the mortality risk of low-risk patients, supporting the potential clinical utility of the proposed approach.

Beyond predictive performance, MorphDistill provides practical advantages for deployment in computational pathology workflows. By distilling knowledge from multiple foundation models into a single encoder, the framework produces a compact and computationally efficient representation while preserving strong predictive performance. As shown in Figure 7 and Table 7, MorphDistill processes 1,000 patches in approximately 1.5 seconds, compared with an average runtime of 3.06 seconds across foundation models, corresponding to a 2× speedup in feature extraction. In addition to improved runtime efficiency, MorphDistill maintains a relatively compact architecture with 86 million parameters, substantially smaller than several large-scale pathology foundation models. These properties are particularly important for whole-slide image analysis, where thousands of patches must be processed per slide. Consequently, the proposed framework not only improves prognostic representation learning but also enables more efficient and scalable deployment in clinical computational pathology pipelines.

More broadly, this work suggests a new direction for representation learning in computational pathology. Rather than training task-specific models from scratch or relying on a single foundation encoder, it may be more effective to leverage and integrate the complementary knowledge embedded in multiple pretrained models. Such an approach offers several advantages: it avoids the computational expense of large-scale pretraining, synthesizes information from diverse datasets and architectures, and enables the development of task-specific representations that can evolve as new foundation models become available. Similar multi-teacher distillation strategies could therefore be applied to other cancer types and clinical prediction tasks.

## 6. Limitation and Future Directions:

Despite the promising results, several limitations should be acknowledged. Although we have validated MorphDistill on the independent TCGA colorectal cancer cohort (READ and COAD), demonstrating improved generalizability beyond the Alliance/CALGB 89803 clinical trial cohort, the evaluation is still limited to stage III and a subset of other disease stages within these datasets. Further validation on additional independent cohorts spanning broader disease stages (e.g., stage I/II and stage IV) and diverse patient populations will be necessary to fully assess the generalizability of the proposed representation. Second, the current framework relies solely on histopathological morphology derived from H&E-stained whole-slide images. While tissue morphology contains substantial prognostic information, survival outcomes are also influenced by clinical variables, genomic alterations, and treatment-specific factors. Future work could explore multimodal integration, combining MorphDistill embeddings with molecular and clinical data to further improve prognostic modeling. Third, the proposed framework distills knowledge from ten publicly available pathology foundation models, but the optimal selection and weighting of teacher models remain open questions. As new foundation models continue to emerge, investigating adaptive teacher selection or incremental distillation strategies may further improve representation quality. Finally, the current approach performs patch-level representation learning followed by slide-level aggregation using MIL frameworks. Future work may explore end-to-end distillation strategies that jointly optimize representation learning and survival prediction, potentially enabling more direct modeling of spatial interactions within whole-slide images. Overall, extending MorphDistill through larger-scale validation, multimodal learning, and improved distillation strategies represents an important direction for advancing computational pathology–based prognostic modeling.

## 7. Conclusions

In this work, we introduced MorphDistill, a multi-teacher distillation framework for learning task-specific morphological representations for colorectal cancer survival prediction from whole-slide images. By combining dimension-agnostic relational distillation with supervised contrastive regularization, MorphDistill integrates complementary knowledge from multiple pathology foundation models into a unified CRC-specific representation. Evaluation on the Alliance/CALGB 89803 cohort shows that MorphDistill improves predictive performance relative to the strongest baseline foundation models, achieving an approximately 8% improvement in AUC (0.68 vs. 0.63) along with a C-index of 0.661 and hazard ratio of 2.52 (95% CI: 1.73–3.65) for five-year survival prediction. The learned representation demonstrates robust performance across MIL aggregation frameworks, treatment regimens, demographic subgroups, and tumor locations, with strong generalizability confirmed by external validation on the independent TCGA colorectal cancer cohorts (COAD and READ), where MorphDistill achieved a superior C-index of 0.6151±0.07, outperforming the best MIL baseline (ABMIL, C-index 0.6034) by approximately 4%. In addition, MorphDistill provides computational efficiency, producing a compact encoder suitable for large-scale whole-slide image analysis. Overall, MorphDistill establishes a new paradigm for task-specific representation learning in computational pathology: rather than training new models from scratch or relying on a single encoder, it distills complementary knowledge from multiple foundation models into a unified representation. Future work will focus on external validation across larger multi-institutional cohorts, extension to additional cancer types and prediction tasks, and integration with molecular and clinical data for multimodal prognostic modeling. The source code and implementation are publicly available at https://github.com.mcas.ms/hikmatkhan/MorphDistill.

## Declaration of Competing Interest

All authors declare no financial or non-financial competing interests.

## Ethics Approval and Consent to Participate

This study was reviewed and approved by the Institutional Review Board of Ohio State University (IRB #2018C0098). All procedures involving human data were conducted in accordance with the ethical standards of the institutional research committee, the national research regulations, and with the 1964 Declaration of Helsinki and its later amendments. Given the retrospective nature of the study and because data were archival and de-identified, the requirement for informed consent to participate was formally waived by the Institutional Review Board.

## Funding

The project described was supported in part by R01 CA276301 (PIs: Niazi and Chen) from the National Cancer Institute. The project was also supported partly by Pelotonia under IRP CC13702 (PIs: Niazi, Vilgelm, and Roy), The Ohio State University Department of Pathology and Comprehensive Cancer Center. The content is solely the responsibility of the authors and does not necessarily represent the official views of the National Cancer Institute, National Institutes of Health, or The Ohio State University.

**Author Contributions**

H.K. performed data preprocessing, experimental design, validation, and manuscript writing and editing. U.S., M.N.G., A.P., W.L.F., W.C., and M.K.K.N. assisted with manuscript editing and reviewing.

**Acknowledgements**

The authors thank Dr. Fang-Shu Ou and the Alliance Statistics and Data Management Center for data acquisition, curation, and management. The authors also gratefully acknowledge the Ohio Supercomputer Center for providing high-performance computing resources under its contract with The Ohio State University College of Medicine. We also thank the Department of Pathology and the Comprehensive Cancer Center at The Ohio State University for their valuable support.

**Clinicaltrials.gov Identifier**

NCT00003835 (CALGB 89803); Registry: CTRP (Clinical Trial Reporting Program); Registry Identifier: NCI-2012-01844

**Support**

The data from CALGB 89803 were obtained directly from the Alliance for Clinical Trials in Oncology, a National Clinical Trials Network cooperative group; however, all analyses and conclusions in this manuscript are the sole responsibility of the authors and do not necessarily reflect the opinions or views of the clinical trial investigators, the NCTN, the NCORP or the NCI.

**Data and Code Availability**

The MorphDistill source code, including Stage I encoder training, Stage II survival prediction, and feature extraction pipelines, is publicly available at https://github.com.mcas.ms/hikmatkhan/MorphDistill. The patch-level pre-training datasets used in Stage I — CRC-100K, STARC-9, and SPIDER-Colorectal — are publicly accessible through their respective original publications. The Alliance/CALGB 89803 whole-slide images and associated clinical outcomes used for Stage II survival prediction are available to qualified investigators through the Alliance for Clinical Trials in Oncology upon reasonable request and in accordance with institutional data sharing agreements. The external validation cohort from The Cancer Genome Atlas (TCGA-COAD and TCGA-READ) is publicly available via the GDC Data Portal (https://portal.gdc.cancer.gov/). For further inquiries regarding data access, code usage, or reproducibility, please contact the corresponding author at Hikmat.khan@osumc.edu or Hikmat.khan179@gmail.com.

**Videos:** A short overview of this work is available at:

- **YouTube Short:**
  - https://www.youtube.com/shorts/5JT7UU_JHwE
  - https://www.youtube.com/shorts/CLIxJ2WDj1A
- **Full Video:**
  - https://www.youtube.com/watch?v=oq78CdJhhto
  - https://www.youtube.com/watch?v=JLHg5vrwcTQ

- https://youtu.be/zo3xrPfuX2g?si=boYAFC-ZpR-40cCu&t=469
- https://www.youtube.com/watch?v=Dow8bv37CtI

## References

1. Rawla, P.; Sunkara, T.; Barsouk, A. Epidemiology of colorectal cancer: incidence, mortality, survival, and risk factors. *Gastroenterology Review/Przegląd Gastroenterologiczny* **2019**, *14*, 89-103.
2. Siegel, R.L.; Kratzer, T.B.; Giaquinto, A.N.; Sung, H.; Jemal, A. Cancer statistics, 2025. *Ca* **2025**, *75*, 10.
3. Pollheimer, M.J.; Kornprat, P.; Lindtner, R.A.; Harbaum, L.; Schlemmer, A.; Rehak, P.; Langner, C. Tumor necrosis is a new promising prognostic factor in colorectal cancer. *Human pathology* **2010**, *41*, 1749-1757.
4. Fan, S.; Cui, X.; Zheng, L.; Ma, W.; Zheng, S.; Wang, J.; Qi, L.; Ye, Z. Prognostic value of desmoplastic stromal reaction, tumor budding and tumor-stroma ratio in stage II colorectal cancer. *Journal of Gastrointestinal Oncology* **2022**, *13*, 2903.
5. Su, Z.; Chen, W.; Annem, S.; Sajjad, U.; Rezapour, M.; Frankel, W.L.; Gurcan, M.N.; Niazi, M.K.K. Adapting SAM to histopathology images for tumor bud segmentation in colorectal cancer. In Proceedings of the Proceedings of SPIE--the International Society for Optical Engineering, 2024; p. 129330C.
6. Hugen, N.; Verhoeven, R.; Radema, S.; De Hingh, I.; Pruijt, J.; Nagtegaal, I.; Lemmens, V.; De Wilt, J. Prognosis and value of adjuvant chemotherapy in stage III mucinous colorectal carcinoma. *Annals of oncology* **2013**, *24*, 2819-2824.
7. Huh, J.W.; Lee, J.H.; Kim, H.R. Prognostic significance of tumor-infiltrating lymphocytes for patients with colorectal cancer. *Archives of surgery* **2012**, *147*, 366-372.
8. Mesker, W.E.; Junggeburt, J.M.; Szuhai, K.; de Heer, P.; Morreau, H.; Tanke, H.J.; Tollenaar, R.A. The carcinoma–stromal ratio of colon carcinoma is an independent factor for survival compared to lymph node status and tumor stage. *Analytical Cellular Pathology* **2007**, *29*, 387-398.
9. Pai, R.K.; Banerjee, I.; Shivji, S.; Jain, S.; Hartman, D.; Buchanan, D.D.; Jenkins, M.A.; Schaeffer, D.F.; Rosty, C.; Como, J. Quantitative pathologic analysis of digitized images of colorectal carcinoma improves prediction of recurrence-free survival. *Gastroenterology* **2022**, *163*, 1531-1546. e1538.
10. Ilse, M.; Tomczak, J.; Welling, M. Attention-based deep multiple instance learning. In Proceedings of the International conference on machine learning, 2018; pp. 2127-2136.
11. Lu, M.Y.; Chen, B.; Williamson, D.F.; Chen, R.J.; Liang, I.; Ding, T.; Jaume, G.; Odintsov, I.; Le, L.P.; Gerber, G. A visual-language foundation model for computational pathology. *Nature Medicine* **2024**, *30*, 863-874.
12. Wang, X.; Zhao, J.; Marostica, E.; Yuan, W.; Jin, J.; Zhang, J.; Li, R.; Tang, H.; Wang, K.; Li, Y. A pathology foundation model for cancer diagnosis and prognosis prediction. *Nature* **2024**, *634*, 970-978.
13. Xu, H.; Usuyama, N.; Bagga, J.; Zhang, S.; Rao, R.; Naumann, T.; Wong, C.; Gero, Z.; González, J.; Gu, Y. A whole-slide foundation model for digital pathology from real-world data. *Nature* **2024**, *630*, 181-188.
14. Saillard, C.; Jenatton, R.; Llinares-López, F.; Mariet, Z.; Cahané, D.; Durand, E.; Vert, J.-P. *H-optimus-0*, 2024.
15. Aben, N.; de Jong, E.D.; Gatopoulos, I.; Känzig, N.; Karasikov, M.; Lagré, A.; Moser, R.; van Doorn, J.; Tang, F. Towards large-scale training of pathology foundation models. *arXiv preprint arXiv:2404.15217* **2024**.
16. Kang, M.; Song, H.; Park, S.; Yoo, D.; Pereira, S. Benchmarking self-supervised learning on diverse pathology datasets. In Proceedings of the Proceedings of the IEEE/CVF Conference on Computer Vision and Pattern Recognition, 2023; pp. 3344-3354.
17. Filiot, A.; Jacob, P.; Kain, A.M.; Saillard, C. Phikon-v2, A large and public feature extractor for biomarker prediction. *ArXiv* **2024**, *abs/2409.09173*.
18. Chen, R.J.a.D.T.a.L.M.Y.a.W.D.F.K.a.J.G.a.S.A.H.a.C.B.a.Z.A.a.; et al. Towards a general-purpose foundation model for computational pathology. *Nature Medicine* **2024**, *30*, 850--862.
19. Chen, R.J.; Ding, T.; Lu, M.Y.; Williamson, D.F.; Jaume, G.; Song, A.H.; Chen, B.; Zhang, A.; Shao, D.; Shaban, M. Towards a general-purpose foundation model for computational pathology. *Nature Medicine* **2024**, *30*, 850-862.
20. Zimmermann, E.; Vorontsov, E.; Viret, J.; Casson, A.; Zelechowski, M.; Shaikovski, G.; Tenenholtz, N.; Hall, J.; Klimstra, D.; Yousfi, R. Virchow2: Scaling self-supervised mixed magnification models in pathology. *arXiv preprint arXiv:2408.00738* **2024**.
21. Khan, H.; Su, Z.; Zhang, H.; Wang, Y.; Ning, B.; Wei, S.; Guo, H.; Li, Z.; Niazi, M.K.K. Predicting Neoadjuvant Chemotherapy Response in Triple-Negative Breast Cancer Using Pre-Treatment Histopathologic Images. *Cancers* **2025**, *17*, 2423.

22. Wang, X.; Yang, S.; Zhang, J.; Wang, M.; Zhang, J.; Yang, W.; Huang, J.; Han, X. Transformer-based unsupervised contrastive learning for histopathological image classification. *Medical image analysis* **2022**, *81*, 102559.
23. Vorontsov, E.; Bozkurt, A.; Casson, A.; Shaikovski, G.; Zelechowski, M.; Liu, S.; Severson, K.; Zimmermann, E.; Hall, J.; Tenenholtz, N. Virchow: A million-slide digital pathology foundation model. *arXiv preprint arXiv:2309.07778* **2023**.
24. Filiot, A.; Jacob, P.; Mac Kain, A.; Saillard, C. Phikon-v2, a large and public feature extractor for biomarker prediction. *arXiv preprint arXiv:2409.09173* **2024**.
25. Ma, J.; Guo, Z.; Zhou, F.; Wang, Y.; Xu, Y.; Li, J.; Yan, F.; Cai, Y.; Zhu, Z.; Jin, C. Towards a generalizable pathology foundation model via unified knowledge distillation. *arXiv preprint arXiv:2407.18449* **2024**.
26. Chen, Y.; Su, Z.; Khan, H.; Niazi, M.K.K. RANGER: Sparsely-Gated Mixture-of-Experts with Adaptive Retrieval Re-ranking for Pathology Report Generation. *arXiv preprint arXiv:2603.04348* **2026**.
27. Khan, H.; Chen, W.; Niazi, M.K.K. Weakly Supervised Teacher–Student Framework with Progressive Pseudo-mask Refinement for Gland Segmentation. *Journal of Clinical and Translational Pathology* **2026**, *6*, 1-12.
28. Chen, Y.; Su, Z.; Meng, L.; Hasanov, E.; Chen, W.; Parwani, A.; Niazi, M. HistoMet: A Pan-Cancer Deep Learning Framework for Prognostic Prediction of Metastatic Progression and Site Tropism from Primary Tumor Histopathology. *arXiv preprint arXiv:2602.07608* **2026**.
29. Sajjad, U.; Akbar, A.; Khan, H.; Frankel, W.; Gurcan, M.; Parwani, A.; Chen, W.; Niazi, M.K.K. 711 An AI Virtual Agent for Prognostic Interpretation: Translating Black-Box Predictions into Clinical Insights. *Laboratory Investigation* **2026**, *106*.
30. Liu, W.; Khan, H.; Rajab, A.; Niazi, M.K.K.; Jin, Y. 1228 AI-Based Detection of Megakaryocyte Dysplasia in Myelodysplastic Neoplasms. *Laboratory Investigation* **2026**, *106*.
31. Datwani, S.; Khan, H.; Niazi, M.K.K.; Parwani, A.V.; Li, Z. Artificial intelligence in breast pathology: Overview and recent updates. *Human pathology* **2025**, 105819.
32. Khan, H.; Zaidi, S.F.A.; Shah, P.M.; Balakrishnan, K.; Khan, R.; Waqas, M.; Wu, J. MorphGen: Morphology-Guided Representation Learning for Robust Single-Domain Generalization in Histopathological Cancer Classification. *arXiv preprint arXiv:2509.00311* **2025**.
33. Belagali, V.; Kapse, S.; Marza, P.; Das, S.; Li, Z.; Boutaj, S.; Pati, P.; Yellapragada, S.; Nandi, T.N.; Madduri, R.K. TICON: A Slide-Level Tile Contextualizer for Histopathology Representation Learning. *arXiv preprint arXiv:2512.21331* **2025**.
34. Kather, J.N.; Halama, N.; Marx, A. 100,000 histological images of human colorectal cancer and healthy tissue. *(No Title)* **2018**.
35. Subramanian, B.; Jeyaraj, R.; Peterson, M.N.; Guo, T.; Shah, N.; Langlotz, C.; Ng, A.Y.; Shen, J. STARC-9: A Large-scale Dataset for Multi-Class Tissue Classification for CRC Histopathology. In Proceedings of the The Thirty-ninth Annual Conference on Neural Information Processing Systems Datasets and Benchmarks Track, 2025.
36. Nechaev, D.; Pchelnikov, A.; Ivanova, E. SPIDER: A Comprehensive Multi-Organ Supervised Pathology Dataset and Baseline Models. *arXiv preprint arXiv:2503.02876* **2025**.
37. Lu, M.Y.; Williamson, D.F.; Chen, T.Y.; Chen, R.J.; Barbieri, M.; Mahmood, F. Data-efficient and weakly supervised computational pathology on whole-slide images. *Nature biomedical engineering* **2021**, *5*, 555-570.
38. Olenius, T.; Koskenvuo, L.; Koskensalo, S.; Lepistö, A.; Böckelman, C. Long-term survival among colorectal cancer patients in Finland, 1991–2015: a nationwide population-based registry study. *BMC cancer* **2022**, *22*, 356.
39. Sargent, D.J.; Wieand, H.S.; Haller, D.G.; Gray, R.; Benedetti, J.K.; Buyse, M.; Labianca, R.; Seitz, J.F.; O'Callaghan, C.J.; Francini, G. Disease-free survival versus overall survival as a primary end point for adjuvant colon cancer studies: individual patient data from 20,898 patients on 18 randomized trials. *Journal of Clinical Oncology* **2005**, *23*, 8664-8670.
40. van den Berg, I.; Coebergh van den Braak, R.R.; van Vugt, J.L.; Ijzermans, J.N.; Buettner, S. Actual survival after resection of primary colorectal cancer: results from a prospective multicenter study. *World journal of surgical oncology* **2021**, *19*, 96.
41. Tarazi, M.; Guest, K.; Cook, A.J.; Balasubramaniam, D.; Bailey, C.M. Two and five year survival for colorectal cancer after resection with curative intent: A retrospective cohort study. *International Journal of Surgery* **2018**, *55*, 152-155.
42. Zhang, A.; Jaume, G.; Vaidya, A.; Ding, T.; Mahmood, F. Accelerating data processing and benchmarking of ai models for pathology. *arXiv preprint arXiv:2502.06750* **2025**.

43. Shao, Z.; Bian, H.; Chen, Y.; Wang, Y.; Zhang, J.; Ji, X. Transmil: Transformer based correlated multiple instance learning for whole slide image classification. *Advances in neural information processing systems* **2021**, *34*, 2136-2147.
44. Song, A.H.; Chen, R.J.; Ding, T.; Williamson, D.F.; Jaume, G.; Mahmood, F. Morphological prototyping for unsupervised slide representation learning in computational pathology. In Proceedings of the Proceedings of the IEEE/CVF Conference on Computer Vision and Pattern Recognition, 2024; pp. 11566-11578.
45. Li, B.; Li, Y.; Eliceiri, K.W. Dual-stream multiple instance learning network for whole slide image classification with self-supervised contrastive learning. In Proceedings of the Proceedings of the IEEE/CVF conference on computer vision and pattern recognition, 2021; pp. 14318-14328.
46. Nakanishi, R.; Morooka, K.i.; Omori, K.; Toyota, S.; Tanaka, Y.; Hasuda, H.; Koga, N.; Nonaka, K.; Hu, Q.; Nakaji, Y. Artificial Intelligence-Based Prediction of Recurrence after Curative Resection for Colorectal Cancer from Digital Pathological Images: R. Nakanishi et al. *Annals of Surgical Oncology* **2023**, *30*, 3506-3514.
47. Tang, W.; Zhou, F.; Huang, S.; Zhu, X.; Zhang, Y.; Liu, B. Feature re-embedding: Towards foundation model-level performance in computational pathology. In Proceedings of the Proceedings of the IEEE/CVF conference on computer vision and pattern recognition, 2024; pp. 11343-11352.
48. Sajjad, U.; Akbar, A.R.; Su, Z.; Knight, D.; Frankel, W.L.; Gurcan, M.N.; Chen, W.; Niazi, M.K.K. Morphology-Aware Prognostic model for Five-Year Survival Prediction in Colorectal Cancer from H&E Whole Slide Images. *arXiv preprint arXiv:2510.14800* **2025**.
49. Dosovitskiy, A.; Beyer, L.; Kolesnikov, A.; Weissenborn, D.; Zhai, X.; Unterthiner, T.; Dehghani, M.; Minderer, M.; Heigold, G.; Gelly, S. An image is worth 16x16 words: Transformers for image recognition at scale. *arXiv preprint arXiv:2010.11929* **2020**.
50. Loshchilov, I.; Hutter, F. Decoupled weight decay regularization. *arXiv preprint arXiv:1711.05101* **2017**.
51. Krogh, A.; Hertz, J. A simple weight decay can improve generalization. *Advances in neural information processing systems* **1991**, *4*.
52. Loshchilov, I.; Hutter, F. Sgdr: Stochastic gradient descent with warm restarts. *arXiv preprint arXiv:1608.03983* **2016**.
53. Wang, Z.; Wang, P.; Liu, K.; Wang, P.; Fu, Y.; Lu, C.-T.; Aggarwal, C.C.; Pei, J.; Zhou, Y. A comprehensive survey on data augmentation. *IEEE Transactions on Knowledge and Data Engineering* **2025**.
54. Kingma, D.P.; Ba, J. Adam: A method for stochastic optimization. *arXiv preprint arXiv:1412.6980* **2014**.
55. Yuan, W.; Chen, Y.; Zhu, B.; Yang, S.; Zhang, J.; Mao, N.; Xiang, J.; Li, Y.; Ji, Y.; Luo, X. Pancancer outcome prediction via a unified weakly supervised deep learning model. *Signal transduction and targeted therapy* **2025**, *10*, 285.

# Supplementary Tables and Figures:

**Table S1.** Demographic and clinical characteristics of the Alliance/CALGB 89803 cohort. The cohort comprises 424 stage III colorectal cancer patients with 431 whole-slide images, stratified into deceased (n=103) and surviving (n=321) groups based on five-year follow-up.

| Alliance Cohort Characteristics | | |
|---|---|---|
| Number of slides | | 431 |
| Number of patients | | 424 |
| Mean age (years) | | 60.47 |
| Median household income (USD) | | 43194.59 |
| Race | White | 398 |
| | Black | 33 |
| Sex | Male | 240 |
| | Female | 191 |
| Treatment | 5FU/LV | 219 |
| | CPT-11/5FU/LV | 212 |
| Zubrod performance scale | 0 | 328 |
| | 1 | 98 |
| | 2 | 2 |
| Tumor location | Cecum | 101 |
| | Ascending colon | 64 |
| | Hepatic flexure | 28 |
| | Transverse colon | 46 |
| | Splenic flexure | 19 |
| | Descending colon | 19 |
| | Sigmoid colon | 149 |
| Grade | I | 20 |
| | II | 300 |
| | III | 108 |
| | IV | 0 |
| Stage | I | 6 |
| | II | 42 |
| | III | 350 |
| | IV | 8 |
| | V | 20 |
| Small blood/lymphatic vessel invasion | No | 285 |
| | Yes | 138 |
| Extramural vascular invasion | No | 431 |

**Table S2.** Dataset-wise distribution of morphological classes. Number of patches used for training and validation of MorphDistill from CRC-100K, STARC-9, and SPIDER-Colorectal datasets. Each dataset provides complementary tissue types, together spanning 18 distinct morphological classes relevant to colorectal cancer histopathology.

| Dataset | Morphological Class | Training | Validation |
|---|---|---|---|
| CRC-100K [34] | Adipose (ADI) | 10,407 | 1,338 |
| | Background (BACK) | 10,566 | 641 |
| | Debris (DEB) | 11,512 | 629 |
| | Lymphocytes (LYM) | 11,557 | 1,154 |
| | Mucus (MUC) | 8,896 | 771 |
| | Smooth muscle (MUS) | 13,536 | 593 |
| | Normal colon mucosa (NORM) | 8,763 | 476 |
| | Cancer-associated stroma (STR) | 10,446 | 421 |
| | Colorectal adenocarcinoma epithelium (TUM) | 14,317 | 1,157 |
| | **Total** | **100,000** | **7,180** |
| | Adipose (ADI) | 70,000 | 10,300 |
| | Lymphocytes (LYM) | 70,000 | 10,500 |
| | Smooth muscle (MUS) | 70,000 | 10,100 |
| | Fibrous connective tissue (FCT) | 70,000 | 10,400 |
| | Mucus (MUC) | 70,000 | 10,500 |
| | Necrotic / cellular debris (NCS) | 70,000 | 10,100 |
| STARC-9 | Blood (BLD) | 70,000 | 9,500 |
| [35] | Tumor epithelium (TUM) | 70,000 | 10,400 |
| | Normal (NOR) | 70,000 | 10,200 |
| | **Total** | **630,000** | **92,000** |
| | Adenocarcinoma high grade | 5,039 | 1,260 |
| | Adenocarcinoma low grade | 4,853 | 1,213 |
| | Adenoma high grade | 4,394 | 1,099 |
| | Adenoma low grade | 4,554 | 1,139 |
| | Hyperplastic polyp | 4,714 | 1,179 |
| | Sessile serrated lesion | 3,994 | 999 |
| | Inflammation | 4,418 | 1,105 |
| | Necrosis | 4,385 | 1,096 |
| | Mucus | 4,569 | 1,142 |
| | Muscle | 4,693 | 1,173 |
| SPIDER- | Fat | 4,865 | 1,213 |
| Colorectal [36] | Stroma healthy | 6,400 | 1,601 |
| | Vessels | 4,865 | 1,260 |
| | **Total** | **61,743** | **15,479** |

**Table S3.** Combined morphological classes across all datasets. Patch distribution for training and validation after harmonizing class labels from CRC-100K, STARC-9, and SPIDER-Colorectal into 18 unified morphological categories. This comprehensive collection enables supervised contrastive regularization during Stage I training.

| **Dataset** | **Morphological Class** | **Training** | **Validation** |
|---|---|---|---|
| | Adenocarcinoma high grade | 47,197 | 7,038 |
| | Adenocarcinoma low grade | 46,406 | 6,992 |
| | Adenoma high grade | 4,394 | 1,099 |
| | Adenoma low grade | 4,554 | 1,139 |
| | Hyperplastic Polyp | 4,714 | 1,179 |
| | Sessile Serrated | 3,994 | 999 |
| | Inflammation | 4,418 | 1,105 |
| | Necrosis (NCS) | 74,385 | 11,196 |
| | Mucus (MUC) | 83,465 | 12,413 |
| | Lymphocyte (LYM) | 81,557 | 11,654 |
| | Adipose (ADI) | 85,272 | 12,851 |
| | Muscle (MUS) | 88,229 | 11,866 |
| | Stroma (STR) | 86,846 | 12,422 |
| | Vessels | 74,865 | 10,760 |
| | Normal (NORM) | 78,763 | 10,676 |
| | Fibrosis (FCT) | 70,000 | 10,400 |
| | Background (BACK) | 10,566 | 641 |
| Combined | Debris (DEB) | 81,512 | 10,729 |
| | **Total** | **791,743** | **114,659** |

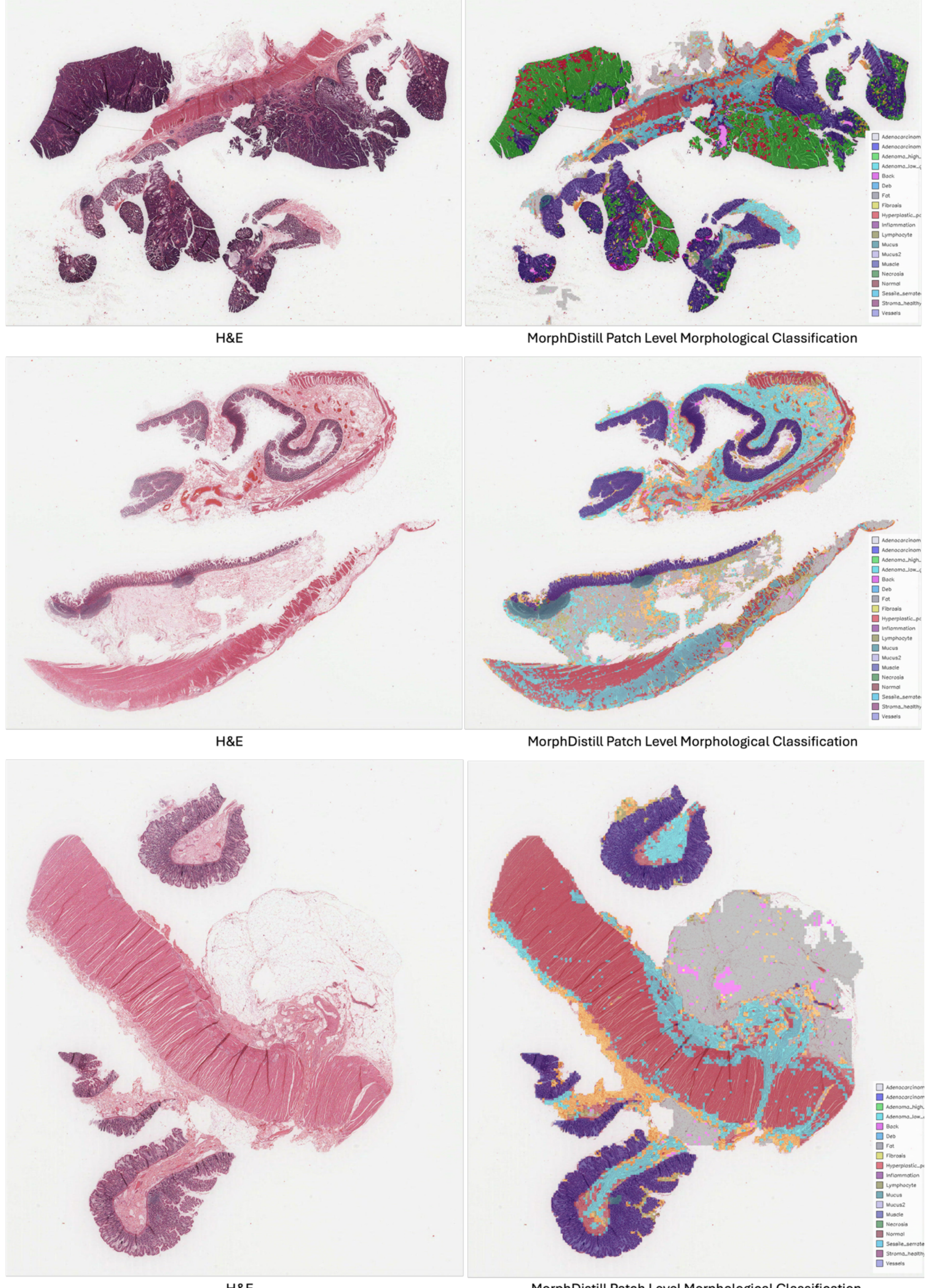

**Figure S1:** Morphological inference by the MorphDistill encoder on a whole-slide image. (Left) Original H&E-stained slide from the Alliance cohort. (Right) Patch-level tissue classification map generated by the frozen MorphDistill encoder, with colors indicating different tissue types. The clear delineation of diverse morphological classes demonstrates the encoder's ability to preserve CRC-relevant tissue semantics after unifying knowledge from ten foundation models (best viewed in color).

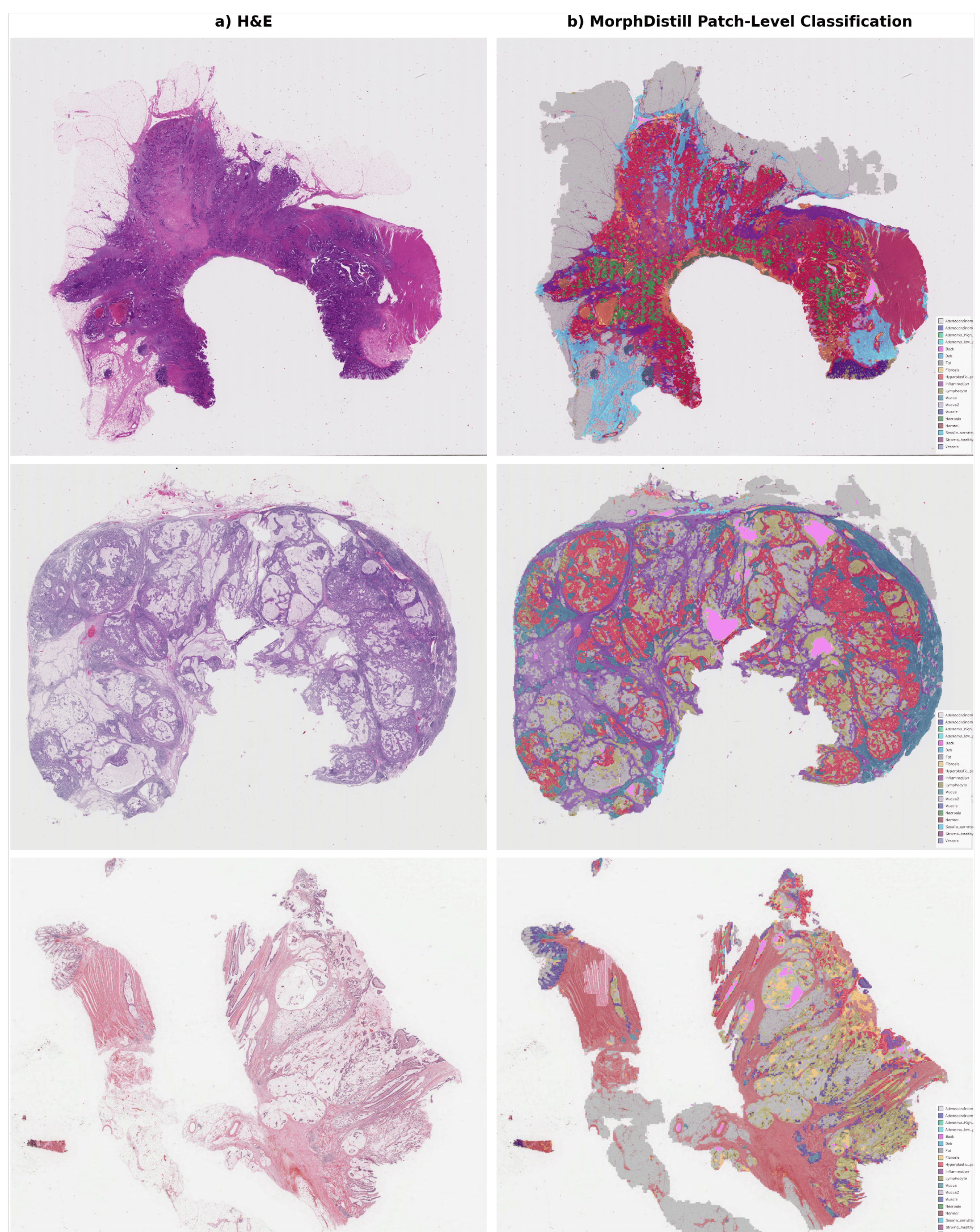

**Figure S2:** Morphological inference by the MorphDistill encoder on a whole-slide image. (Left) Original H&E-stained slide from the Alliance cohort. (Right) Patch-level tissue classification map generated by the frozen MorphDistill encoder, with colors indicating different tissue types. The clear delineation of diverse morphological classes demonstrates the encoder's ability to preserve CRC-relevant tissue semantics after unifying knowledge from ten foundation models (best viewed in color).

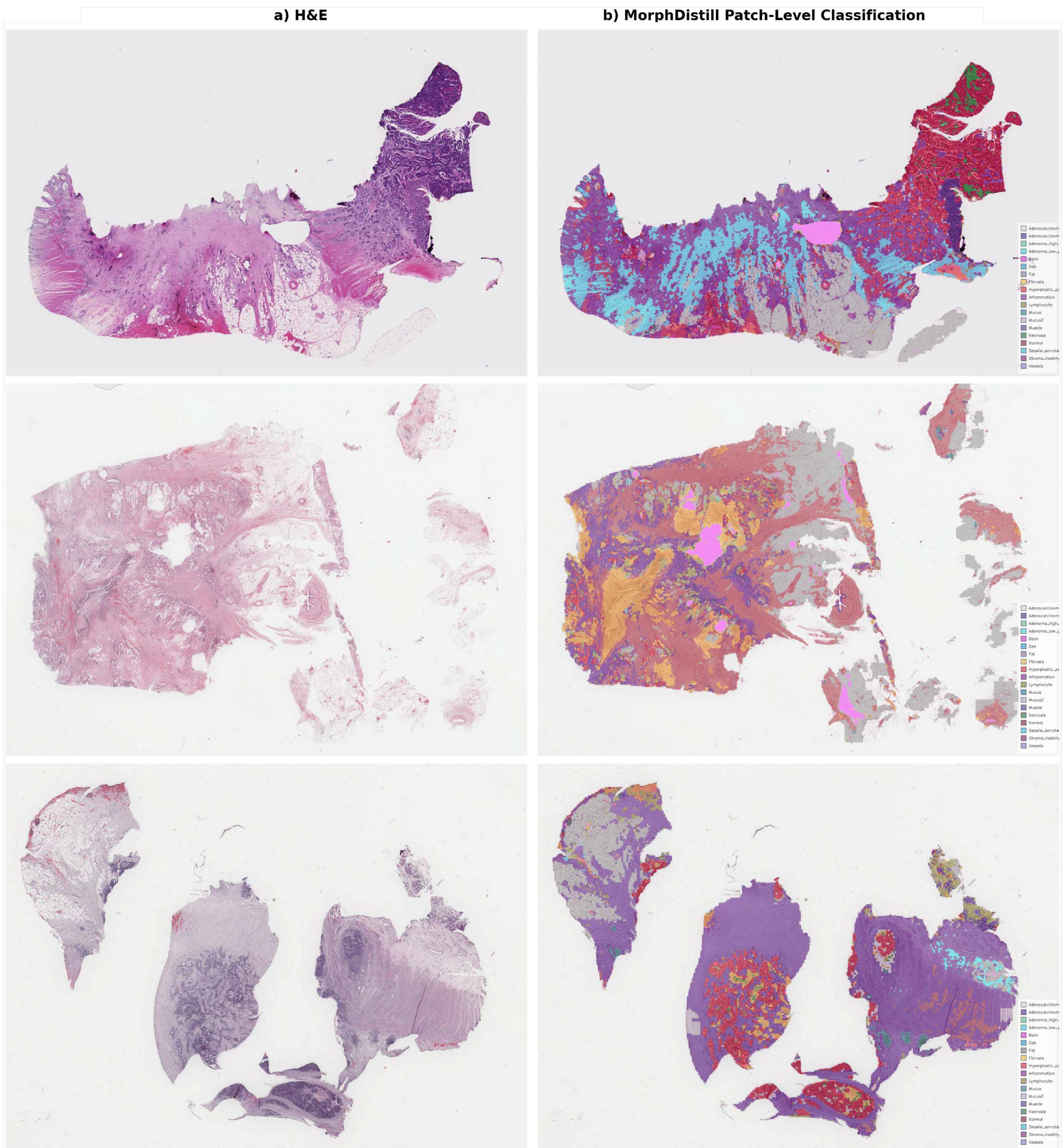

**Figure S3:** Morphological inference by the MorphDistill encoder on a whole-slide image. (Left) Original H&E-stained slide from the Alliance cohort. (Right) Patch-level tissue classification map generated by the frozen MorphDistill encoder, with colors indicating different tissue types. The clear delineation of diverse morphological classes demonstrates the encoder's ability to preserve CRC-relevant tissue semantics after unifying knowledge from ten foundation models (best viewed in color).

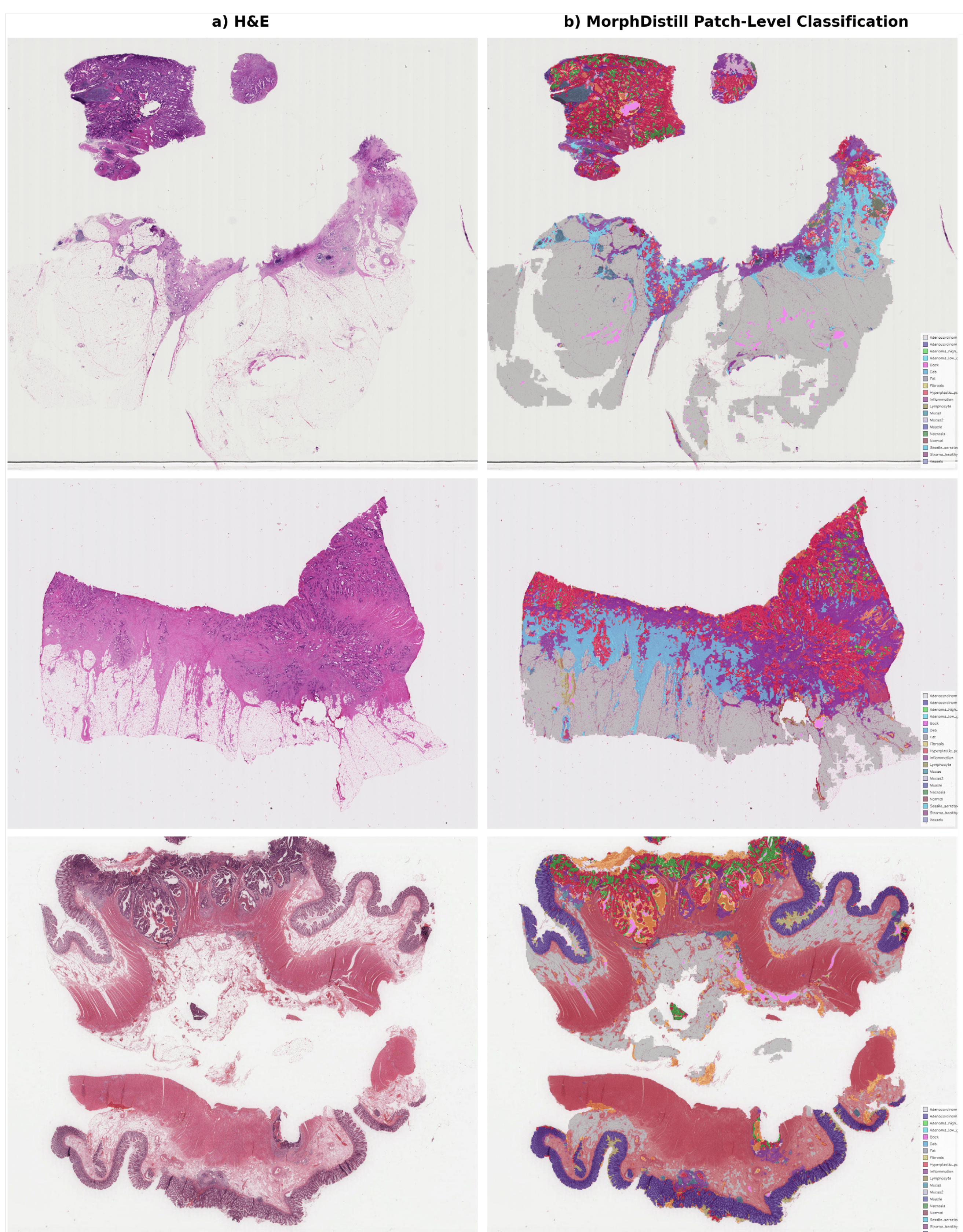

**Figure S4:** Morphological inference by the MorphDistill encoder on a whole-slide image. (Left) Original H&E-stained slide from the Alliance cohort. (Right) Patch-level tissue classification map generated by the frozen MorphDistill encoder, with colors indicating different tissue types. The clear delineation of diverse morphological classes demonstrates the encoder's ability to preserve CRC-relevant tissue semantics after unifying knowledge from ten foundation models (best viewed in color).

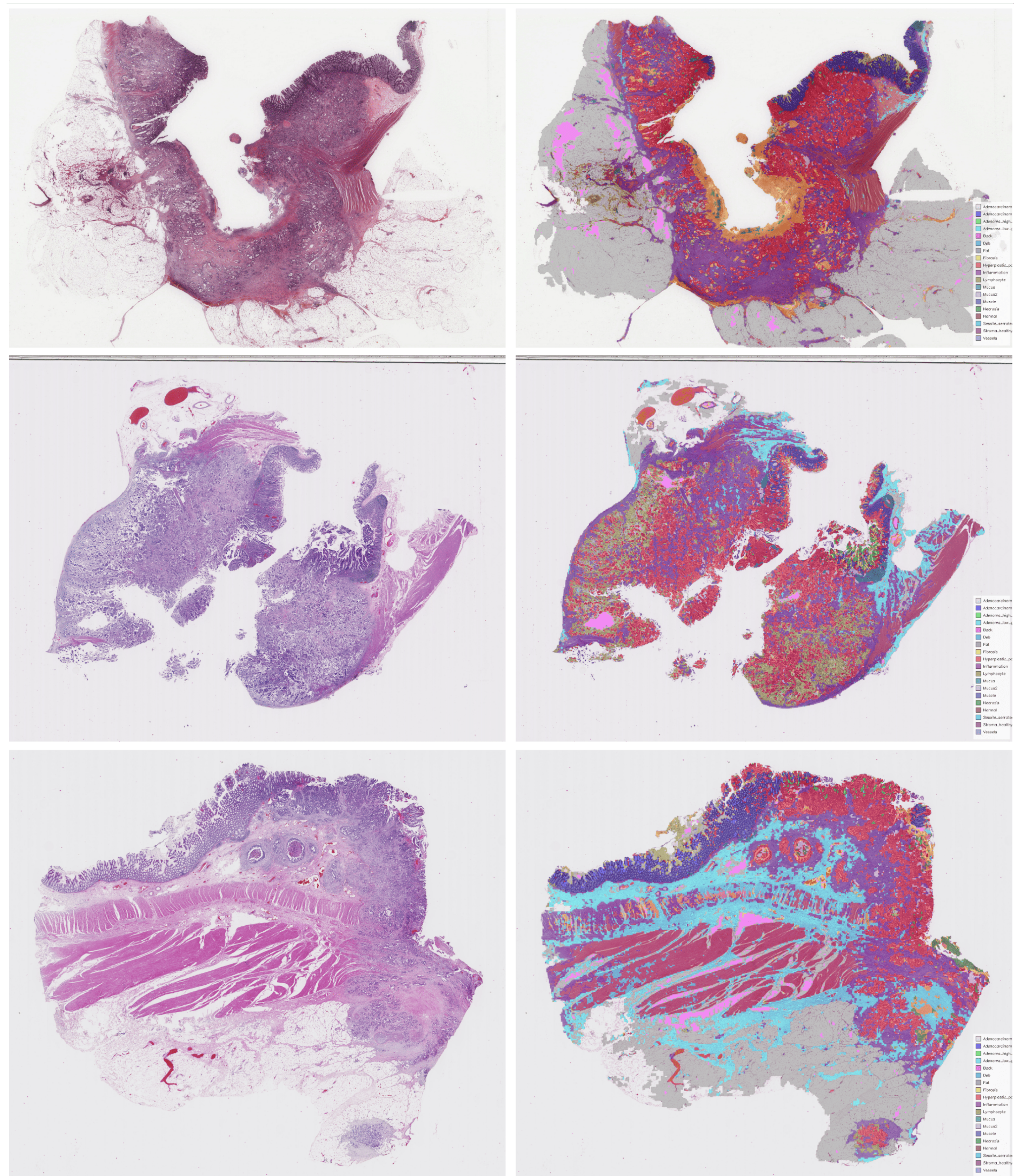

**Figure S5:** Morphological inference by the MorphDistill encoder on a whole-slide image. (Left) Original H&E-stained slide from the Alliance cohort. (Right) Patch-level tissue classification map generated by the frozen MorphDistill encoder, with colors indicating different tissue types. The clear delineation of diverse morphological classes demonstrates the encoder's ability to preserve CRC-relevant tissue semantics after unifying knowledge from ten foundation models (best viewed in color).

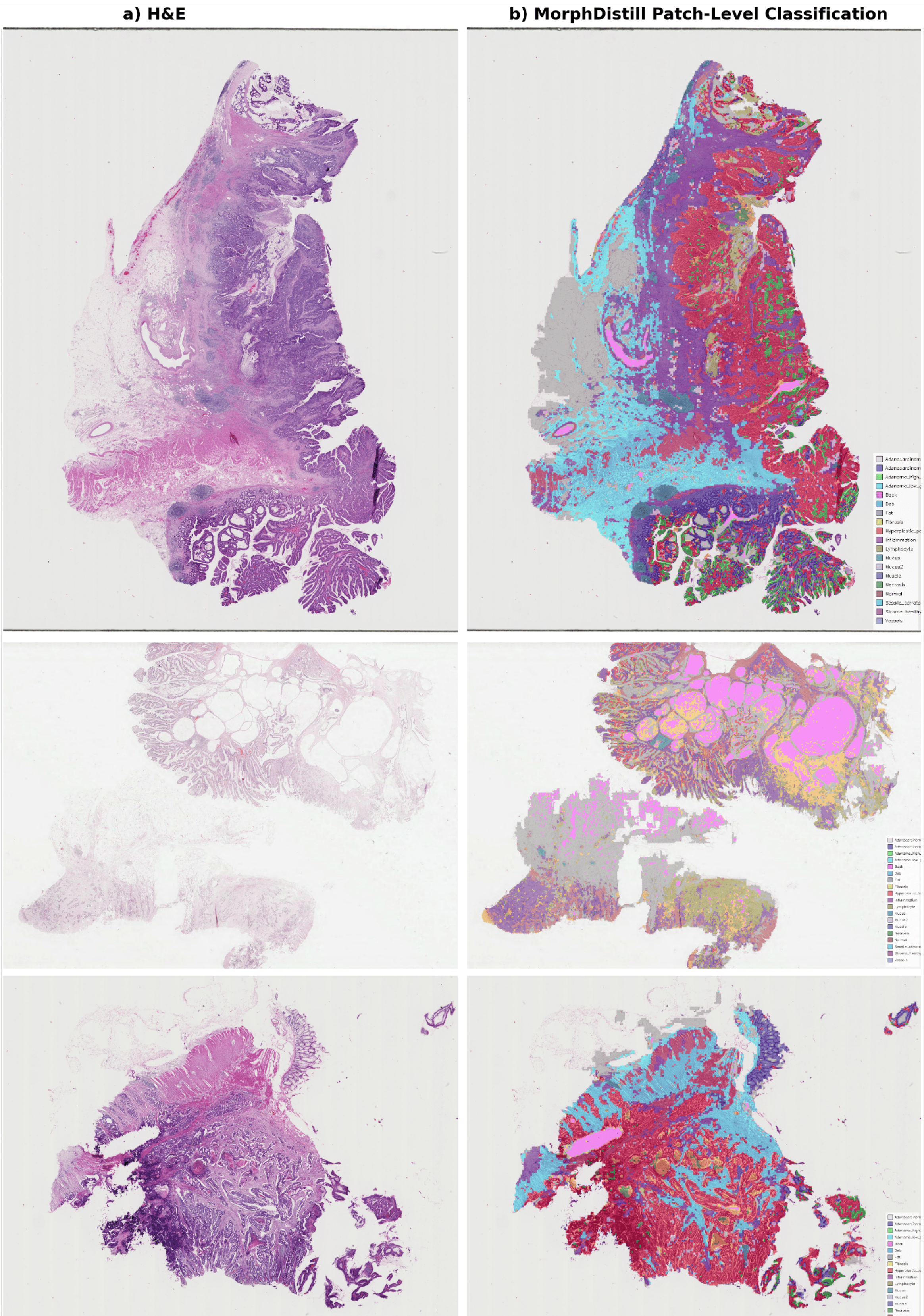

**Figure S6:** Morphological inference by the MorphDistill encoder on a whole-slide image. (Left) Original H&E-stained slide from the Alliance cohort. (Right) Patch-level tissue classification map generated by the frozen MorphDistill encoder, with colors indicating different tissue types. The clear delineation of diverse morphological classes demonstrates the encoder's ability to preserve CRC-relevant tissue semantics after unifying knowledge from ten foundation models (best viewed in color).

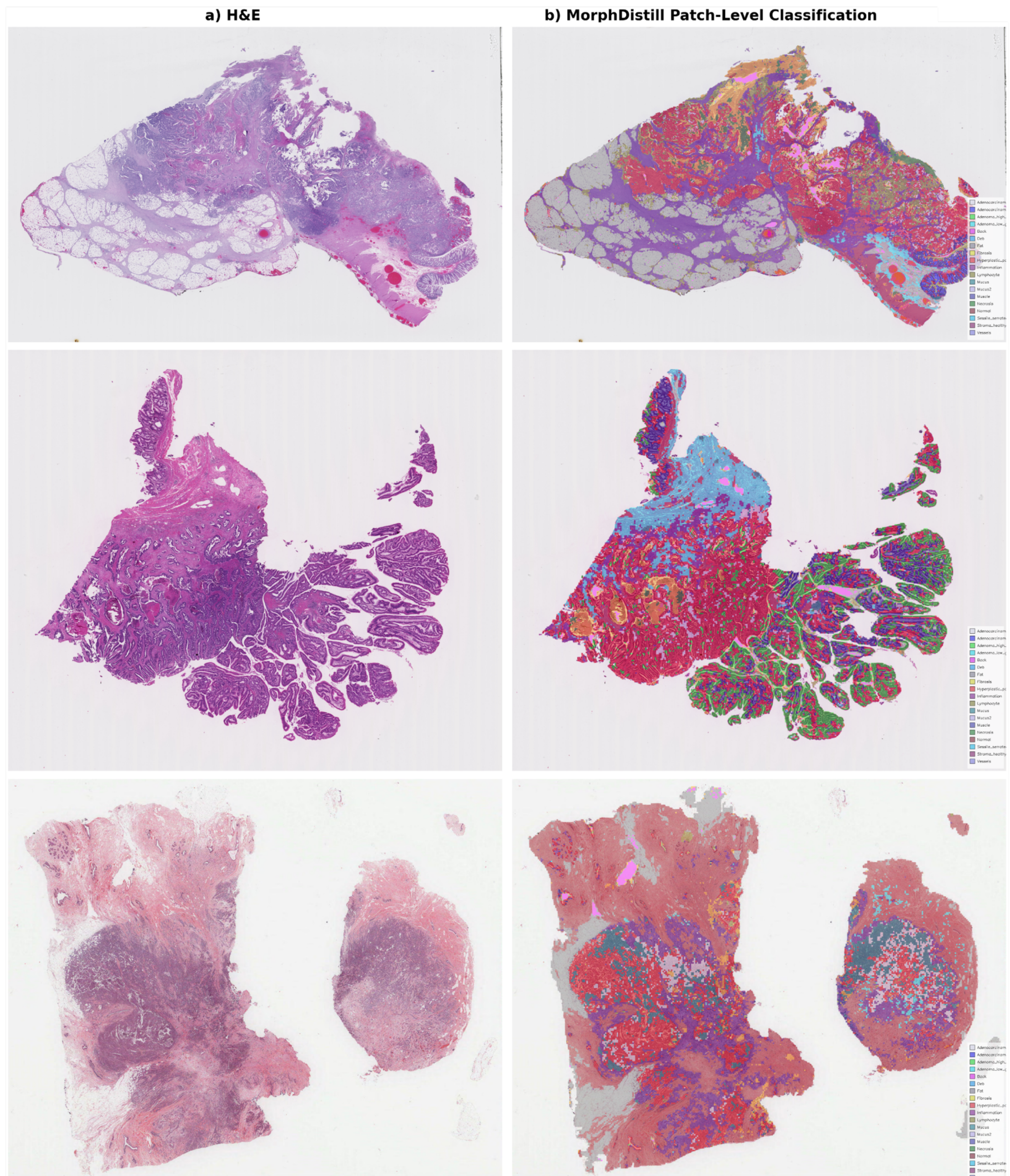

**Figure S7:** Morphological inference by the MorphDistill encoder on a whole-slide image. (Left) Original H&E-stained slide from the Alliance cohort. (Right) Patch-level tissue classification map generated by the frozen MorphDistill encoder, with colors indicating different tissue types. The clear delineation of diverse morphological classes demonstrates the encoder's ability to preserve CRC-relevant tissue semantics after unifying knowledge from ten foundation models (best viewed in color).

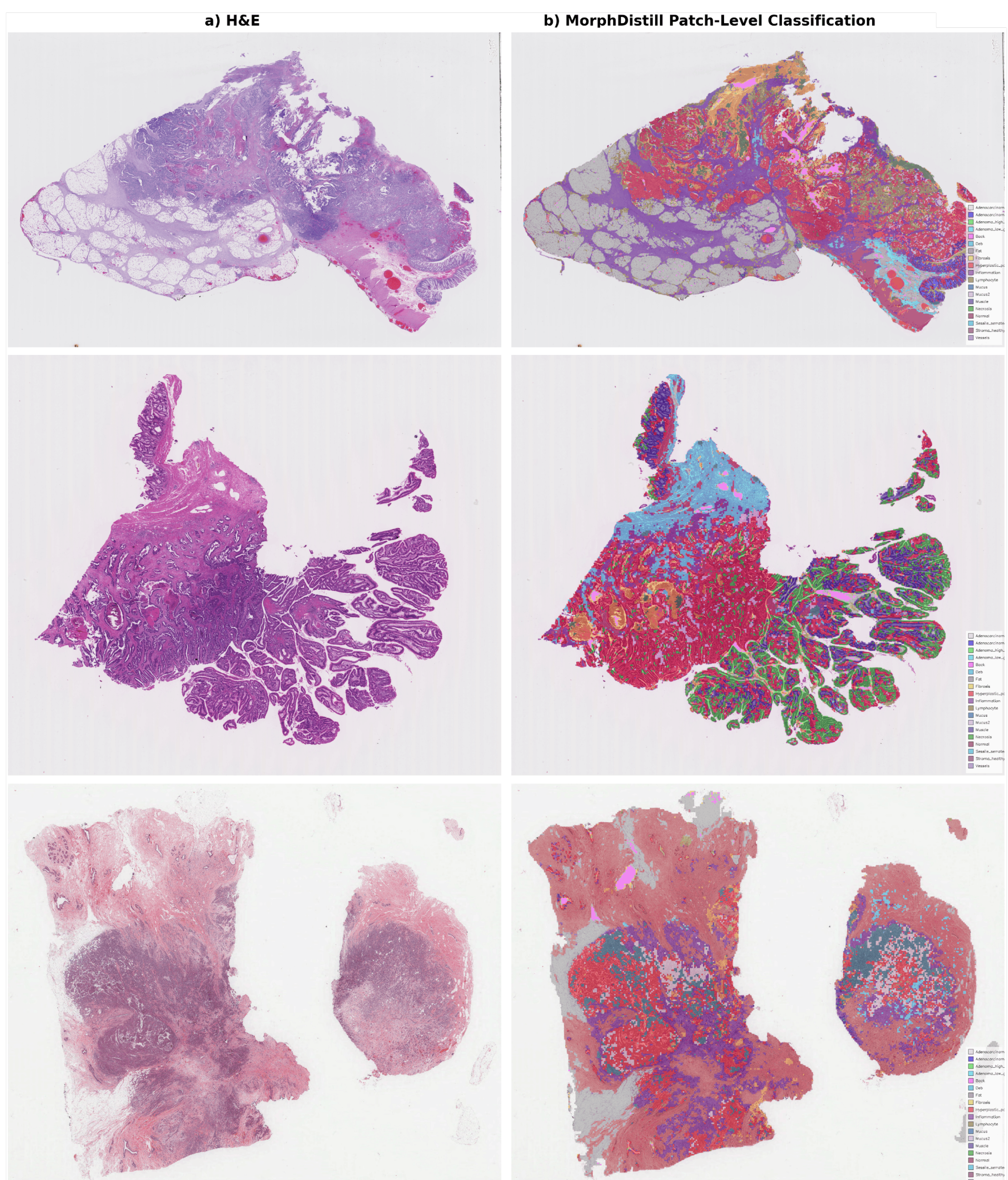

**Figure S8:** Morphological inference by the MorphDistill encoder on a whole-slide image. (Left) Original H&E-stained slide from the Alliance cohort. (Right) Patch-level tissue classification map generated by the frozen MorphDistill encoder, with colors indicating different tissue types. The clear delineation of diverse morphological classes demonstrates the encoder's ability to preserve CRC-relevant tissue semantics after unifying knowledge from ten foundation models (best viewed in color).

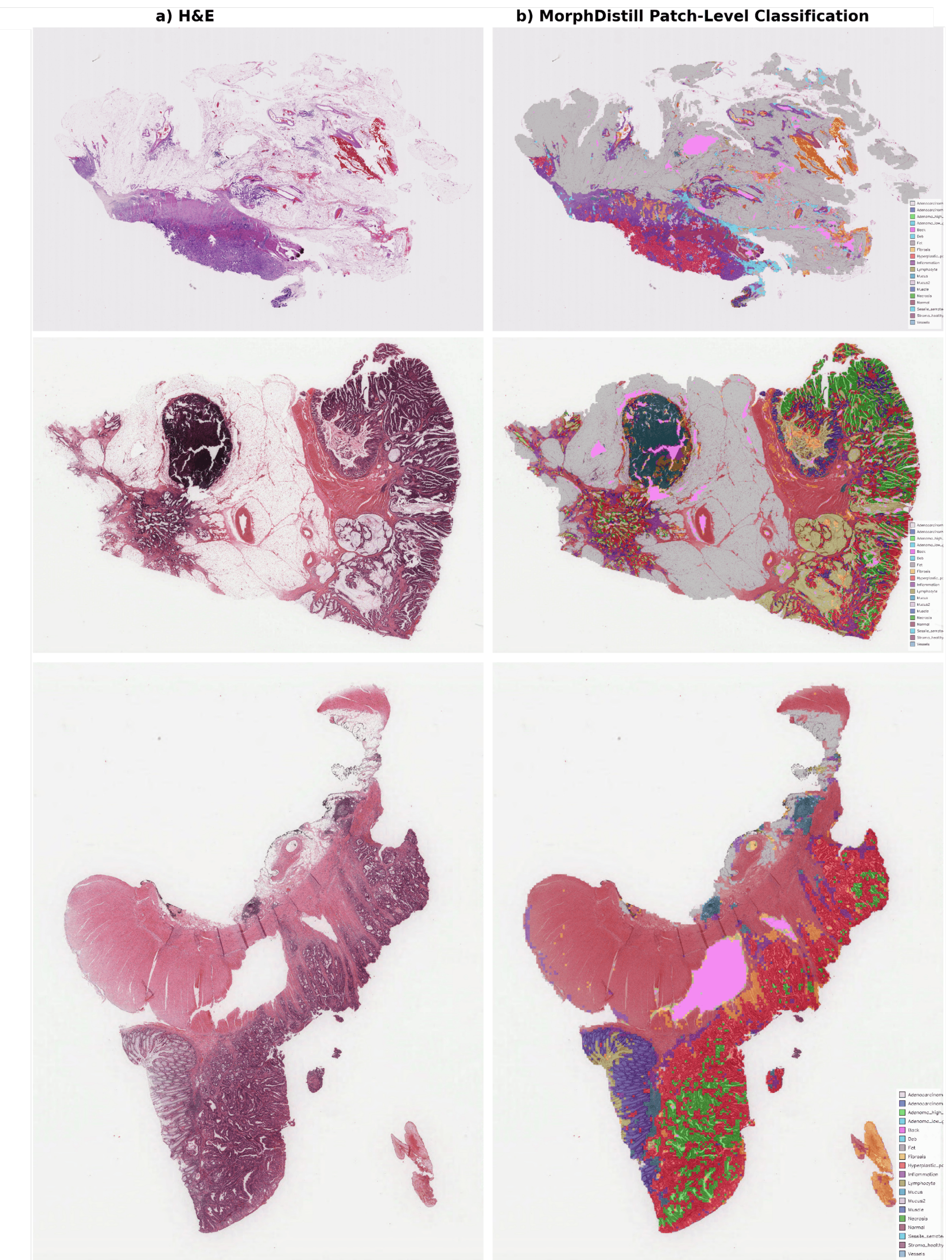

**Figure S9:** Morphological inference by the MorphDistill encoder on a whole-slide image. (Left) Original H&E-stained slide from the Alliance cohort. (Right) Patch-level tissue classification map generated by the frozen MorphDistill encoder, with colors indicating different tissue types. The clear delineation of diverse morphological classes demonstrates the encoder's ability to preserve CRC-relevant tissue semantics after unifying knowledge from ten foundation models (best viewed in color).

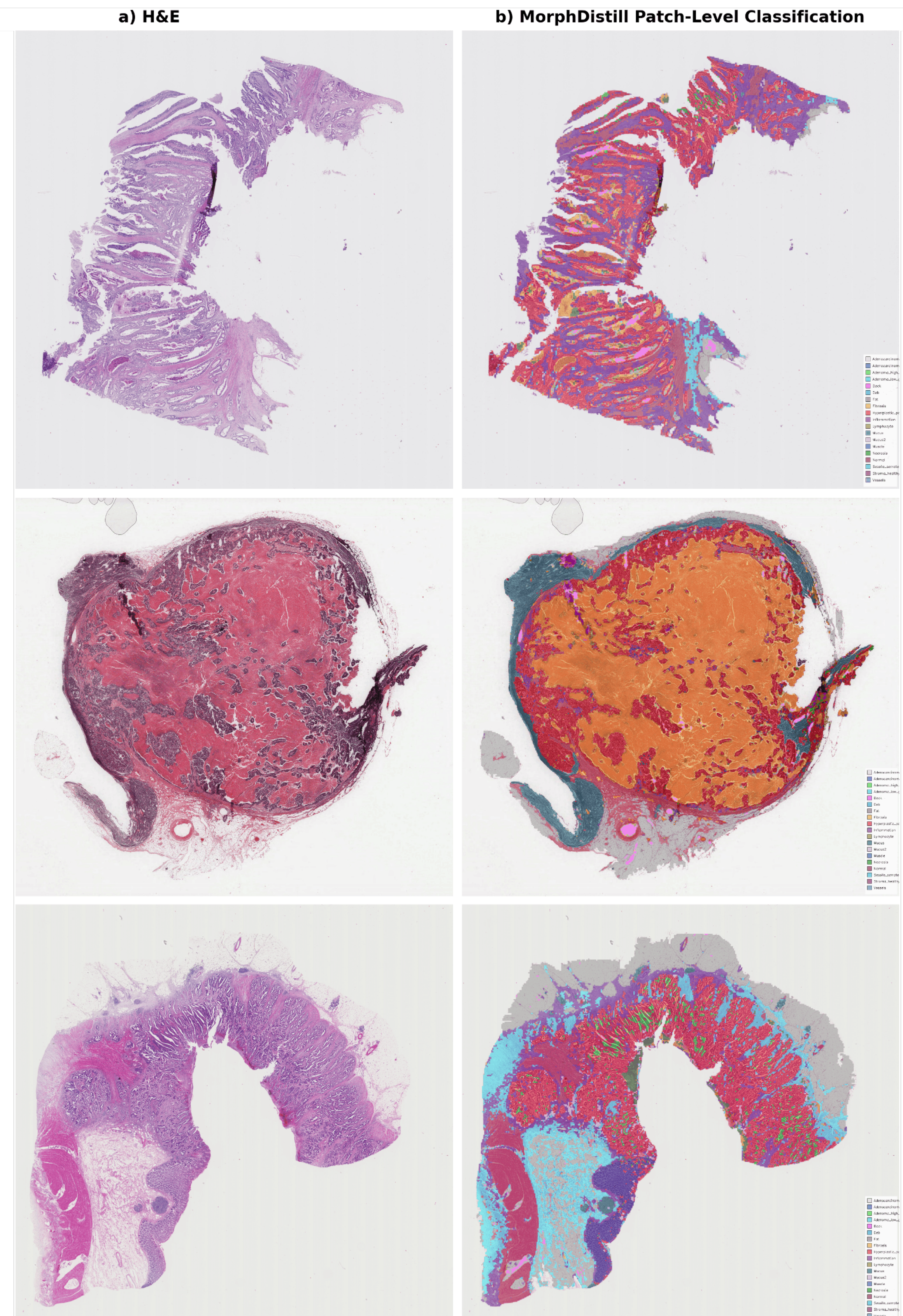

**Figure S10:** Morphological inference by the MorphDistill encoder on a whole-slide image. (Left) Original H&E-stained slide from the Alliance cohort. (Right) Patch-level tissue classification map generated by the frozen MorphDistill encoder, with colors indicating different tissue types. The clear delineation of diverse morphological classes demonstrates the encoder's ability to preserve CRC-relevant tissue semantics after unifying knowledge from ten foundation models (best viewed in color).

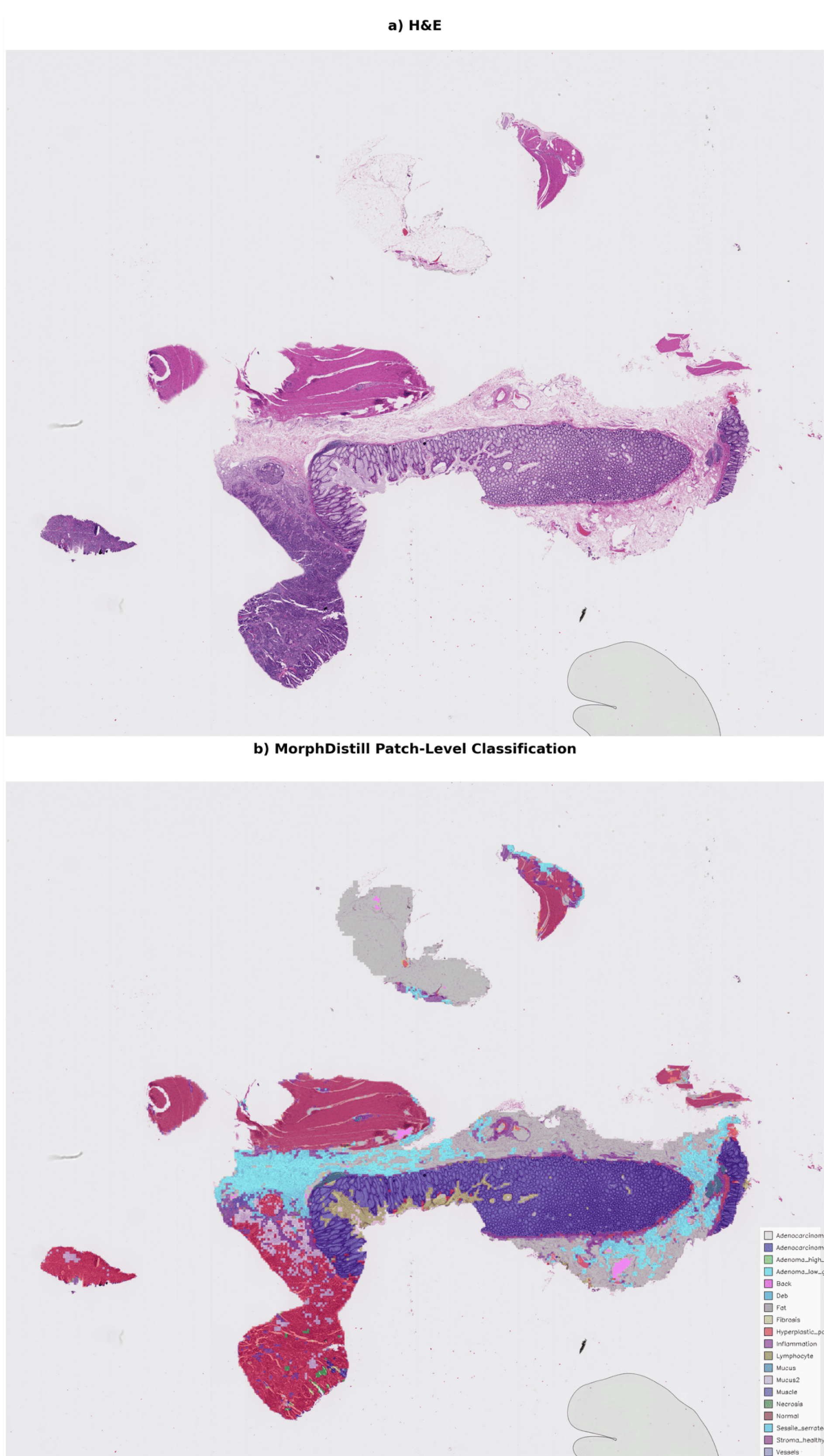

**Figure S11:** Morphological inference by the MorphDistill encoder on a whole-slide image. (a) Original H&E-stained slide from the Alliance cohort. (b) Patch-level tissue classification map generated by the frozen MorphDistill encoder, with colors indicating different tissue types. The clear delineation of diverse morphological classes — including adenocarcinoma, lymphocytes, mucus, muscle, stroma, and vessels, etc — demonstrates the encoder's ability to preserve CRC-relevant tissue semantics after unifying knowledge from ten foundation models (best viewed in color).

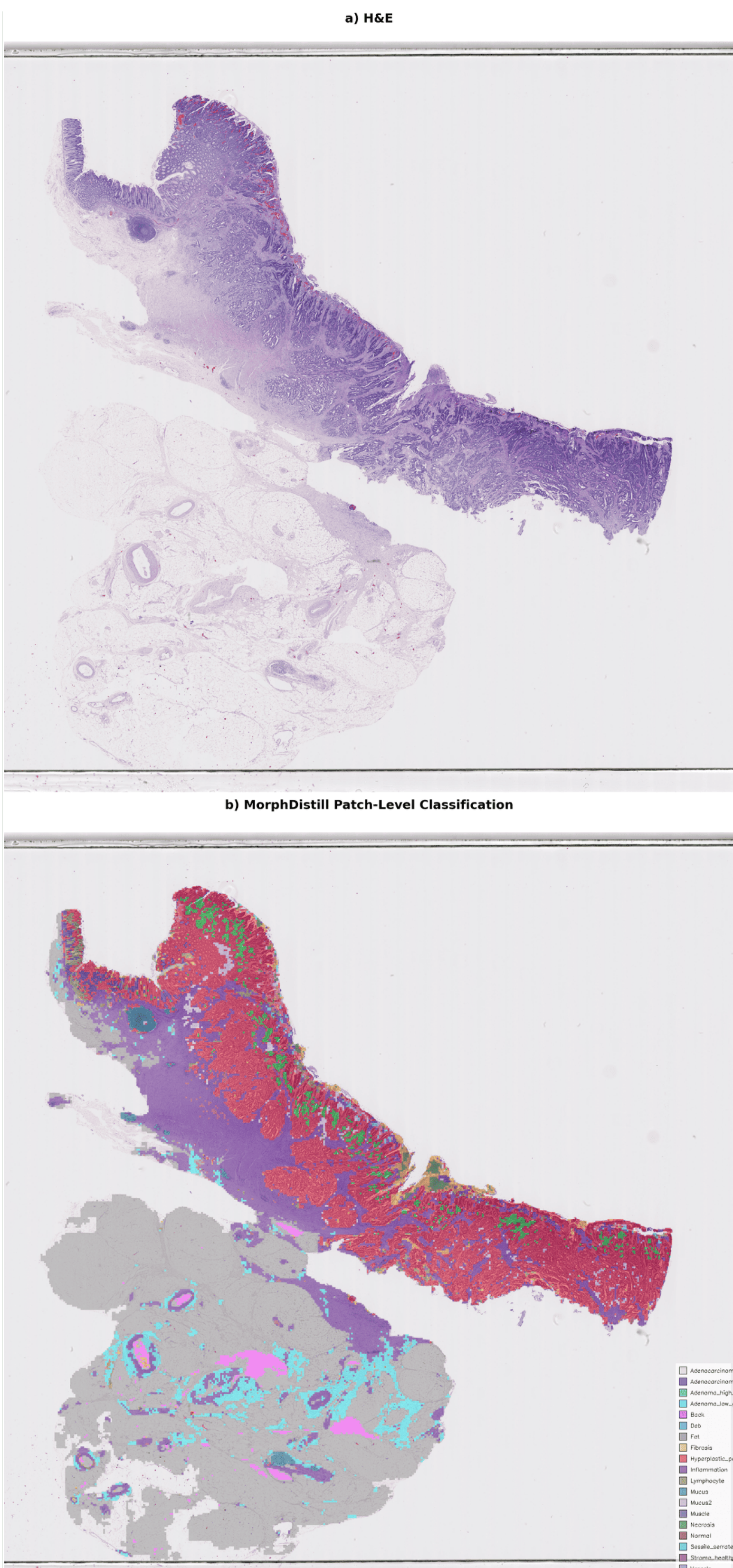

**Figure S13:** Morphological inference by the MorphDistill encoder on a whole-slide image. (a) Original H&E-stained slide from the Alliance cohort. (b) Patch-level tissue classification map generated by the frozen MorphDistill encoder, with colors indicating different tissue types. The clear delineation of diverse morphological classes — including adenocarcinoma, lymphocytes, mucus, muscle, stroma, and vessels, etc — demonstrates the encoder's ability to preserve CRC-relevant tissue semantics after unifying knowledge from ten foundation models (best viewed in color).

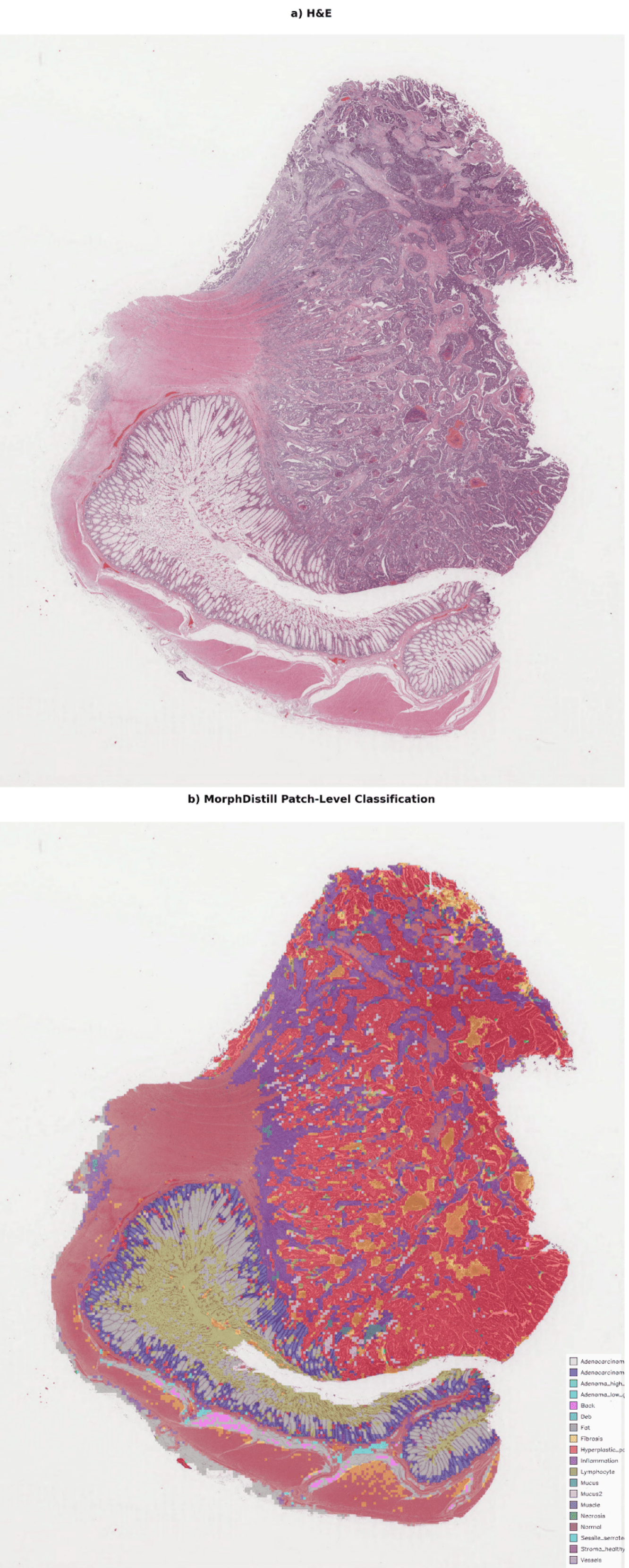

**Figure S14:** Morphological inference by the MorphDistill encoder on a whole-slide image. (a) Original H&E-stained slide from the Alliance cohort. (b) Patch-level tissue classification map generated by the frozen MorphDistill encoder, with colors indicating different tissue types. The clear delineation of diverse morphological classes — including adenocarcinoma, lymphocytes, mucus, muscle, stroma, and vessels, etc — demonstrates the encoder's ability to preserve CRC-relevant tissue semantics after unifying knowledge from ten foundation models (best viewed in color).

a) H&E

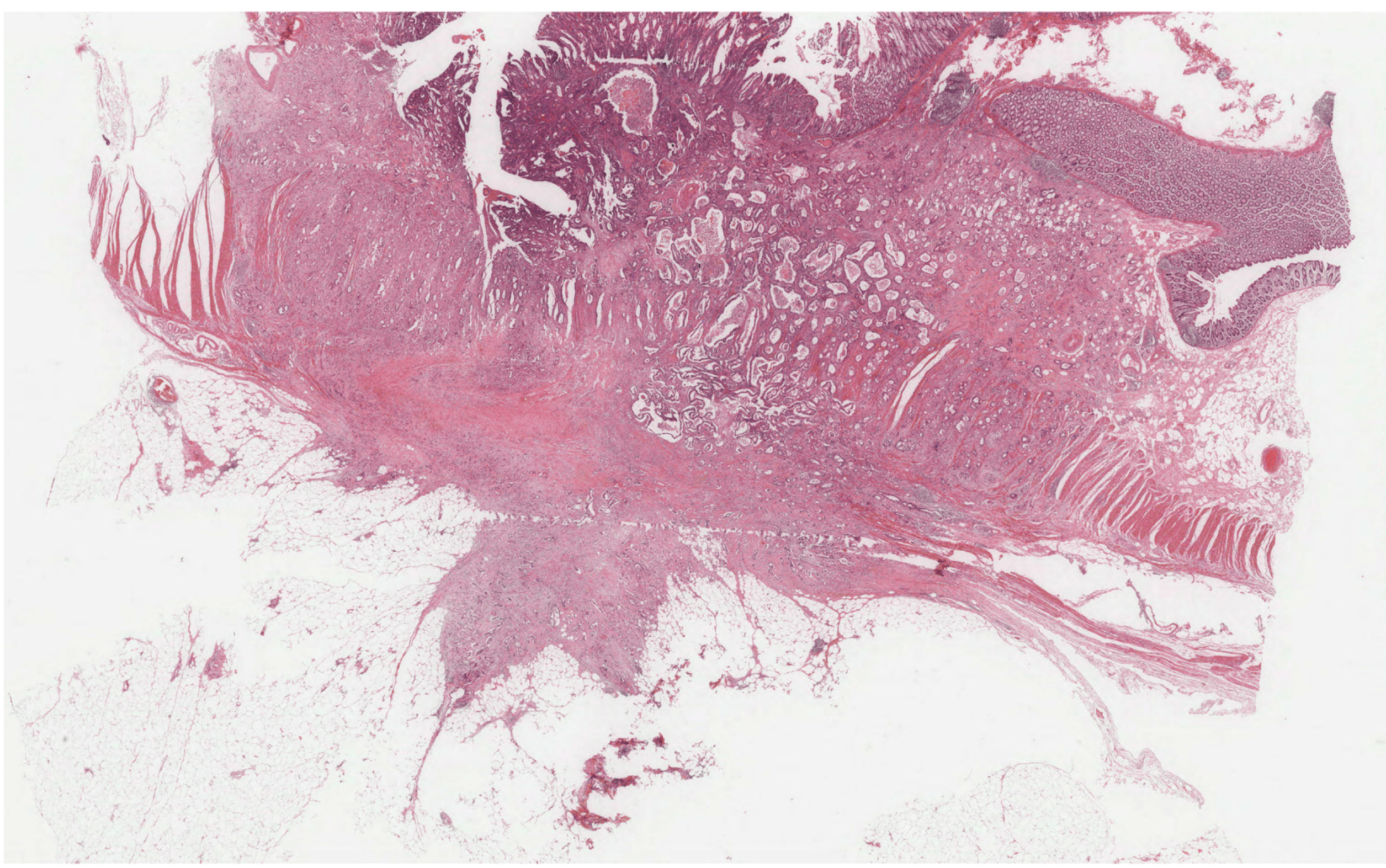

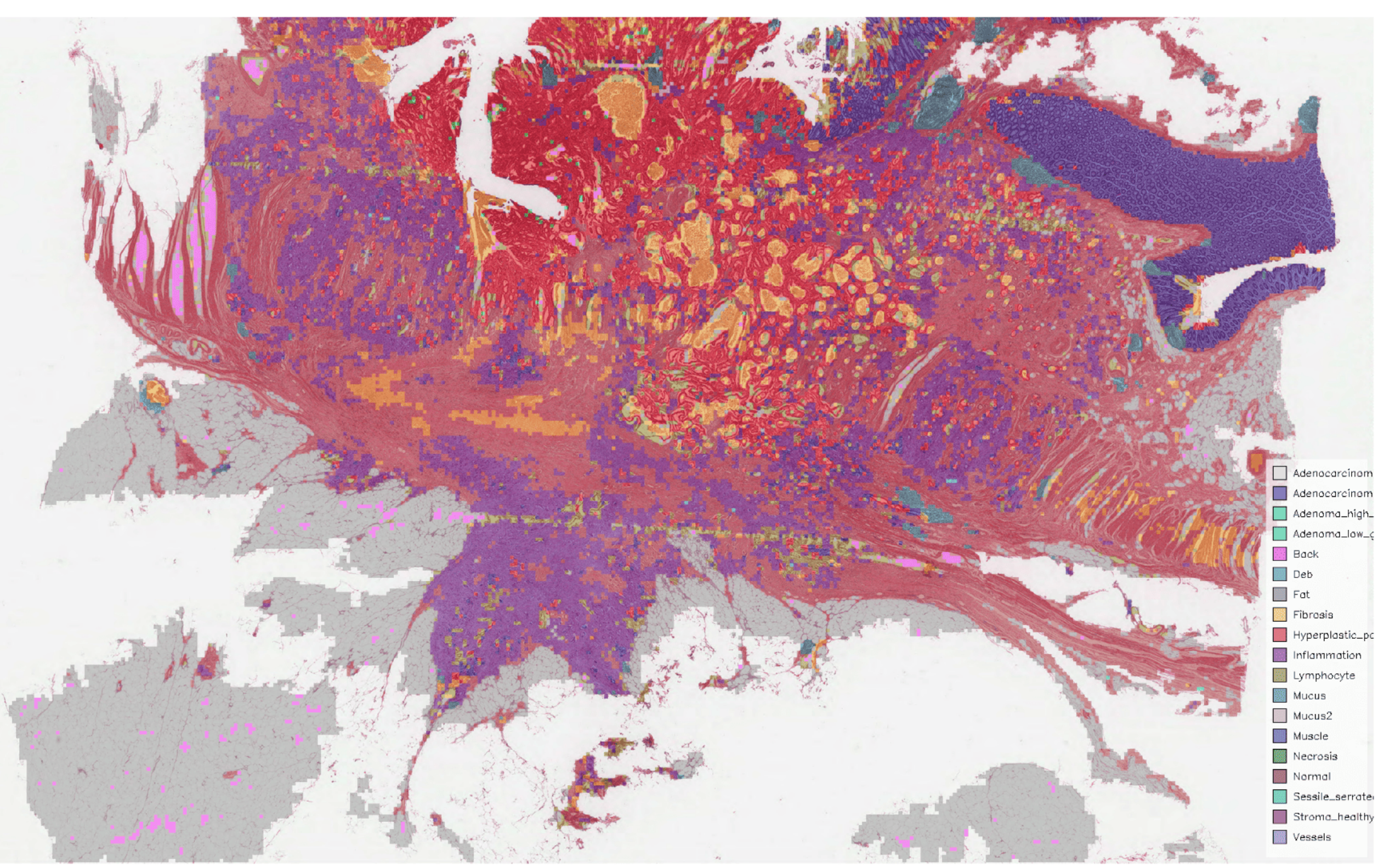

**Figure S15:** Morphological inference by the MorphDistill encoder on a whole-slide image. (a) Original H&E-stained slide from the Alliance cohort. (b) Patch-level tissue classification map generated by the frozen MorphDistill encoder, with colors indicating different tissue types. The clear delineation of diverse morphological classes — including adenocarcinoma, lymphocytes, mucus, muscle, stroma, and vessels, etc — demonstrates the encoder's ability to preserve CRC-relevant tissue semantics after unifying knowledge from ten foundation models (best viewed in color).

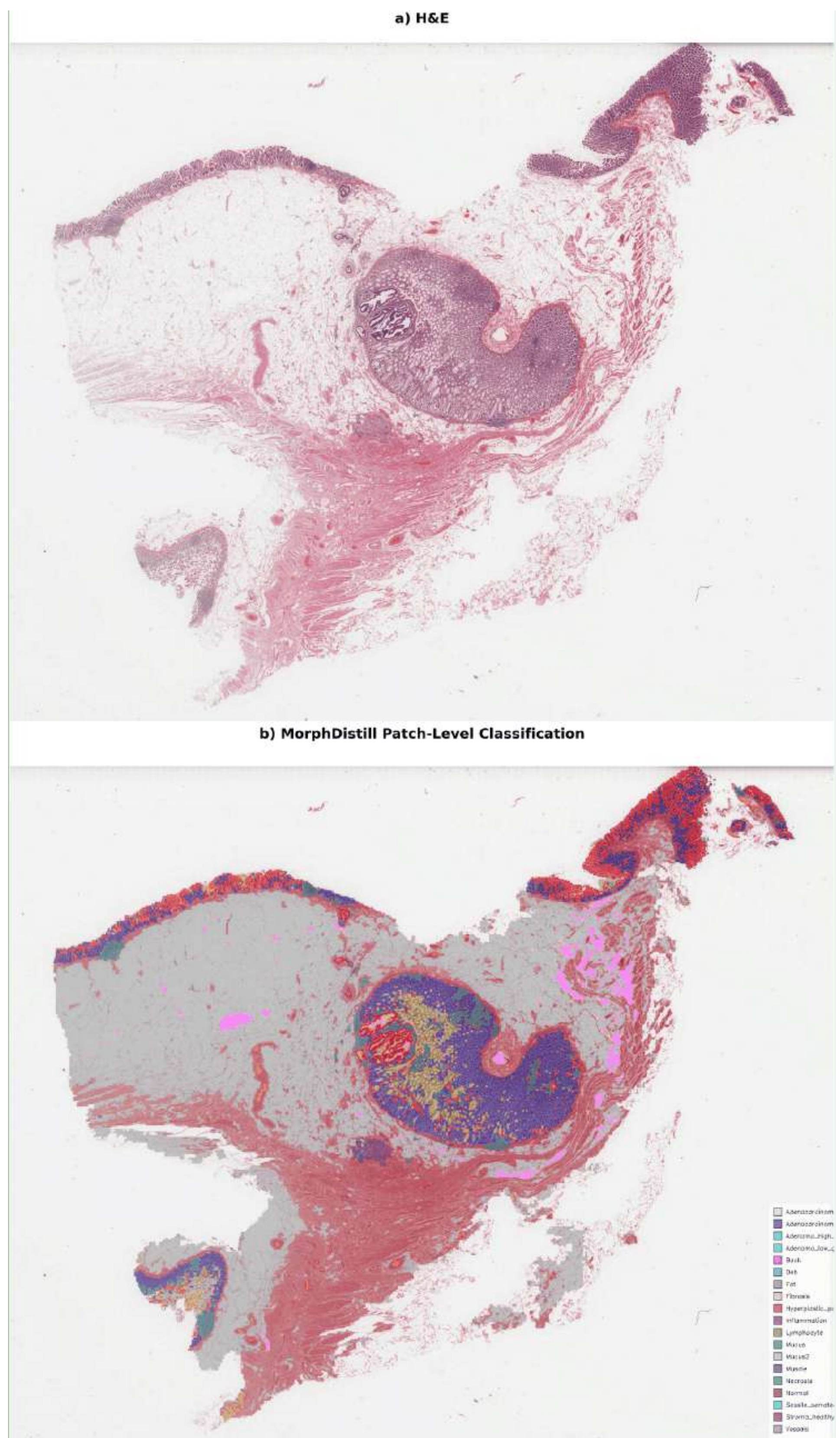

**Figure S16:** Morphological inference by the MorphDistill encoder on a whole-slide image. (a) Original H&E-stained slide from the Alliance cohort. (b) Patch-level tissue classification map generated by the frozen MorphDistill encoder, with colors indicating different tissue types. The clear delineation of diverse morphological classes — including adenocarcinoma, lymphocytes, mucus, muscle, stroma, and vessels etc. — demonstrates the encoder's ability to preserve CRC-relevant tissue semantics after unifying knowledge from ten foundation models (best viewed in color).

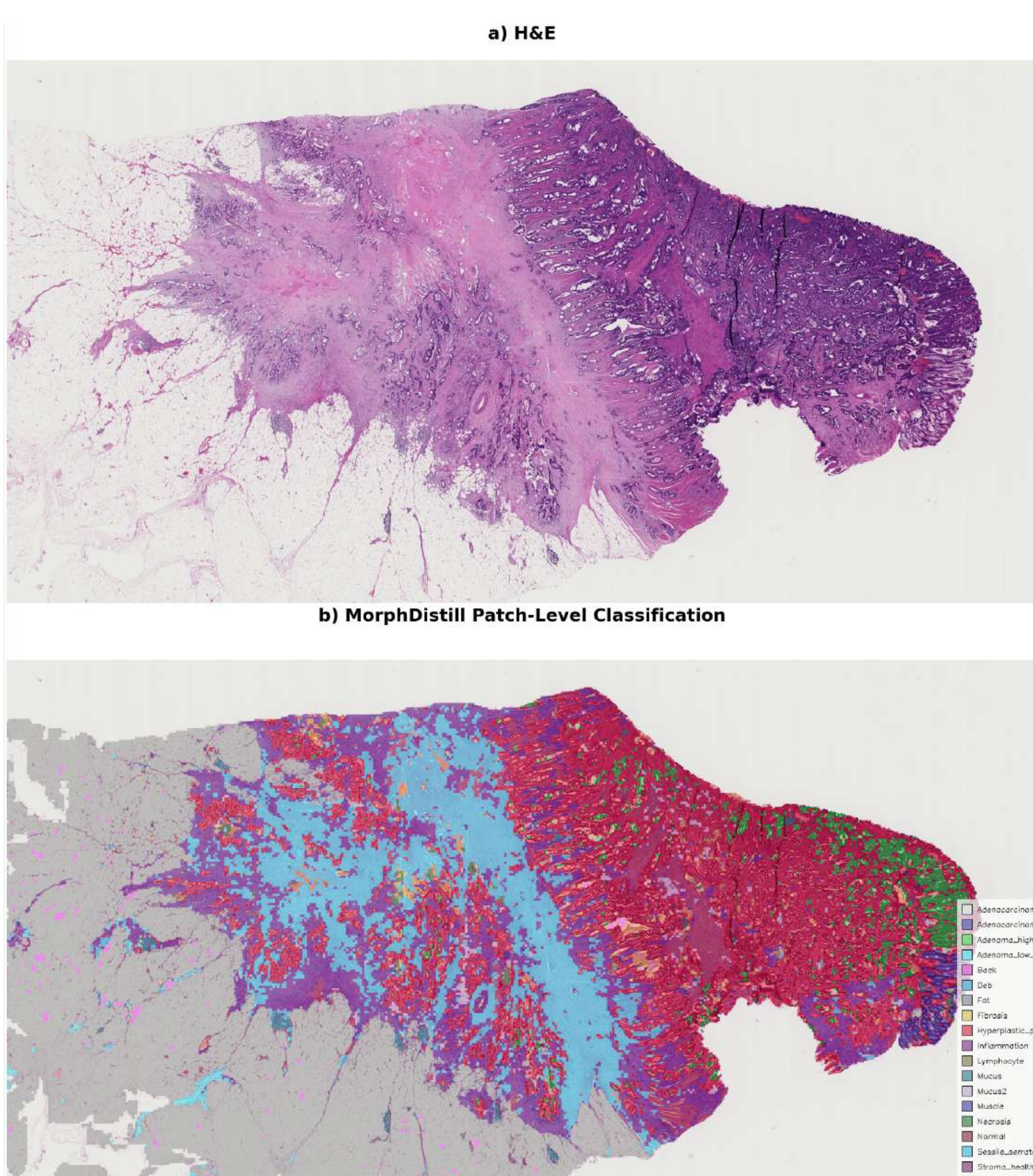

**Figure S17:** Morphological inference by the MorphDistill encoder on a whole-slide image. (a) Original H&E-stained slide from the Alliance cohort. (b) Patch-level tissue classification map generated by the frozen MorphDistill encoder, with colors indicating different tissue types. The clear delineation of diverse morphological classes — including adenocarcinoma, lymphocytes, mucus, muscle, stroma, and vessels etc. — demonstrates the encoder's ability to preserve CRC-relevant tissue semantics after unifying knowledge from ten foundation models (best viewed in color).

**a) H&E**

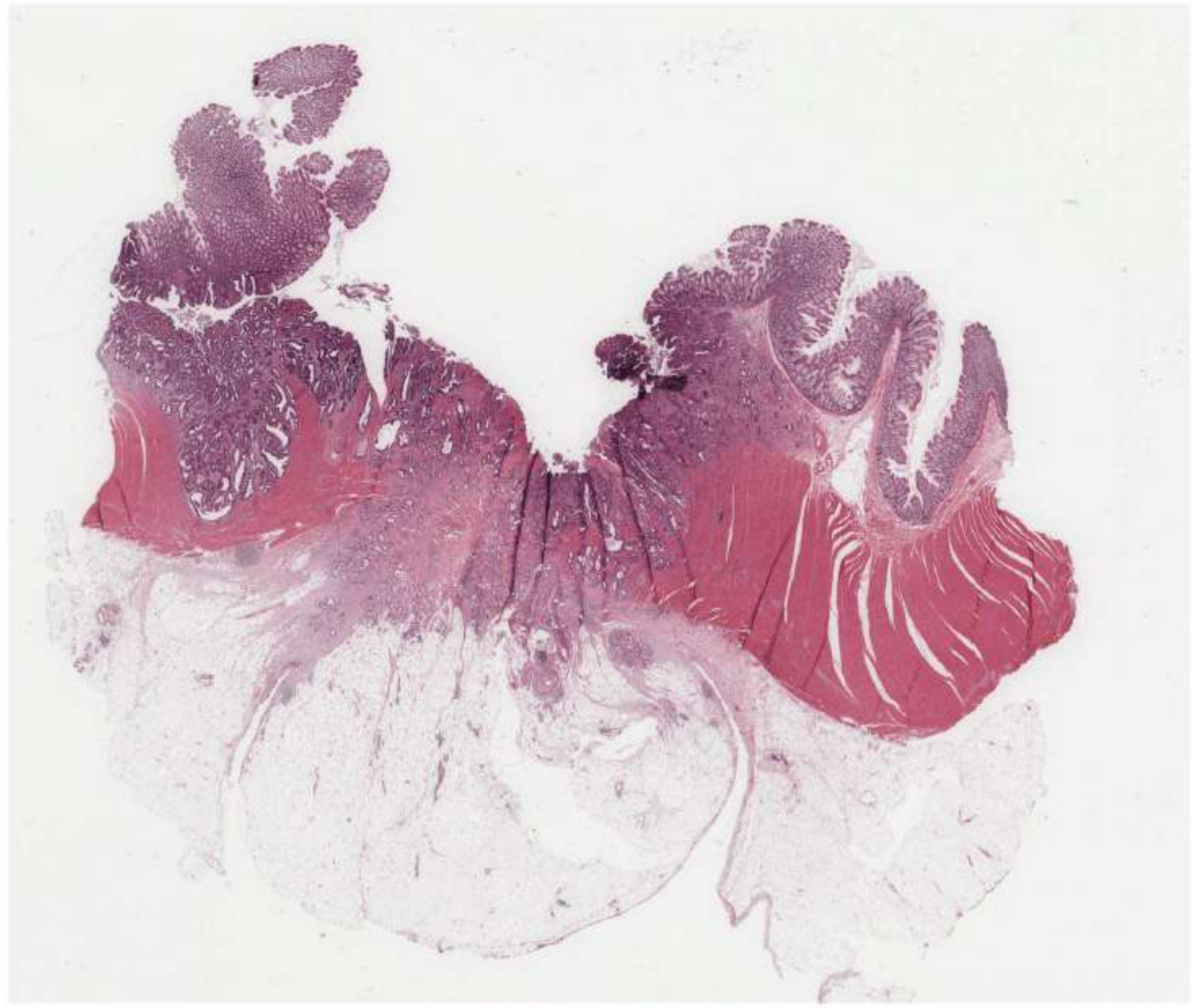

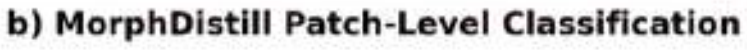

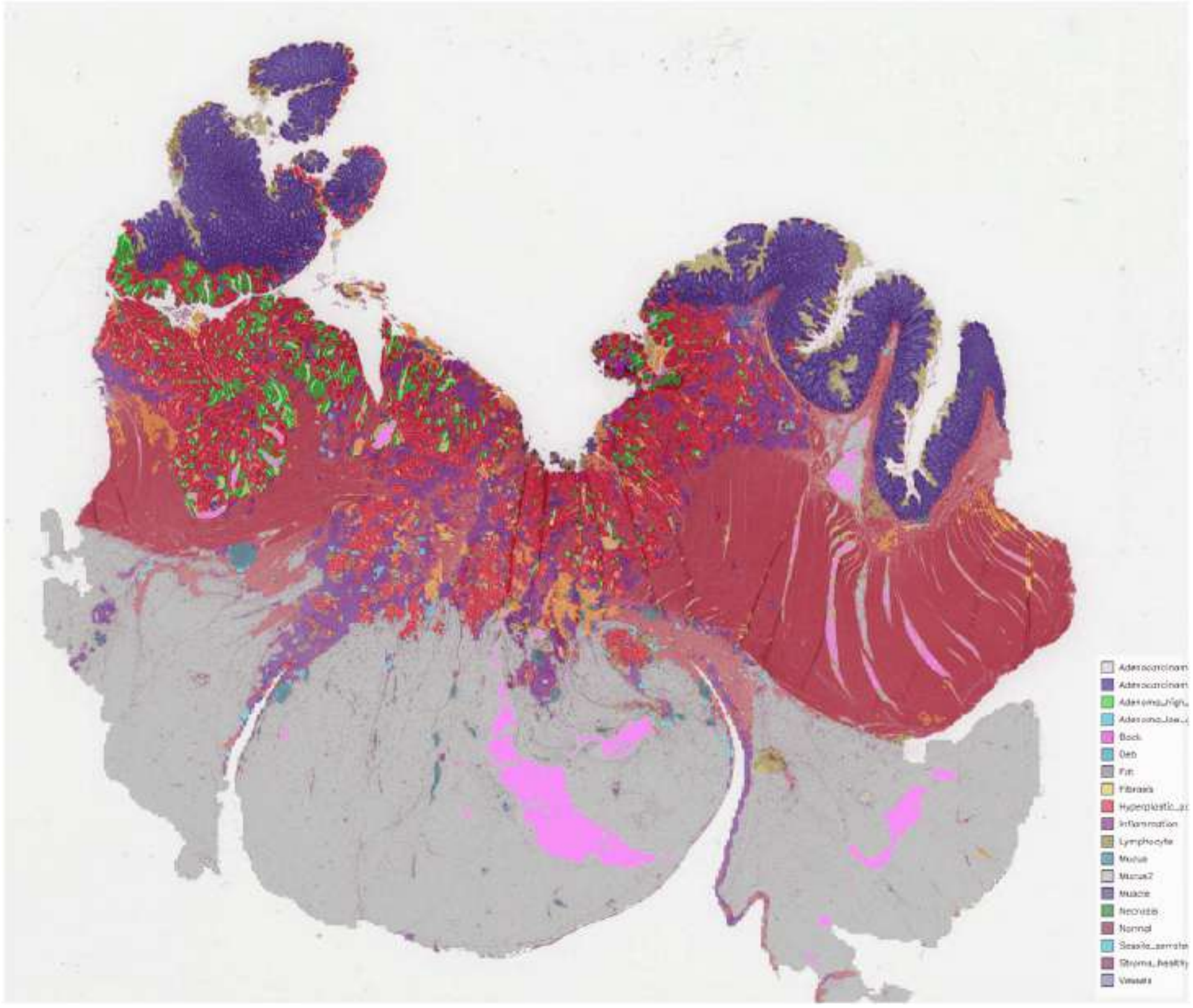

**Figure S18:** Morphological inference by the MorphDistill encoder on a whole-slide image. (a) Original H&E-stained slide from the Alliance cohort. (b) Patch-level tissue classification map generated by the frozen MorphDistill encoder, with colors indicating different tissue types. The clear delineation of diverse morphological classes — including adenocarcinoma, lymphocytes, mucus, muscle, stroma, and vessels etc. — demonstrates the encoder's ability to preserve CRC-relevant tissue semantics after unifying knowledge from ten foundation models (best viewed in color).

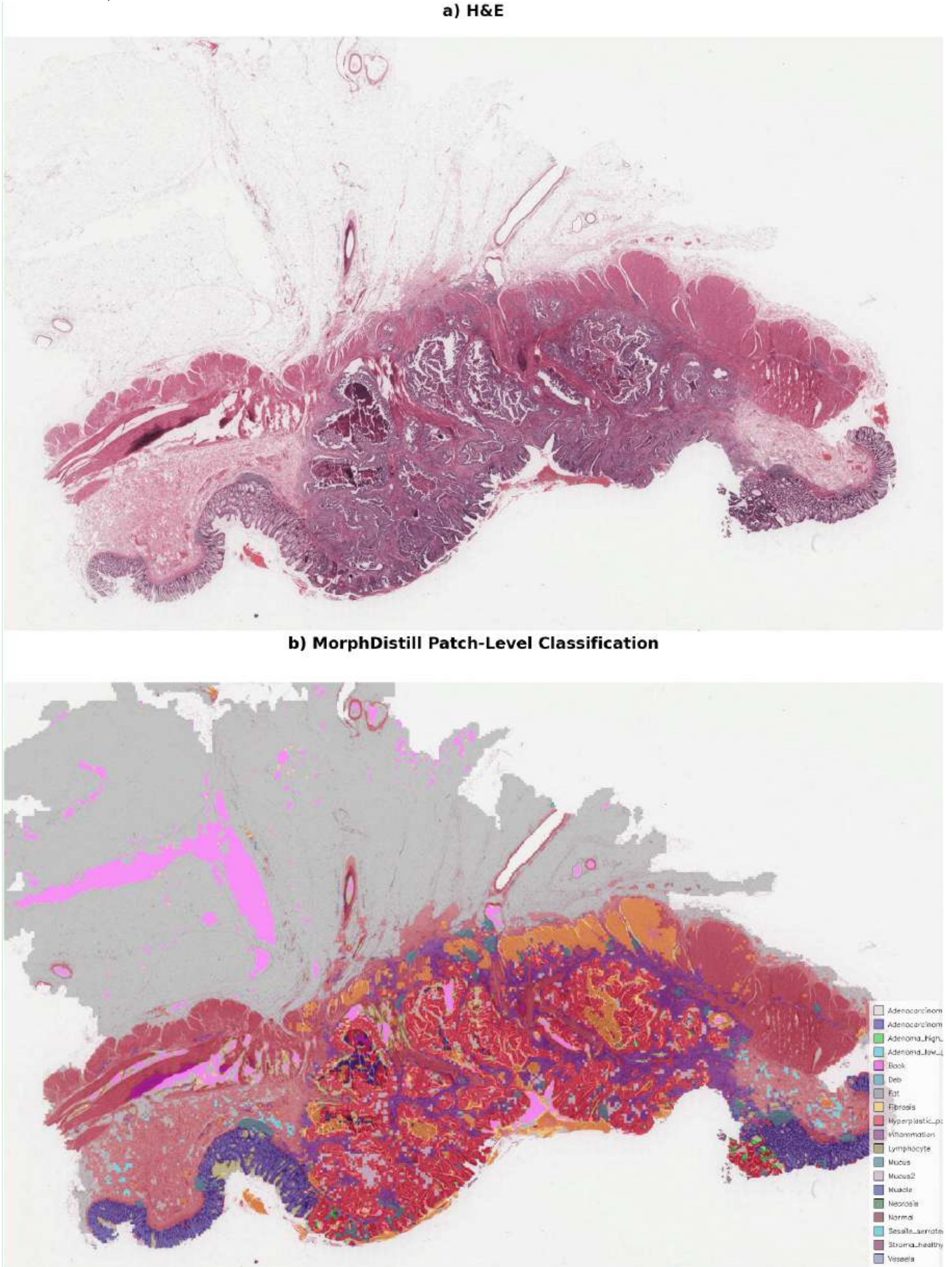

**Figure S19:** Morphological inference by the MorphDistill encoder on a whole-slide image. (a) Original H&E-stained slide from the Alliance cohort. (b) Patch-level tissue classification map generated by the frozen MorphDistill encoder, with colors indicating different tissue types. The clear delineation of diverse morphological classes — including adenocarcinoma, lymphocytes, mucus, muscle, stroma, and vessels etc. — demonstrates the encoder's ability to preserve CRC-relevant tissue semantics after unifying knowledge from ten foundation models (best viewed in color).

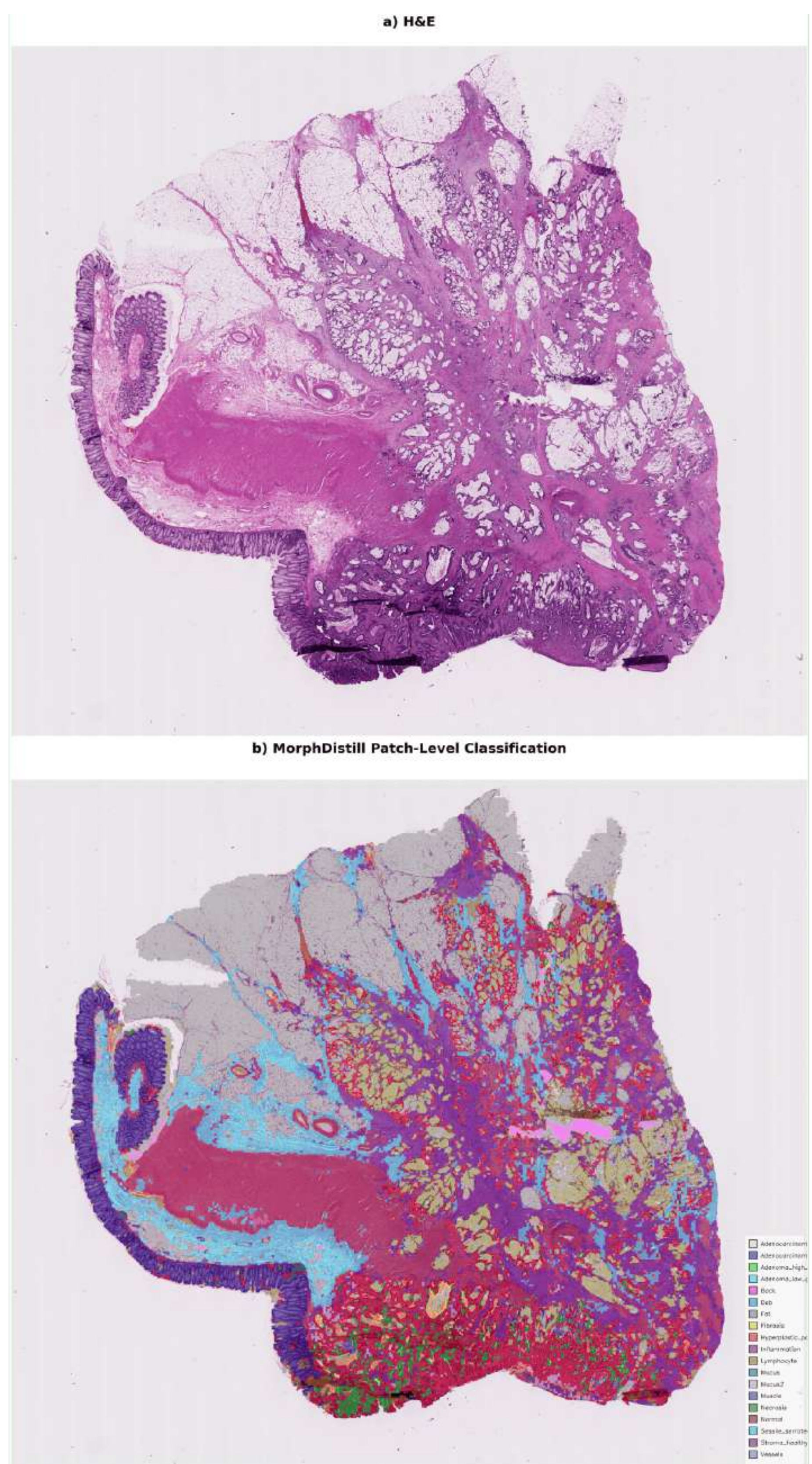

**Figure S19:** Morphological inference by the MorphDistill encoder on a whole-slide image. (a) Original H&E-stained slide from the Alliance cohort. (b) Patch-level tissue classification map generated by the frozen MorphDistill encoder, with colors indicating different tissue types. The clear delineation of diverse morphological classes — including adenocarcinoma, lymphocytes, mucus, muscle, stroma, and vessels etc. — demonstrates the encoder's ability to preserve CRC-relevant tissue semantics after unifying knowledge from ten foundation models (best viewed in color).

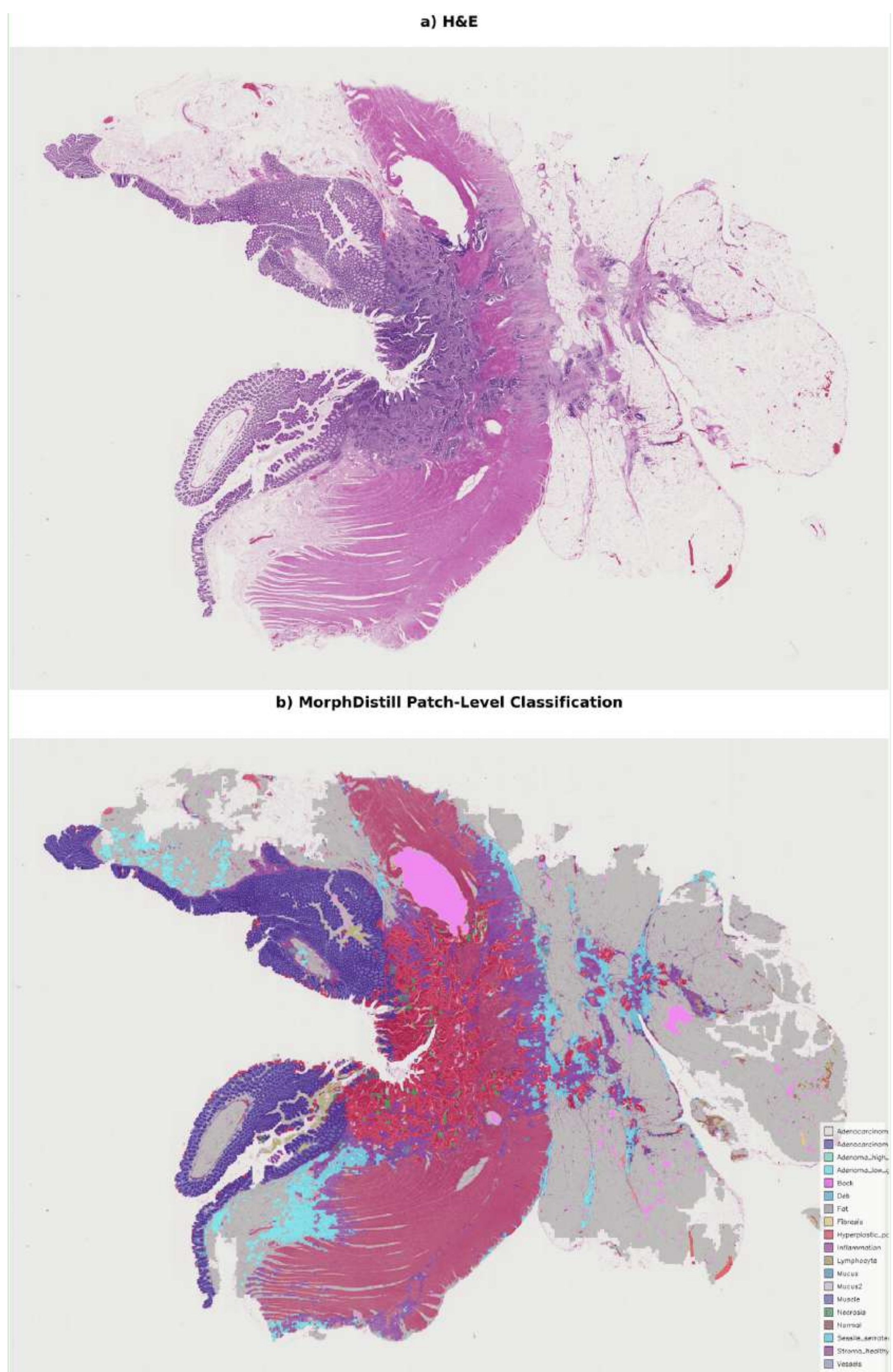

**Figure S20:** Morphological inference by the MorphDistill encoder on a whole-slide image. (a) Original H&E-stained slide from the Alliance cohort. (b) Patch-level tissue classification map generated by the frozen MorphDistill encoder, with colors indicating different tissue types. The clear delineation of diverse morphological classes — including adenocarcinoma, lymphocytes, mucus, muscle, stroma, and vessels etc. — demonstrates the encoder's ability to preserve CRC-relevant tissue semantics after unifying knowledge from ten foundation models (best viewed in color).

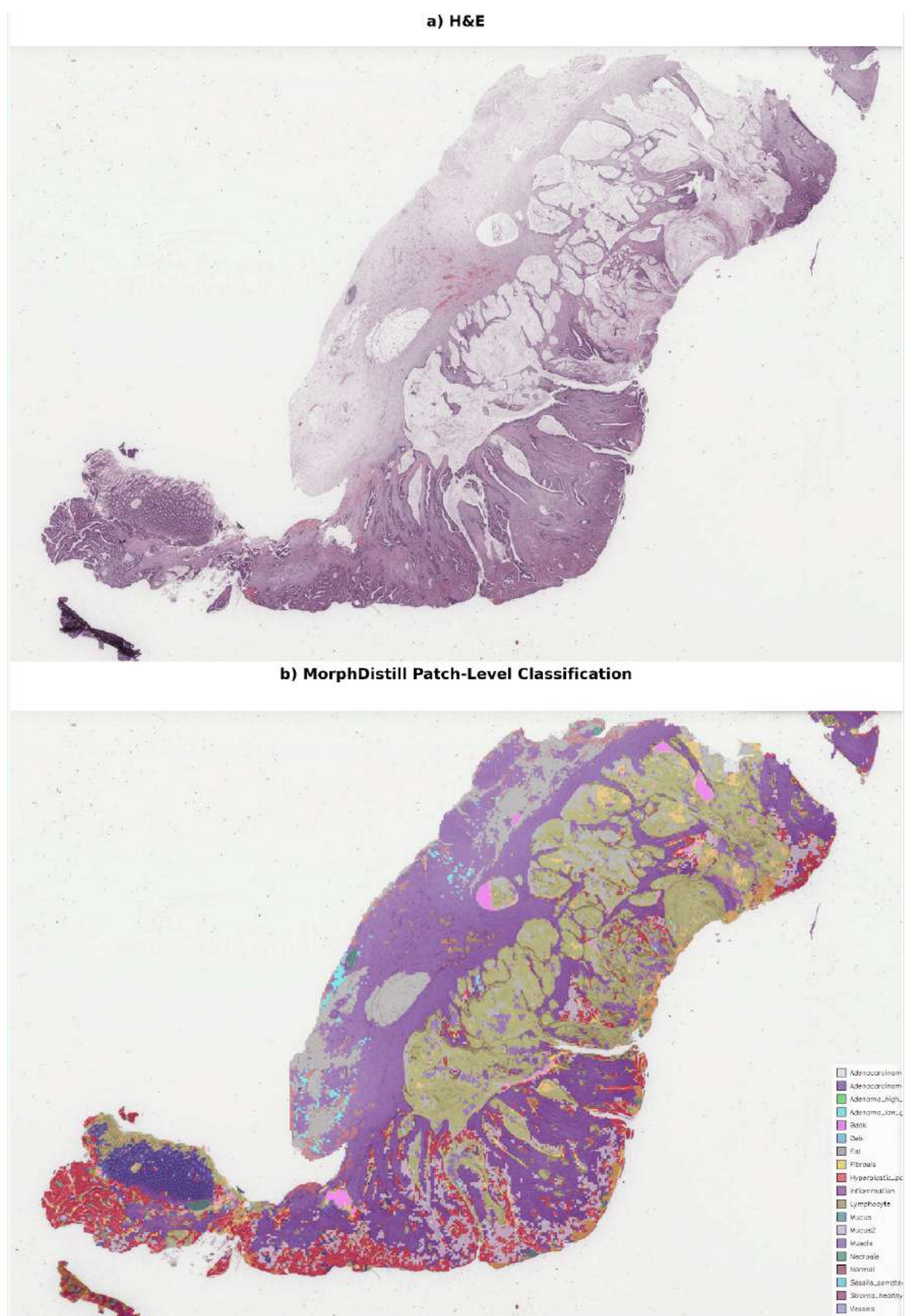

**Figure S21:** Morphological inference by the MorphDistill encoder on a whole-slide image. (a) Original H&E-stained slide from the Alliance cohort. (b) Patch-level tissue classification map generated by the frozen MorphDistill encoder, with colors indicating different tissue types. The clear delineation of diverse morphological classes — including adenocarcinoma, lymphocytes, mucus, muscle, stroma, and vessels etc. — demonstrates the encoder's ability to preserve CRC-relevant tissue semantics after unifying knowledge from ten foundation models (best viewed in color).

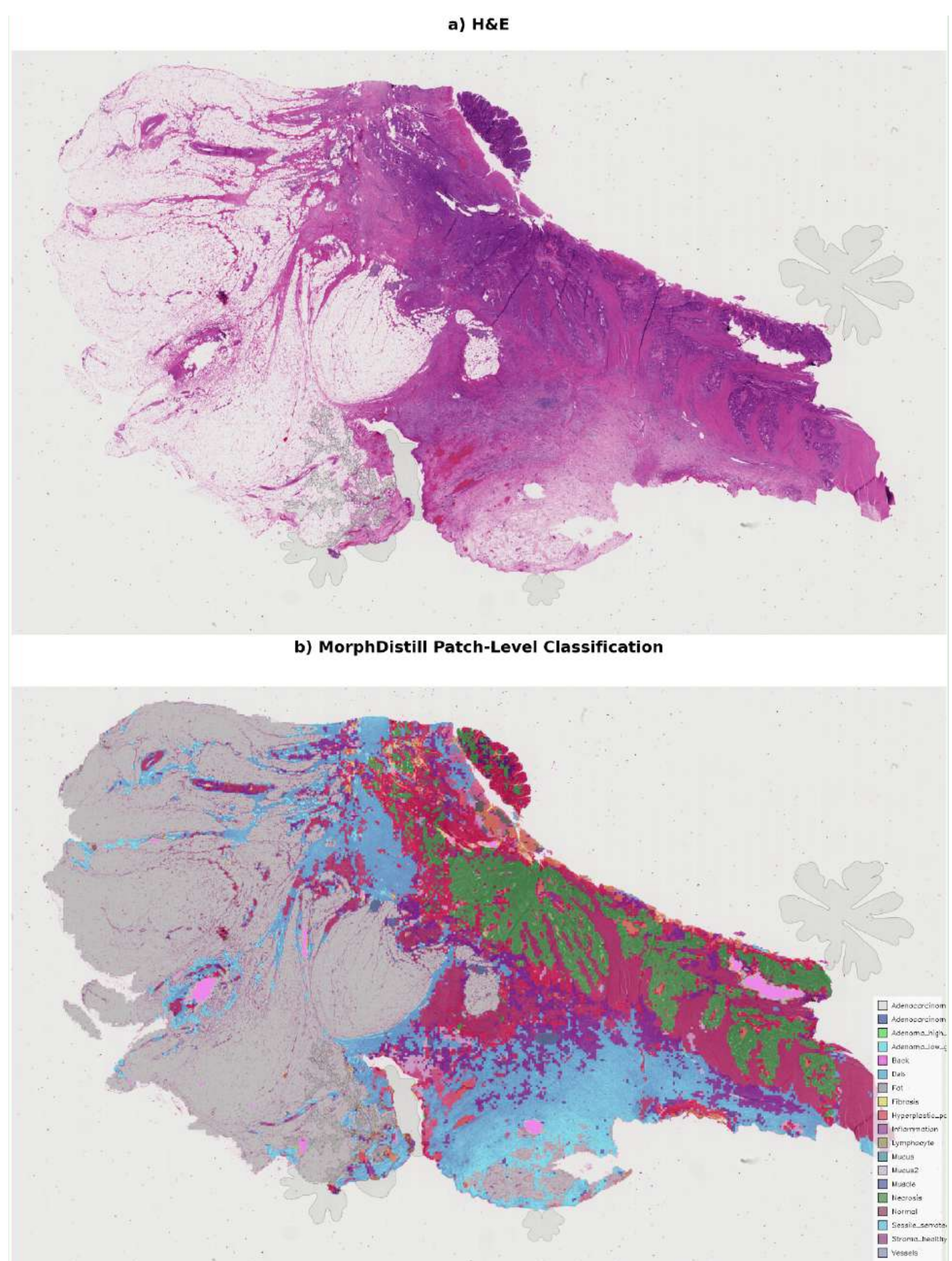

**Figure S22:** Morphological inference by the MorphDistill encoder on a whole-slide image. (a) Original H&E-stained slide from the Alliance cohort. (b) Patch-level tissue classification map generated by the frozen MorphDistill encoder, with colors indicating different tissue types. The clear delineation of diverse morphological classes — including adenocarcinoma, lymphocytes, mucus, muscle, stroma, and vessels etc. — demonstrates the encoder's ability to preserve CRC-relevant tissue semantics after unifying knowledge from ten foundation models (best viewed in color).

**a) H&E**

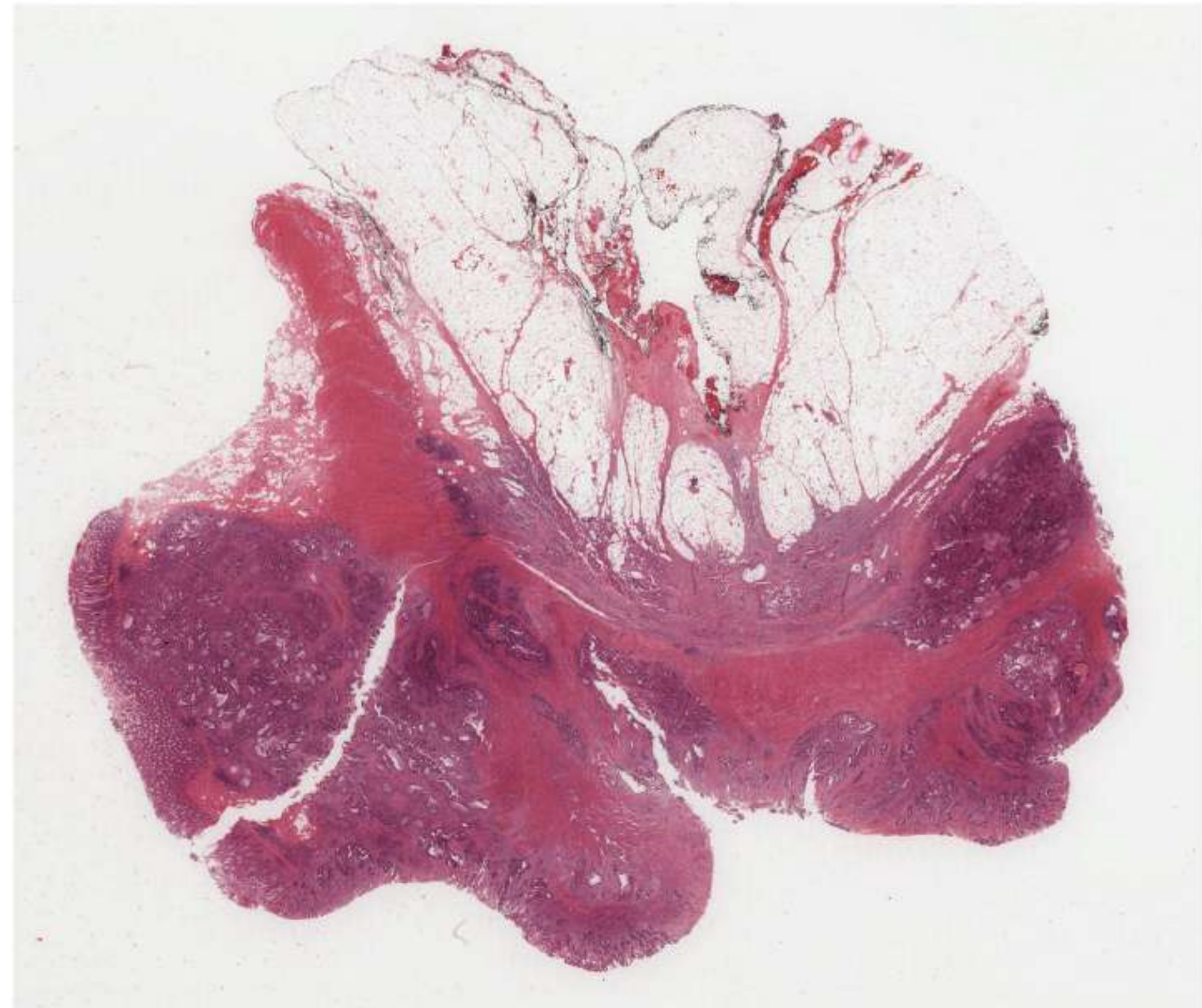

**b) MorphDistill Patch-Level Classification**

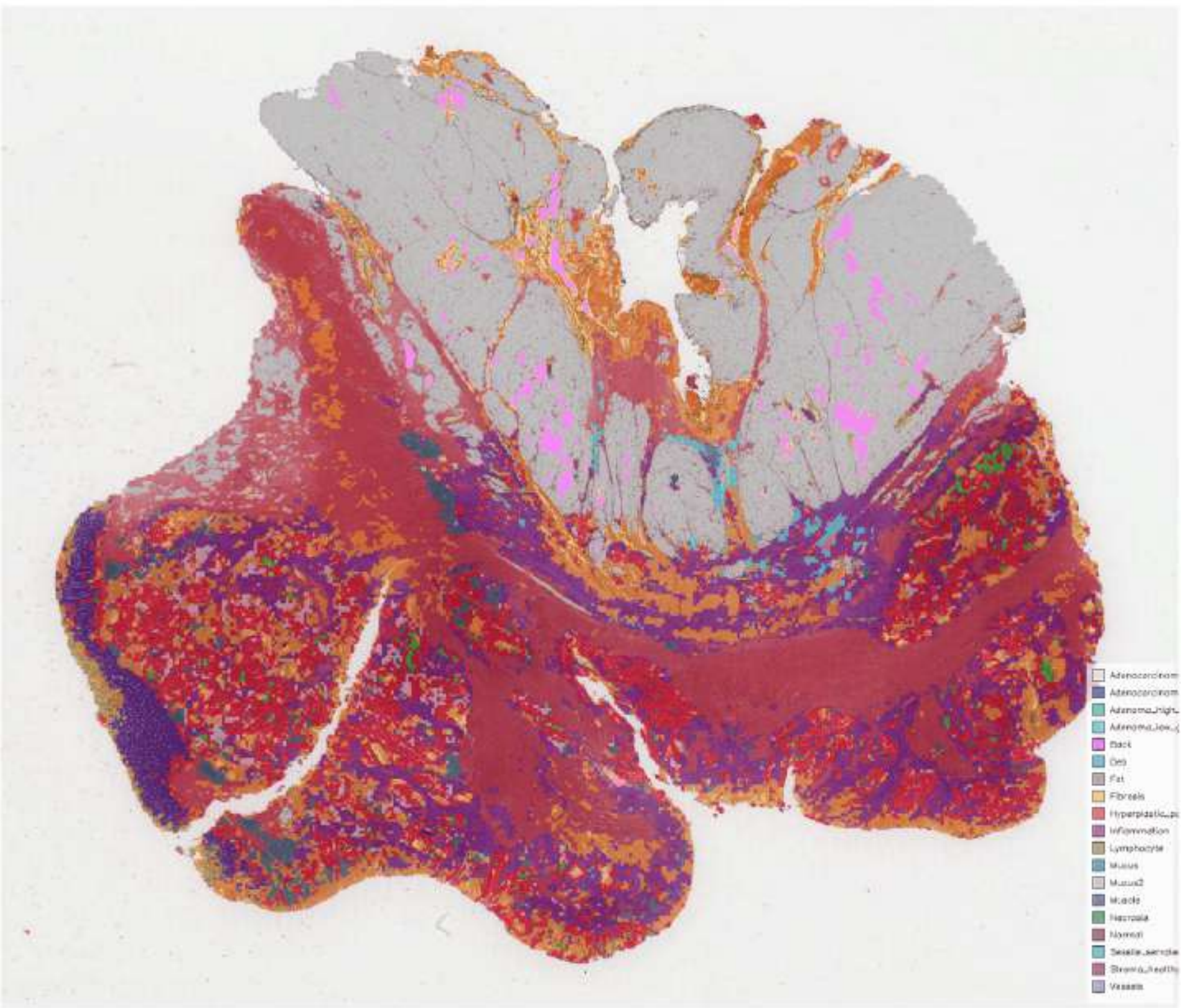

**Figure S23:** Morphological inference by the MorphDistill encoder on a whole-slide image. (a) Original H&E-stained slide from the Alliance cohort. (b) Patch-level tissue classification map generated by the frozen MorphDistill encoder, with colors indicating different tissue types. The clear delineation of diverse morphological classes — including adenocarcinoma, lymphocytes, mucus, muscle, stroma, and vessels etc. — demonstrates the encoder's ability to preserve CRC-relevant tissue semantics after unifying knowledge from ten foundation models (best viewed in color).

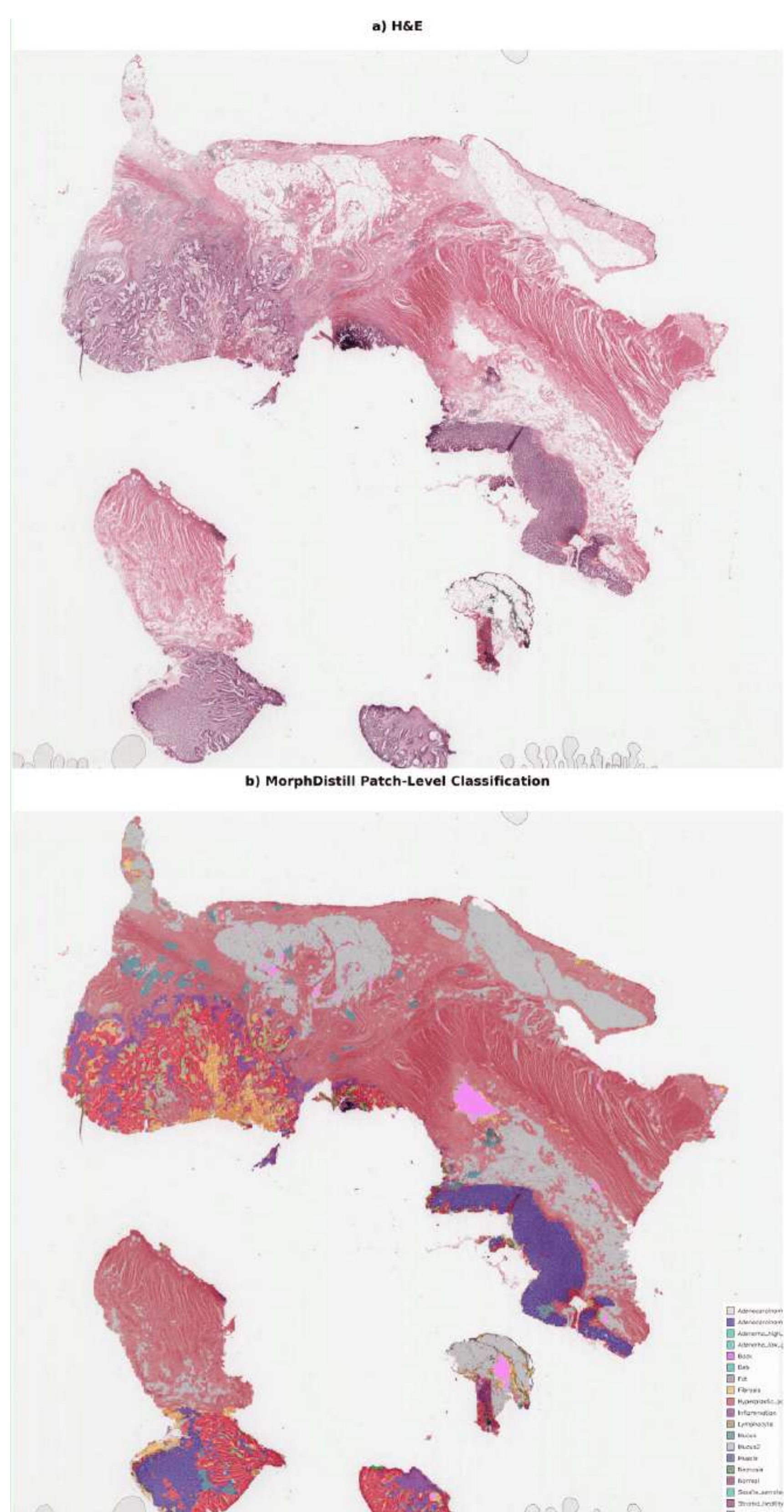

**Figure S24:** Morphological inference by the MorphDistill encoder on a whole-slide image. (a) Original H&E-stained slide from the Alliance cohort. (b) Patch-level tissue classification map generated by the frozen MorphDistill encoder, with colors indicating different tissue types. The clear delineation of diverse morphological classes — including adenocarcinoma, lymphocytes, mucus, muscle, stroma, and vessels etc. — demonstrates the encoder's ability to preserve CRC-relevant tissue semantics after unifying knowledge from ten foundation models (best viewed in color).

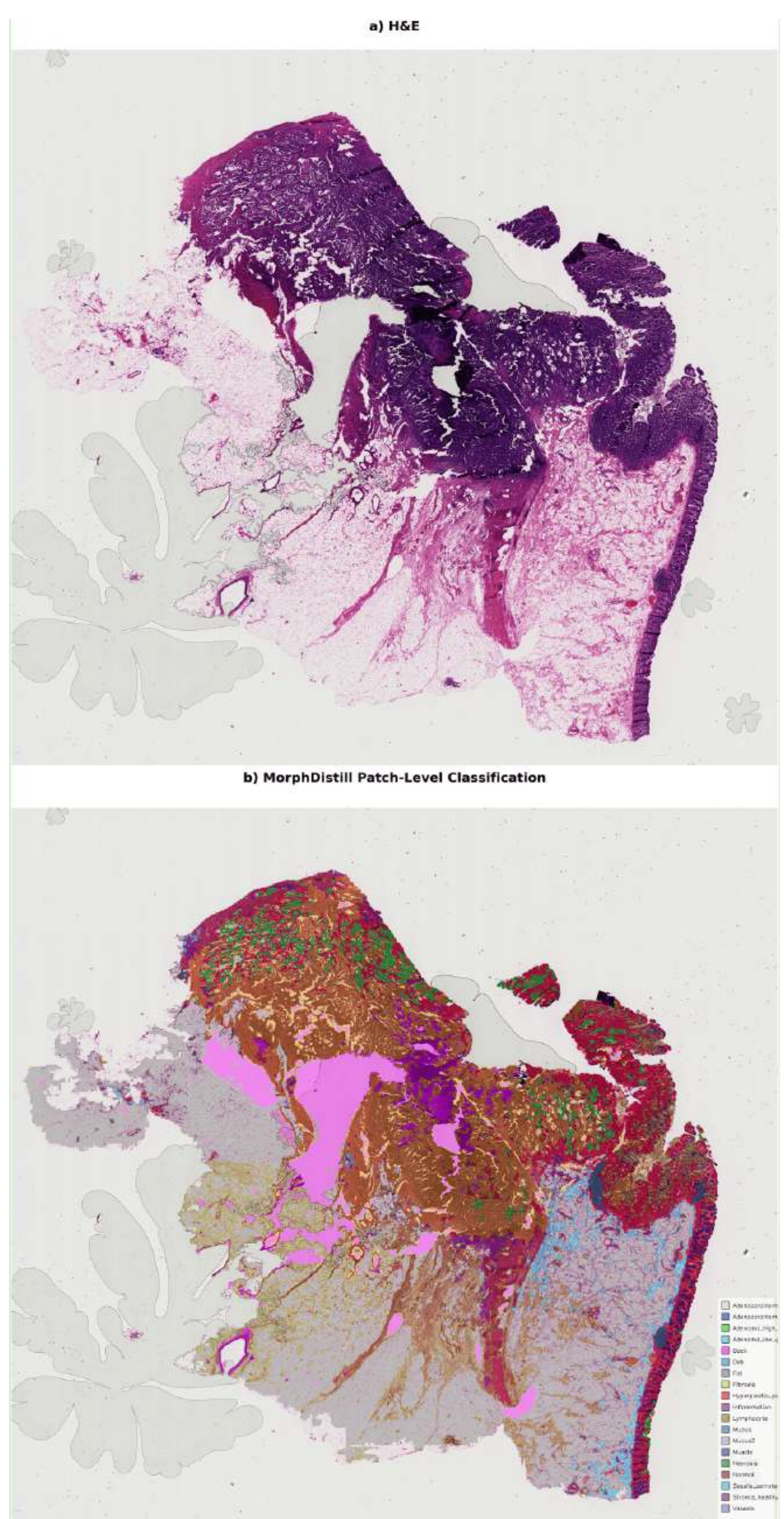

**Figure S25:** Morphological inference by the MorphDistill encoder on a whole-slide image. (a) Original H&E-stained slide from the Alliance cohort. (b) Patch-level tissue classification map generated by the frozen MorphDistill encoder, with colors indicating different tissue types. The clear delineation of diverse morphological classes — including adenocarcinoma, lymphocytes, mucus, muscle, stroma, and vessels etc. — demonstrates the encoder's ability to preserve CRC-relevant tissue semantics after unifying knowledge from ten foundation models (best viewed in color).

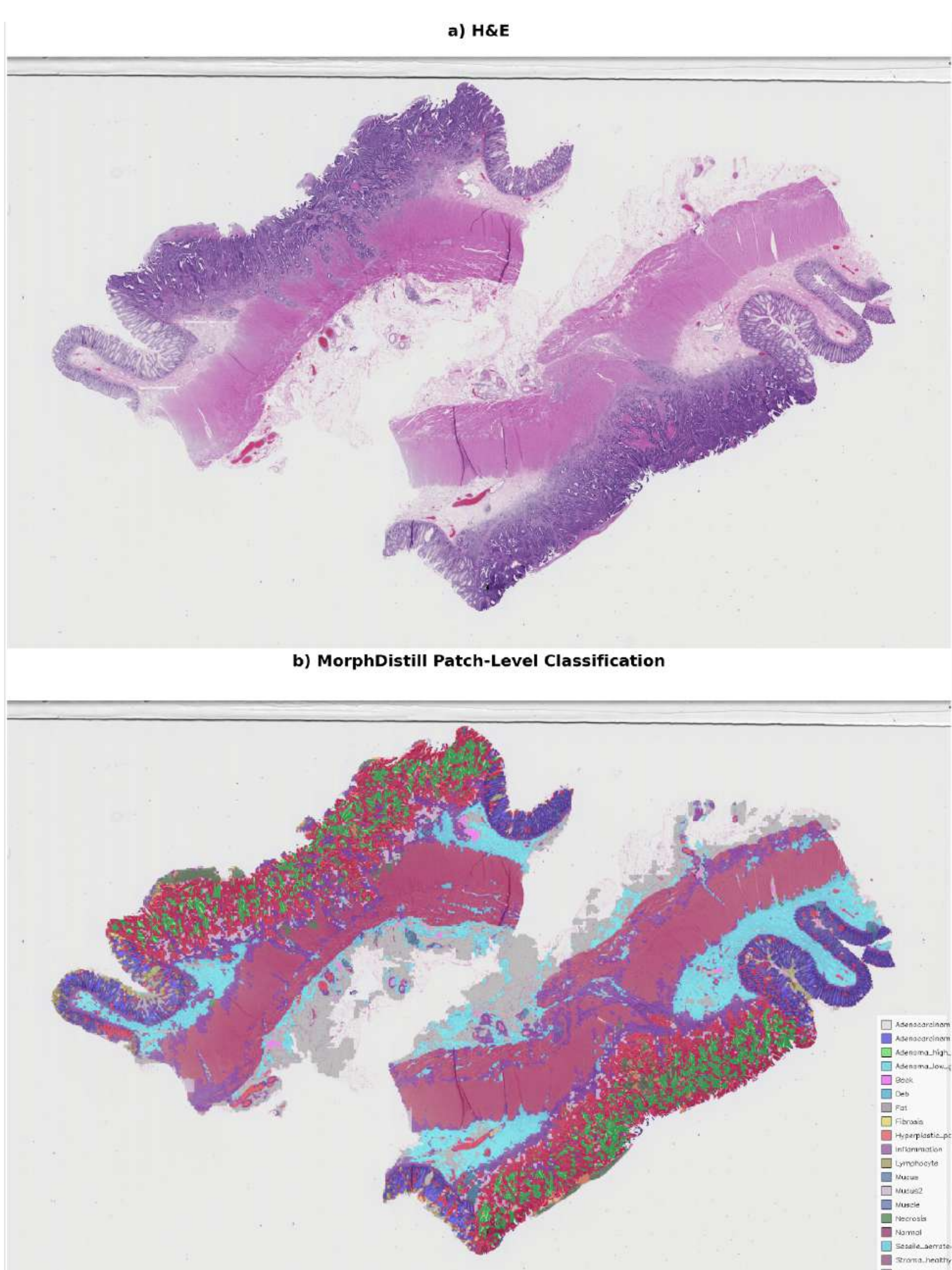

**Figure S26:** Morphological inference by the MorphDistill encoder on a whole-slide image. (a) Original H&E-stained slide from the Alliance cohort. (b) Patch-level tissue classification map generated by the frozen MorphDistill encoder, with colors indicating different tissue types. The clear delineation of diverse morphological classes — including adenocarcinoma, lymphocytes, mucus, muscle, stroma, and vessels etc. — demonstrates the encoder's ability to preserve CRC-relevant tissue semantics after unifying knowledge from ten foundation models (best viewed in color).